\documentclass{article} % For LaTeX2e
\usepackage{iclr2022_conference,times}

\usepackage{amsmath,amsfonts,bm}
\usepackage{nicefrac}
\usepackage{xspace}
\usepackage{graphicx}
\usepackage{algorithm}
\usepackage{algorithmicx}
\usepackage[noend]{algpseudocode}
\algrenewcommand\algorithmicindent{4px}
\usepackage{tcolorbox}
\usepackage{multirow}
\usepackage{caption}
\usepackage{tikz}
\usepackage{enumitem}
\usepackage[frozencache,cachedir=.]{minted}

\definecolor{gray}{HTML}{AAAAAA}
\definecolor{seaborn1}{HTML}{4C72B0}
\definecolor{seaborn2}{HTML}{55A868}
\definecolor{seaborn3}{HTML}{C44E52}
\definecolor{seaborn4}{HTML}{8172B2}
\definecolor{seaborn5}{HTML}{CCB974}
\definecolor{seaborn6}{HTML}{64B5CD}

\def\figref#1{Fig.~\ref{#1}}
\def\secref#1{Sec.~\ref{#1}}
\def\appref#1{App.~\ref{#1}}
\def\eqnref#1{Eq.~\eqref{#1}}
\def\algref#1{Alg.~\ref{#1}}
\def\tabref#1{Tab.~\ref{#1}}

\def\1{\bm{1}}

\DeclareMathAlphabet{\mathsfit}{\encodingdefault}{\sfdefault}{m}{sl}
\SetMathAlphabet{\mathsfit}{bold}{\encodingdefault}{\sfdefault}{bx}{n}

\newcommand{\p}{P}

\DeclareMathOperator*{\argmax}{arg\,max}

\usepackage{xspace}
\makeatletter
\DeclareRobustCommand\onedot{\futurelet\@let@token\@onedot}
\def\@onedot{\ifx\@let@token.\else.\null\fi\xspace}
\def\eg{\emph{e.g}\onedot} 
\def\ie{\emph{i.e}\onedot} 
\def\cf{\emph{c.f}\onedot} 
\def\etc{\emph{etc}\onedot} 
\def\wrt{w.r.t\onedot} 

\makeatother

\newcommand{\CP}{CP\xspace}
\newcommand{\APS}{\textsc{APS}\xspace}
\newcommand{\RAPS}{\textsc{RAPS}\xspace}
\newcommand{\Thr}{\textsc{Thr}\xspace}
\newcommand{\ThrL}{\textsc{ThrL}\xspace}
\newcommand{\ThrLP}{\textsc{ThrLP}\xspace}
\newcommand{\CovT}{CoverTr\xspace}
\newcommand{\CT}{ConfTr\xspace}
\newcommand{\Bel}{Bel\xspace}
\newcommand{\Belc}{Bel\xspace}

\newcommand\Cov{\relax\ifmmode\text{Cover}\else Cover\xspace\fi}
\newcommand\Acc{\relax\ifmmode\text{Acc}\else Acc\xspace\fi}
\newcommand\Ineff{\relax\ifmmode\text{Ineff}\else Ineff\xspace\fi}
\newcommand\Lcov{\relax\ifmmode\mathcal{L}_{\text{cov}}\else $\mathcal{L}_{\text{cov}}$\xspace\fi}
\newcommand\Lclass{\relax\ifmmode\mathcal{L}_{\text{class}}\else $\mathcal{L}_{\text{class}}$\xspace\fi}
\newcommand\Ical{\relax\ifmmode I_{\text{cal}}\else $I_{\text{cal}}$\xspace\fi}
\newcommand\Itest{\relax\ifmmode I_{\text{test}}\else $I_{\text{test}}$\xspace\fi}
\newcommand\Bcal{\relax\ifmmode B_{\text{cal}}\else $B_{\text{cal}}$\xspace\fi}
\newcommand\Bpred{\relax\ifmmode B_{\text{pred}}\else $B_{\text{pred}}$\xspace\fi}
\newcommand\MisCov{\relax\ifmmode\text{MisCover}\else MisCover\xspace\fi}

\newcommand{\red}[1]{\noindent{\color{red}{#1}}}
\renewcommand{\red}[1]{\noindent{#1}}

\usepackage{hyperref}
\usepackage{url}

\title{Learning Optimal Conformal Classifiers}
\author{%
David Stutz$^{1,2}$, Krishnamurthy (Dj) Dvijotham$^1$, Ali Taylan Cemgil$^1$, Arnaud Doucet$^1$\\
$^1$ DeepMind, $^2$ Max Planck Institute for Informatics, Saarland Informatics Campus
}

\iclrfinalcopy
\begin{document}

\maketitle

\begin{abstract}
    Modern deep learning based classifiers 
    show very high accuracy on test data but this does \emph{not} provide sufficient guarantees for safe deployment, especially in high-stake AI applications such as medical diagnosis.
    Usually, predictions are obtained without a reliable uncertainty estimate or a formal guarantee.
    \emph{Conformal prediction (\CP)} addresses these issues by using the classifier's \red{predictions, \eg, its probability estimates,} to predict \emph{confidence sets} containing the true class with a user-specified probability.
    However, using \CP as a separate processing step after training prevents the underlying model from adapting to the prediction of confidence sets.
    Thus, this paper explores strategies to differentiate through \CP \emph{during training} with the goal of training model with the conformal wrapper \emph{end-to-end}.
    In our approach, \textbf{conformal training (\CT)}, we specifically ``simulate'' conformalization on mini-batches during training.
    \red{Compared to standard training, \CT reduces the average confidence set size (\emph{inefficiency}) of state-of-the-art \CP methods applied after training.}
    Moreover, it allows to ``shape'' the confidence sets predicted at test time, which is difficult for standard \CP.
    On experiments with several datasets, we show \CT can influence how inefficiency is distributed across classes, or guide the composition of confidence sets in terms of the included classes, while retaining the guarantees offered by \CP.
\end{abstract}
\section{Introduction}

In classification tasks, for input $x$, we approximate the posterior distribution over classes $y \in [K] := \{1, \ldots, K\}$, denoted $\pi_y(x) \approx \p(Y = y| X = x)$.
Following Bayes' decision rule, the \emph{single} class with highest posterior probability is predicted \red{for optimizing a 0-1 classification loss}.
This way, deep networks $\pi_{\theta,y}(x)$ with parameters $\theta$ achieve impressive accuracy on held-out test sets.
However, this does not \emph{guarantee} safe deployment.
\emph{Conformal prediction (\CP)} \citep{Vovk2005} uses a post-training calibration step to \emph{guarantee} a user-specified \emph{coverage}:
by allowing to predict confidence sets $C(X) \subseteq [K]$, \CP guarantees the true class $Y$ to be included with confidence level $\alpha$, \ie $\p(Y \in C(X)) \geq 1 - \alpha$ when the calibration examples $(X_i, Y_i)$, $i \in \Ical$ are drawn exchangeably from the test distribution.
This is usually achieved in two steps:
In the \emph{prediction step}, so-called \emph{conformity scores} (\wrt to a class $k \in [K]$) are computed to construct the confidence sets $C(X)$.
During the \emph{calibration step}, these conformity scores on the calibration set \wrt the true class $Y_i$ are ranked to determine a cut-off threshold $\tau$ for the predicted probabilities $\pi_\theta(x)$ guaranteeing coverage $1 - \alpha$.
This is called \emph{marginal} coverage as it holds only unconditionally, \ie, the expectation
is being taken not only \wrt $(X, Y)$ but also over the distribution of all possible calibration sets, rather than \wrt the conditional distribution $p(Y| X)$.

\CP
also outputs intuitive uncertainty estimates:
larger confidence sets $|C(X)|$ generally convey higher uncertainty.
Although \CP is agnostic to details of the underlying model $\pi_\theta(x)$, the obtained uncertainty estimates depend strongly on the model's performance.
If the underlying classifier is poor, \CP results in too large and thus uninformative confidence sets.
``Uneven'' coverage is also a common issue, where lower coverage is achieved on more difficult classes.
To address such problems, the threshold \CP method of \citep{SadinleJASA2019} explicitly minimizes inefficiency.
\cite{RomanoNIPS2020} and \cite{CauchoisARXIV2020} propose methods that perform favorably in terms of (approximate) conditional coverage.
The \emph{adaptive prediction sets (\APS)} method of \cite{RomanoNIPS2020} is further extended by \cite{AngelopoulosICLR2020} to return smaller confidence sets.
These various objectives are typically achieved by changing the definition of the conformity scores.
In all cases, \CP is used as a post-training calibration step.
In contrast, our work does \emph{not} focus on advancing \CP itself, \eg, through new conformity scores, but develops a novel training procedure for the classifier $\pi_\theta$. \red{After training, any of the above \CP methods can readily be applied.}

Indeed, while the flexibility of \CP regarding the underlying model appears attractive, it is also a severe limitation:
Learning the model parameters $\theta$ is \emph{not} informed about the post-hoc ``conformalization'', \ie, they are are not tuned towards any specific objective such as reducing expected confidence set size (\emph{inefficiency}).
During training, the model will typically be trained to minimize cross-entropy loss.
At test time, in contrast, it is used to obtain a set predictor $C(X)$ with specific properties such as low inefficiency.
In concurrent work, \cite{BellottiARXIV2021} addresses this issue by learning a set predictor $C(X)$ through thresholding logits:
Classes with logits exceeding $1$ are included in $C(X)$ and training aims to minimize inefficiency while targeting coverage $1 - \alpha$.
In experiments using linear models only, this approach is shown to decrease inefficiency.
However, \citep{BellottiARXIV2021} ignores the crucial calibration step of \CP during training and does \emph{not} allow to optimize losses beyond marginal coverage or inefficiency.
In contrast, our work subsumes \citep{BellottiARXIV2021}, but additionally considers the calibration step during training, which is crucial for further decreasing inefficiency.
Furthermore, we aim to allow fine-grained control over class-conditional inefficiency or the composition of the confidence sets by allowing to optimize arbitrary losses defined on confidence sets.

\begin{figure}
    \vspace*{-0.3cm}
    \centering
    \begin{tikzpicture}
        \node[anchor=north west] at (0,0){\includegraphics[width=13cm]{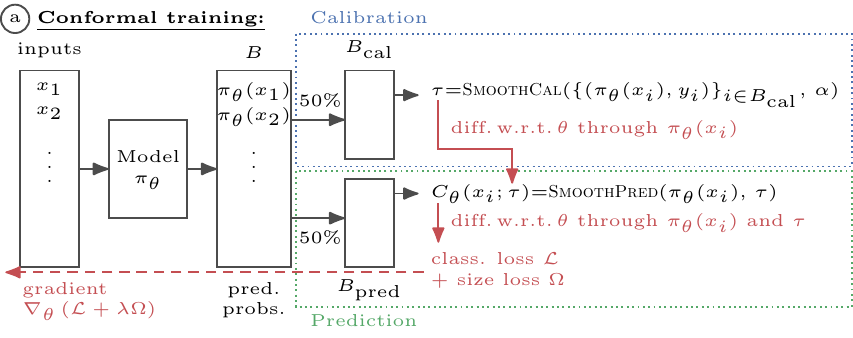}};
        \node[anchor=north west,fill=white] at (8.75,-3.65){\includegraphics[width=4.25cm]{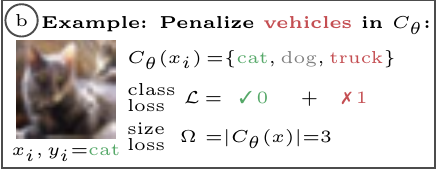}};
    \end{tikzpicture}
    \vspace*{-12px}
    \caption{
    \textbf{Illustration of \emph{conformal training (\CT)}:}
    We develop differentiable prediction and calibration steps for \emph{conformal prediction (\CP)}, \textsc{SmoothCal} and \textsc{SmoothPred}.
    During training, this allows \CT to ``simulate'' \CP on each mini-batch $B$ by calibrating on the first half $B_{\text{cal}}$ and predicting confidence sets on the other half $B_{\text{pred}}$ (\cf \textcircled{a}).
    \red{\CT can optimize} arbitrary losses on the predicted confidence sets, \eg, reducing average confidence set size (\emph{inefficiency}) using a size loss $\Omega$ or penalizing specific classes from being included using a classification loss $\mathcal{L}$ (\cf \textcircled{b}). \red{\emph{After} training using our method, \emph{any} existing \CP method can be used to obtain a coverage guarantee.}
    }
    \label{fig:introduction}
    \vspace*{-0.1cm}
\end{figure}

Our \textbf{contributions} can be summarized as follows:
\vspace*{-6px}
\begin{enumerate}[leftmargin=0.65cm]
    \item We propose \textbf{conformal training (\CT)}, a procedure allowing to train model and conformal wrapper \emph{end-to-end}.
    This is achieved by developing smooth implementations of recent \CP methods \red{for use during training}.
    On each mini-batch, \CT ``simulates'' conformalization, using half of the batch for calibration, and the other half for \red{prediction and loss computation}, \cf \figref{fig:introduction} \textcircled{a}. \red{After training, \emph{any} existing \CP method can provide a coverage guarantee.}
    \vspace*{-2px}
    \item In experiments, \red{using \CT for training} consistently reduces the inefficiency of conformal predictors such as \emph{threshold \CP (\Thr)} \citep{SadinleJASA2019} or \APS \citep{RomanoNIPS2020} \red{applied \emph{after} training}.
    We further improve over \citep{BellottiARXIV2021}, illustrating the importance of the calibration step during training.
    \vspace*{-2px}
    \item Using carefully constructed losses, \CT allows to ``shape'' the confidence sets obtained at test time:
    We can reduce \emph{class-conditional} inefficiency or ``coverage confusion'', \ie, the likelihood of two or more classes being included in the same confidence sets, \cf \figref{fig:introduction} \textcircled{b}.
    Generally, in contrast to \citep{BellottiARXIV2021}, \CT allows to optimize arbitrary losses on the confidence sets.
\end{enumerate}
Because \CT is agnostic to the \CP method used at test time, our work is complementary to most related work, \red{\ie, \emph{any} advancement in terms of \CP is directly applicable to \CT.}
\red{For example, this might include conditional or application-specific guarantees as in \citep{SadinleARXIV2016,BatesARXIV2021}.}
Most importantly, \CT preserves the coverage guarantee obtained through \CP.
\section{Differentiable Conformal Predictors}
\label{sec:conformal-prediction}

We are interested in training the model $\pi_\theta$ end-to-end with the conformal wrapper in order to allow fine-grained control over the confidence sets $C(X)$.
Before developing differentiable \CP methods \red{for training} in \secref{subsec:differentiable-conformal-predictors}, we review two \red{recently proposed} conformal predictors \red{that we use at test time. These}
consist of two steps, \red{see \secref{subsec:conformal-predictors}}:
for \emph{prediction} (on the test set) we need to define the confidence sets $C_\theta(X; \tau)$ which depend on the model parameters $\theta$ through the predictions $\pi_\theta$ and where the threshold $\tau$ is determined during \emph{calibration} on a held-out calibration set $(X_i, Y_i), i \in \Ical$ in order to obtain coverage.

\subsection{Conformal Predictors}
\label{subsec:conformal-predictors}

\begin{figure}
    \centering
    \vspace*{-0.2cm}
    \begin{minipage}[t]{1.0\textwidth}
        \small
        \centering
        \begin{tabular}{|l|c|c|c|c|}
            \hline
            \multicolumn{5}{|c|}{\CP Baseline Comparison by \Ineff}\\
            \hline
            Dataset, $\alpha$ & \ThrL & \Thr & \APS & \RAPS\\
            \hline
            \hline
            \hline
            CIFAR10, $0.05$ & 2.22 & \bfseries 1.64 & 2.06 & 1.74\\
            CIFAR10, $0.01$ & 3.92 & \bfseries 2.93 & 3.30 & 3.06\\
            \hline
            CIFAR100, $0.01$ & 19.22 & \bfseries 10.63 & 16.62 & 14.25\\
            \hline
        \end{tabular}
        \hskip 2px
        \raisebox{-1.2cm}{\includegraphics[height=2.5cm]{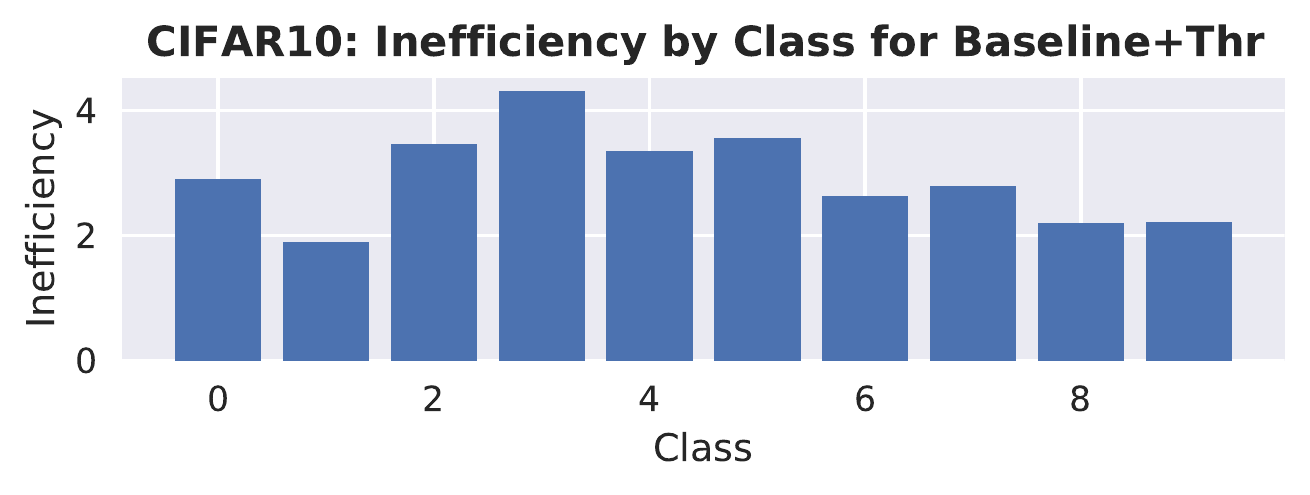}}
        
    \end{minipage}
    \vspace*{-12px}
    \caption{
    \textbf{Baseline \CP Results on CIFAR:}
    \textit{Left:} Inefficiency (\Ineff, lower is better) for the \CP methods discussed in \secref{sec:conformal-prediction}.
    Coverage (\Cov), omitted here, is empirically close to $1 - \alpha$.
    \Thr clearly outperforms all approaches \wrt inefficiency.
    \textit{Right:} Inefficiency distribution across CIFAR10 classes (for $\alpha{=}0.01$) is plotted, with more difficult classes yielding higher inefficiency.
    }
    \vspace*{-0.1cm}
    \label{fig:baseline}
\end{figure}

The \textbf{threshold conformal predictor (\Thr)} \citep{SadinleJASA2019} constructs the confidence sets by thresholding probabilities:
$C_\theta(x;\tau) := \{k: \pi_{\theta,k}(x) =: E_\theta(x, k) \geq \tau\}$.
Here, the subscript $C_\theta$ makes the dependence on the model $\pi_\theta$ and its parameters $\theta$ explicit.
During calibration, $\tau$ is computed as the $\alpha (1 + \nicefrac{1}{|\Ical|})$-quantile of the so-called conformity scores $E_\theta(x_i, y_i) = \pi_{\theta,y_i}(x_i)$.
The conformity scores indicate, for each example, the threshold that ensures coverage.
Marginal coverage of $(1 - \alpha)$ is guaranteed on \red{a} test example $(X, Y)$.
In practice, \Thr can also be applied on logits (\ThrL) or log-probabilities (\ThrLP) instead of probabilities.

\textbf{Adaptive Prediction Sets (\APS)} \citep{RomanoNIPS2020} constructs confidence sets based on the ordered probabilities.
Specifically, $C_\theta(x;\tau) := \{k: E_\theta(x, k) \,\leq\, \tau\}$ with:
\begin{align}
    E_\theta(x, k) := \pi_{\theta,y^{(1)}}(x) + \ldots + \pi_{\theta,y^{(k - 1)}}(x) + U\pi_{\theta,y^{(k)}}(x)\label{eq:aps},
\end{align}
where $\pi_{\theta,y^{(1)}}(x) \geq \ldots \geq \pi_{\theta, y^{(K)}}(x)$ and $U$ is a uniform random variable in $[0,1]$ to break ties.
Similar to \Thr, the conformity scores $E_\theta(x_i, y_i)$ \wrt the true classes $y_i$ are used for calibration, but the $(1 - \alpha)(1 + \nicefrac{1}{|\Ical|})$-quantile is required to ensure marginal coverage on test examples.

Performance of \CP is then measured using two metrics: (empirical and marginal) \textbf{coverage (\Cov)} as well as \textbf{inefficiency (\Ineff)}.
Letting \Itest be a test set of size $|\Itest|$, these metrics are computed as
\begin{align}
    \Cov := \frac{1}{|\Itest|} \sum_{i \in \Itest} \delta[y_i \in C(x_i)]\quad\text{and}\quad \Ineff := \frac{1}{|\Itest|} \sum_{i \in \Itest} |C(x_i)|,\label{eq:coverage-inefficiency}
\end{align}
where $\delta$ denotes an indicator function that is $1$ when its argument is true and $0$ otherwise.
Due to the marginal coverage guarantee provided by \CP (\cf \citep{RomanoNIPS2020} or \appref{sec:app-guarantee}), 
the empirical coverage, when averaged across several calibration/test splits, is $\Cov \approx 1 - \alpha$.
Thus, we concentrate on inefficiency as the main metric to compare across \CP methods and models.
\red{With \emph{accuracy}, we refer to the (top-1) accuracy with respect to the $\argmax$-predictions, \ie, $\argmax_k \pi_{\theta,k}(x)$, obtained by the underlying model $\pi$.}
As shown in \figref{fig:baseline} (left), \Thr clearly outperforms \ThrL and \APS \wrt inefficiency (lower is better) averaged across random \Ical/\Itest splits (details in \secref{sec:app-experimental-setup}).

\CP is intended to be used as a ``wrapper'' around $\pi_\theta$.
``Better'' \CP methods generally result in lower inefficiency for a \emph{fixed} model $\pi_\theta$.
For example, following \figref{fig:baseline} (left), regularized \APS (\RAPS) \citep{AngelopoulosICLR2020} recently showed how to improve inefficiency compared to \APS by modifying the conformity score -- without outperforming \Thr, however.
Fine-grained control over inefficiency, \eg, conditioned on the class or the composition of the $C(X)$ is generally \emph{not} possible.
Integrating \CP into the training procedure promises a higher degree of control, however, requires differentiable \CP implementations, \eg, for \Thr or \APS.

\subsection{Differentiable Prediction and Calibration Steps}
\label{subsec:differentiable-conformal-predictors}

Differentiating through \CP involves differentiable prediction and calibration steps:
We want $C_\theta(x; \tau)$ to be differentiable \wrt the predictions $\pi_\theta(x)$, and $\tau$ to be differentiable \wrt to the predictions $\pi_\theta(x_i)$, $i \in \Ical$ used for calibration.
We emphasize that, ultimately, this allows to differentiate through both calibration and prediction \wrt the \emph{model parameters $\theta$}, on which the predictions $\pi_\theta(x)$ and thus the conformity scores $E_\theta(x,k)$ depend.
\red{For brevity, we focus on \Thr, see \algref{alg:ct} and discuss \APS in \appref{sec:app-differentiable-aps}.}

\textbf{Prediction} involves thresholding the conformity scores $E_\theta(x, k)$, which can be smoothed using the sigmoid function $\sigma(z) = \nicefrac{1}{1 + \exp(-z)}$ and a temperature hyper-parameter $T$:
$C_{\theta,k}(x;\tau) := \sigma\left(\nicefrac{(E_\theta(x, k) - \tau)}{T}\right)$.
Essentially, $C_{\theta,k}(x;\tau) \in [0,1]$ represents a \emph{soft} assignment of class $k$ to the confidence set, \ie, can be interpreted as the probability of $k$ being included.
For $T \rightarrow 0$, the ``hard'' confidence set will be recovered, \ie, $C_{\theta,k}(x;\tau) = 1$ for $k \in C_\theta(x;\tau)$ and $0$ otherwise.
\red{For \Thr, the conformity scores are naturally differentiable \wrt to the parameters $\theta$ because $E(x, k) = \pi_{\theta,k}(x)$.}

As the conformity scores are already differentiable, \textbf{calibration} 
\red{merely involves a differentiable quantile computation. This can be accomplished using any smooth sorting approach \citep{BlondelICML2020,CuturiNIPS2019,Williamson2020}.
These often come with a ``dispersion'' hyper-parameter $\epsilon$ such that smooth sorting approximates ``hard'' sorting for $\epsilon \rightarrow 0$.}
Overall, this results in the threshold $\tau$ being differentiable \wrt the predictions of the calibration examples $\{(\pi_\theta(x_i),y_i\}_{i \in \Ical}$ and the model's parameters $\theta$.

\red{As this approximation is using smooth operations, the coverage guarantee seems lost.
However, in the limit of $T,\epsilon \rightarrow 0$ we recover the original non-smooth computations and the corresponding coverage guarantee.
Thus, it is reasonable to assume that, in practice, we \emph{empirically} obtain coverage close to $(1 - \alpha)$.
We found that this is sufficient because these smooth variants are \emph{only} used during training.
At test time, we use the original (non-smooth) implementations and the coverage guarantee follows directly from \citep{RomanoNIPS2020,SadinleJASA2019}.}

\section{Conformal Training (\CT): \textit{Learning} Conformal Prediction}
\label{sec:conformal-training}

\begin{figure}
    \vspace*{-0.2cm}
    \centering
    \begin{minipage}[t]{0.52\textwidth}
        \vspace*{-5px}
        
        \begin{tcolorbox}[arc=0mm,colback=white,colframe=seaborn1,boxrule=1px,bottom=0.25mm,top=0.25mm,left=0.25mm,right=0.25mm]
            \begin{algorithmic}[1]
                \footnotesize 
                \Function{Predict}{$\pi_\theta(x)$, $\tau$}
                    \State compute $E_\theta(x, k)$, $k{\in}[K]$
                    \State \Return $C_\theta(x;\tau) = \{k: E_\theta(x, k) \geq \tau\}$
                \EndFunction
            \end{algorithmic}
        \end{tcolorbox}
        \vskip 3px
        \begin{tcolorbox}[arc=0mm,colback=white,colframe=seaborn1,boxrule=1px,bottom=0.25mm,top=0.25mm,left=0.25mm,right=0.25mm]
            \begin{algorithmic}[1]
                \footnotesize 
                \Function{Calibrate}{$\{(\pi_\theta(x_i), y_i\}_{i = 1}^n$, $\alpha$}
                    \State compute $E_\theta(x_i, y_i)$, $i{=}1,\ldots,n$
                    \State \Return \Call{Quantile}{$\{E_\theta(x_i, y_i)\}$, $\alpha(1 + \nicefrac{1}{n})$}
                \EndFunction
            \end{algorithmic}
        \end{tcolorbox}
        \vskip 3px
        \begin{tcolorbox}[arc=0mm,colback=white,colframe=seaborn2,boxrule=1px,bottom=0.25mm,top=0.25mm,left=-0.5mm,right=0.25mm]
            \begin{algorithmic}[1]
                \footnotesize 
                \Function{SmoothPred}{$\pi_\theta(x)$, $\tau$, $T{=}1$}
                    \State \Return $C_{\theta,k}(x;\tau){\,=\,}\sigma(\frac{(E_\theta(x,k) - \tau)}{T})$, $k{\,\in\,}[K]$
                \EndFunction
                \Function{SmoothCal}{$\{(\pi_\theta(x_i), y_i\}_{i = 1}^n$, $\alpha$}
                    \State \Return \Call{SmoothQuant}{$\{E_\theta(x_i,y_i)\}$, $\alpha(1{+}\frac{1}{n})$}
                \EndFunction
            \end{algorithmic}
        \end{tcolorbox}
    \end{minipage}
    \hfill
    \begin{minipage}[t]{0.47\textwidth}
        \vspace*{0px}
        
        \begin{tcolorbox}[arc=0mm,colback=white,colframe=seaborn2,boxrule=1px,bottom=0.25mm,top=0.25mm,left=0.25mm,right=0.25mm]
            \begin{algorithmic}[1]
                \footnotesize 
                \Function{ConformalTraining}{$\alpha$, $\lambda{=}1$}
                    \For{mini-batch $B$}
                        \State randomly split batch $\Bcal\uplus \Bpred = B$
                        \State \{``On-the-fly'' calibration on \Bcal:\}
                        \State $\tau{\,=\,}$\Call{SmoothCal}{$\{(\pi_\theta(x_i), y_i)\}_{i \in \Bcal}$,\,$\alpha$}
                        \State \{Prediction only on $i \in \Bpred$:\}
                        \State $C_\theta(x_i; \tau){\,=}$ \Call{SmoothPred}{$\pi_\theta(x_i)$,\,$\tau$}
                        \State \{\emph{Optional} classification loss:\}
                        \State $\mathcal{L}_B{\,=\,}0$ or $\sum_{i \in \Bpred} \mathcal{L}(C_\theta(x_i; \tau), y_i)$
                        \State $\Omega_B{\,=\,}\sum_{i \in \Bpred}\Omega(C_\theta(x_i;\tau))$
                        \State $\Delta{\,=\,}\nabla_\theta \nicefrac{1}{|\Bpred|}(\mathcal{L}_B + \lambda\Omega_B)$
                        \State update parameters $\theta$ using $\Delta$
                    \EndFor
                \EndFunction
            \end{algorithmic}
        \end{tcolorbox}
    \end{minipage}
    
    \vspace*{-6px}
    \captionof{algorithm}{
    \textbf{Smooth \CP and Conformal Training (\CT):}
    \textit{\color{seaborn1}\bfseries Top left:} At test time, \red{for \Thr}, \textsc{Predict} computes the conformity scores $E_\theta(x, k)$ for each $k{\in}[K]$ and constructs the confidence sets $C_\theta(x;\tau)$ by thresholding with $\tau$.
    \textsc{Calibrate} determines the threshold $\tau$ as the $\alpha (1 + \nicefrac{1}{n})$-quantile of the conformity scores \wrt the true classes $y_i$ on a calibration set $\{(x_i, y_i)\}$ of size $n{:=}|\Ical|$.
    \Thr and \APS use different conformity scores.
    \textit{\color{seaborn2}\bfseries Right and bottom left:}
    \CT calibrates on a part of each mini-batch, \Bcal.
    Thereby, we obtain guaranteed coverage on the other part, \Bpred (in expectation across batches).
    Then, the inefficiency on \Bpred is minimized to update the model parameters $\theta$.
    Smooth implementations of calibration and prediction are used.
    }
    \label{alg:ct}
    \vspace*{-0.1cm}
\end{figure}

The key idea of \textbf{conformal training (\CT)} is to ``simulate'' \CP during training, \ie, performing both calibration and prediction steps on each mini-batch.
This is accomplished using the differentiable \red{conformal predictors as introduced in \secref{subsec:differentiable-conformal-predictors}.}
\CT can be viewed as a generalization of \citep{BellottiARXIV2021} that just differentiates through the prediction step with a fixed threshold, without considering the crucial calibration step, see \appref{sec:app-coverage-training}.
\red{In both cases, \emph{only} the training procedure changes. After training, standard (non-smooth) conformal predictors are applied.}

\subsection{\CT by Optimizing Inefficiency}
\label{subsec:conformal-training}

\CT performs (differentiable) \CP on each mini-batch during stochastic gradient descent (SGD) training.
In particular, as illustrated in \figref{fig:introduction} \textcircled{a}, we split each mini-batch $B$ in half: the first half is used for calibration, \Bcal, and the second one for prediction and loss computation, \Bpred.
That is, on \Bcal, we calibrate $\tau$ by computing the $\alpha(1 + \nicefrac{1}{|\Bcal|})$-quantile of the conformity scores in a differentiable manner. 
It is important to note that we compute $C_\theta(x_i;\tau)$ only for $i \in \Bpred$ and \emph{not} for $i \in \Bcal$.
Then, in expectation across mini-batches and large enough $|\Bcal|$, \red{for $T,\epsilon \rightarrow 0$}, \CP  guarantees coverage $1 - \alpha$ on \Bpred.
\red{Assuming empirical coverage to be close to $(1 - \alpha)$ in practice,} we only need to minimize inefficiency during training:
\begin{align}
    \min_\theta \log\mathbb{E}\left[\Omega(C_\theta(X;\tau))\right]\quad\text{with }\Omega(C_\theta(x;\tau)) = \max\left(0, \sum_{k = 1}^K C_{\theta,k}(x;\tau) - \kappa\right).\label{eq:size-loss}
\end{align}
We emphasize that \CT optimizes the model parameters $\theta$ on which the confidence sets $C_\theta$ depend through the model predictions $\pi_\theta$.
Here, $\Omega$ is a ``smooth'' \emph{size loss} intended to minimize the expected inefficiency, \ie, $\mathbb{E}[|C_\theta(X;\tau)|]$, not to be confused with the statistic in \eqnref{eq:coverage-inefficiency} used for evaluation.
Remember that $C_{\pi,k}(x;\tau)$ can be understood as a soft assignment of class $k$ to the confidence set $C_\theta(x;\tau)$.
By default, we use $\kappa = 1$ in order to not penalize singletons.
However, $\kappa \in \{0,1\}$ can generally be treated as hyper-parameter.
After training, any \CP method can be applied to \red{re-}calibrate $\tau$ on a held-out calibration set \Ical as usual, \ie, the thresholds $\tau$ obtained during training are \emph{not} kept. This \red{ensures that we obtain} a coverage guarantee of \CP.

\subsection{\CT with Classification Loss}

In order to obtain more control over the composition of confidence sets $C_\theta(X;\tau)$ at test time, \CT can be complemented using a generic loss $\mathcal{L}$:
\begin{align}
    \min_\theta \log\left(\mathbb{E}\left[\mathcal{L}(C_\theta(X;\tau), Y) + \lambda \Omega(C_\theta(X;\tau))\right]\right).\label{eq:conformal-training-classification}
\end{align}
While $\mathcal{L}$ can be any arbitrary loss defined directly on the confidence sets $C_\theta$, we propose to use a ``configurable'' \emph{classification loss} \Lclass.
This classification loss is intended to explicitly enforce coverage, \ie, make sure the true label $Y$ is included in $C_\theta(X;\tau)$, and optionally penalize other classes $k$ \emph{not} to be included in $C_\theta$, as illustrated in \figref{fig:introduction} \textcircled{b}.
To this end, we define
\begin{align}
    \Lclass(C_\theta(x;\tau), y) := \sum_{k = 1}^K L_{y,k} \Big[\underbrace{(1 - C_{\theta,k}(x;\tau)) \cdot \delta[y=k]}_{\text{enforce }y\text{ to be in }C} + \underbrace{C_{\theta,k}(x;\tau)\cdot \delta[y\neq k]}_{\text{penalize class }k\neq y\text{ \emph{not} to be in }C}\Big].\label{eq:classification-loss}
\end{align}
As above, $C_{\theta,k}(x;\tau) \in [0,1]$ such that $1 - C_\theta(x;\tau)$ can be understood as the likelihood of $k$ not being in $C_\theta(x;\tau)$.
In \eqnref{eq:classification-loss}, the first term is used to encourage coverage, while the second term can be used to avoid predicting other classes.
This is governed by the \emph{loss matrix} $L$:
For $L = I_K$, \ie, the identity matrix with $K$ rows and columns, this loss simply enforces coverage (perfect coverage if $\Lclass = 0$).
However, setting any $L_{y,k} > 0$ for $y \neq k$ penalizes the model from including class $k$ in confidence sets with ground truth $y$.
Thus, cleverly defining $L$ allows to define rather complex objectives, as we will explore next.
\CT with (optional) classification loss is summarized in \algref{alg:ct} (right) and Python code can be found in \appref{sec:app-code}.

\subsection{\CT with General and Application-Specific Losses}
\label{subsec:conformal-training-applications}

We consider several use cases motivated by medical diagnosis, \eg, breast cancer screening \citep{McKinneyNATURE2020} or classification of dermatological conditions \citep{LiuNATURE2020,RoyARXIV2021,JainJAMA2021}.
In skin condition classification, for example, predicting sets of classes, \eg, the top-$k$ conditions, is already a common strategy for handling uncertainty.
In these cases, we not only care about coverage guarantees but also desirable characteristics of the confidence sets.
These constraints in terms of the predicted confidence sets can, however, be rather complicated and pose difficulties for standard \CP.
We explore several exemplary use cases to demonstrate the applicability of \CT, that are also relevant beyond the considered use cases in medical diagnosis.

First, we consider ``shaping'' class-conditional inefficiency, formally defined as
\begin{align}
    \Ineff[Y = y] := \frac{1}{\sum_{i \in \Itest}\delta[y_i = y]} \sum_{i \in \Itest} \delta[y_i = y] |C(x_i)|.\label{eq:conditional-ineff}
\end{align}
Similarly, we can define inefficiency conditional on a \emph{group} of classes.
For example, we could reduce inefficiency, \ie, uncertainty, on ``low-risk'' diseases at the expense of higher uncertainty on ``high-risk'' conditions.
This can be thought of as re-allocating time spent by a doctor towards high-risk cases.
Using \CT, we can manipulate group- or class-conditional inefficiency using a weighted size loss $\omega \cdot \Omega(C(X;\tau))$ with $\omega := \omega(Y)$ depending on the ground truth $Y$ in \eqnref{eq:size-loss}.

Next, we consider \emph{which} classes are actually included in the confidence sets.
\CP itself does not enforce any constraints on the composition of the confidence sets.
However, with \CT, we can penalize the ``confusion'' between pairs of classes:
for example if two diseases are frequently confused by doctors, it makes sense to train models that avoid confidence sets that contain \emph{both} diseases.
To control such cases, we define the \emph{coverage confusion matrix} as
\begin{align}
    \Sigma_{y,k} := \frac{1}{|\Itest|} \sum_{i \in \Itest} \delta[y_i = y \wedge k \in C(x_i)].\label{eq:coverage-confusion}
\end{align}
The off-diagonals, \ie, $\Sigma_{y,k}$ for $y \neq k$, quantify how often class $k$ is included in confidence sets with true class $y$.
Reducing $\Sigma_{y,k}$ can be accomplished using a positive entry $L_{y,k} > 0$ in \eqnref{eq:classification-loss}.

Finally, we explicitly want to penalize ``overlap'' between groups of classes in confidence sets.
For example, we may \emph{not} want to concurrently include very high-risk conditions among low-risk ones in confidence sets, to avoid unwanted anxiety or tests for the patient.
Letting $K_0 \uplus K_1$ being two disjoint sets of classes, we define \emph{mis-coverage} as
\begin{align}
    \MisCov_{0\rightarrow1} = \frac{1}{\sum_{i \in \Itest} \delta[y_i \in K_0]}\sum_{i \in \Itest} \delta[y_i \in K_0 \wedge \left(\exists k \in K_1: k \in C(x_i)\right)].\label{eq:mis-coverage}
\end{align}
Reducing $\MisCov_{0\rightarrow1}$ means avoiding classes $K_1$ being included in confidence sets of classes $K_0$.
Again, we use $L_{y,k} > 0$ for $y \in K_0, k \in K_1$ to approach this problem.
$\MisCov_{1\rightarrow0}$ is defined analogously and measures the opposite, \ie, classes $K_0$ being included in confidence sets of $K_1$.
\section{Experiments}
\label{sec:experiments}

We present experiments in two parts:
First, in \secref{subsec:experiments-inefficiency}, we demonstrate that \CT can reduce inefficiency of \Thr and \APS compared to \CP applied to a baseline model trained using cross-entropy loss separately \red{(see \tabref{tab:results} for the main results)}.
Thereby, we outperform concurrent work of \cite{BellottiARXIV2021}.
Second, in \secref{subsec:experiments-shaping}, we show how \CT can be used to ``shape'' confidence sets, \ie, reduce class-conditional inefficiency for specific (groups of) classes or coverage confusion of two or more classes, while maintaining the marginal coverage guarantee.
This is impossible using \citep{BellottiARXIV2021} and rather difficult for standard \CP.

We consider several benchmark datasets as well as architectures, \cf \tabref{tab:app-datasets}, and report metrics averaged across $10$ random calibration/test splits for $10$ trained models for each method.
We focus on \red{(non-differentiable)} \Thr and \APS as \CP methods used \emph{after} training \red{and, thus, obtain the corresponding coverage guarantee.}
\Thr, in particular, consistently achieves lower inefficiency for a fixed confidence level $\alpha$ than, \eg, \ThrL (\ie, \Thr on logits) or \RAPS, see \figref{fig:baseline} (left).
We set $\alpha = 0.01$ and use the same $\alpha$ during training using \CT.
Hyper-parameters are optimized for \Thr or \APS individually.
We refer to \appref{sec:app-experimental-setup} for further details on datasets, models, evaluation protocol and hyper-parameter optimization.

\subsection{Reducing Inefficiency with \CT}
\label{subsec:experiments-inefficiency}

In the first part, we focus on the inefficiency reductions of \CT
in comparison to \red{a standard cross-entropy training baseline} and \citep{BellottiARXIV2021} (\Bel). After summarizing the possible inefficiency reductions, we also discuss which \CP method to use during training and how \CT can be used for ensembles and generalizes to lower $\alpha$.

\begin{table}[t]
    \vspace*{-0.2cm}
    \caption{
    \textbf{Main Inefficiency Results}, comparing \citep{BellottiARXIV2021} (\Bel, trained with \ThrL) and \CT (trained with \ThrLP)
    using \Thr or \APS at test time (with $\alpha{=}0.01$).
    \red{We also report improvements relative to the baseline, \ie, standard cross-entropy training, in percentage in parentheses.}
    \CT results in a consistent improvement of inefficiency for both \Thr and \APS.
    Training with \Lclass, \red{using $L = I_K$}, generally works slightly better.
    On CIFAR, the inefficiency reduction is smaller compared to other datasets as \CT is trained on pre-trained ResNet features, see text.
    More results can be found in \appref{sec:app-results}.
    }
    \label{tab:results}
    \vspace*{-6px}
    \centering
    \small
    \hspace*{-0.7cm}
    \begin{minipage}{0.625\textwidth}
    \begin{tabular}{|l||c|c|c|c||c|c|c|}
        \hline
        \multicolumn{8}{|c|}{\textbf{Inefficiency $\downarrow$}, \CT (trained w/ \ThrLP), $\alpha = 0.01$}\\
        \hline
        & \multicolumn{4}{c||}{\Thr}
        & \multicolumn{3}{c|}{\APS}\\
        \hline
        Dataset & Basel. & \Bel & \CT & +\Lclass
        & Basel. & \CT & +\Lclass\\
        \hline\hline
        MNIST & 2.23 & 2.70 & 2.18 & \textbf{2.11} (-5.4\%) & 2.50 & 2.16 & \textbf{2.14} (-14.4\%)\\
        \hline
        F-MNIST & 2.05 & 1.90 & 1.69 & \textbf{1.67} (-18.5\%) & 2.36 & 1.82 & \textbf{1.72} (-27.1\%)\\
        \hline
        EMNIST & 2.66 & 3.48 & 2.66 & \textbf{2.49} (-6.4\%) & 4.23 & \textbf{2.86} & 2.87 (-32.2\%)\\
        %$\alpha{=}0.001$ & 15.73 & 19.33 && 13.65 &&&\\
        \hline
        CIFAR10 & 2.93 & 2.93 & 2.88 & \textbf{2.84} (-3.1\%) & 3.30 & 3.05 & \textbf{2.93} (-11.2\%)\\
        \hline
        CIFAR100 & 10.63 & 10.91 & 10.78 & \textbf{10.44} (-1.8\%) & 16.62 & 12.99 & \textbf{12.73} (-23.4\%)\\
        \hline
    \end{tabular}
    \end{minipage}
    \vspace*{-0.1cm}
\end{table}
\begin{table}[b]
    \vspace*{-0.1cm}
    \caption{
    \textbf{Ensemble Results and Lower Confidence Levels $\alpha$:}
    \textit{Left}: ``Conformalization'' of ensembles using a 2-layer MLP trained on logits, either normally or using \CT.
    The ensemble contains $18$ models with accuracies in between $75.10$ and $82.72\%$.
    Training a model on top of the ensemble clearly outperforms the best model of the ensemble; using \CT further boosts \Ineff.
    \textit{Right}: The inefficiency improvements of \tabref{tab:results} generalize to lower confidence levels $\alpha$ on EMNIST, although \CT is trained with $\alpha{=}0.01$.
    }
    \label{tab:further-results}
    \vspace*{-6px}
    \centering
    \small
    \begin{tabular}{|l|c|c|c|}
        \hline
        \multicolumn{4}{|c|}{\textbf{\red{CIFAR10}:} Ensemble Results}\\
        \hline
        Test
        & \multicolumn{3}{|c|}{\Thr}\\
        \hline
        Method
        & (Models)
        & +MLP
        & +\CT\\
        \hline
        \hline
        Avg. \Ineff & 3.10 & 2.40 & \bfseries 2.35\\
        Best \Ineff & 2.84 & 2.33 & \bfseries 2.30\\
        \hline
    \end{tabular}
    \begin{tabular}{|l|c|c|}
        \hline
        \multicolumn{3}{|c|}{\textbf{EMNIST:} Confidence Levels}\\
        \hline
        \hline
        Method
        & Basel.
        & \CT\\
        \hline
        Test & \multicolumn{2}{c|}{\Thr}\\
        \hline
        \hline
        %\Ineff, $\alpha{=}0.01$ & 2.66 & 2.66\\
        \Ineff, $\alpha{=}0.005$ & 4.10 & \textbf{3.37} (-17.8\%)\\
        \Ineff, $\alpha{=}0.001$ & 15.73 & \textbf{13.65} (-13.2\%)\\
        \hline
    \end{tabular}
    \vspace*{-0.2cm}
\end{table}

\textbf{Main Results:}
In \tabref{tab:results}, we summarize the inefficiency reductions possible through \CT (trained with \ThrLP) in comparison to \Bel (trained with \ThrL) and the baseline.
\Bel does \emph{not} consistently improve inefficiency on all datasets.
Specifically, on MNIST, EMNIST or CIFAR100, inefficiency actually \emph{worsens}.
Our \CT, in contrast, reduces inefficiency consistently, not only for \Thr but also for \APS.
Here, improvements on CIFAR for \Thr are generally less pronounced.
This is likely because we train linear models on top of a pre-trained ResNet \citep{HeCVPR2016} where features are not taking into account conformalization at test time, see \appref{sec:app-results}.
For \APS, in contrast, improvements are still significant.
Across all datasets, training with \Lclass generally performs slightly better, especially for datasets with many classes such as EMNIST ($K{=}52$) or CIFAR100 ($K{=}100$).
Overall, \CT yields significant inefficiency reductions, independent of the \CP method used at test time.

\textbf{Conformal Predictors for Training:}
In \tabref{tab:results}, we use \ThrLP during training, irrespective of the \CP method used at test time.
This is counter-intuitive when using, \eg, \APS at test time.
However, training with \Thr and \APS is rather difficult, as discussed in \appref{sec:app-ablation}.
This is likely caused by limited gradient flow as both \Thr and \APS are defined on the predicted probabilities instead of log-probabilities as used for \ThrLP or in cross-entropy training.
\red{Moreover, re-formulating the conformity scores of \APS in \eqnref{eq:aps} to use log-probabilities is non-trivial.}
\red{In contrast, \Bel has to be trained using \ThrL as a fixed threshold $\tau$ is used during training.
This is because the calibration step is ignored during training.
Also,
fixing $\tau$ is not straightforward for \Thr due to the limited range of the predicted probabilities $\pi_{\theta,k}(x) \in [0,1]$, see \appref{sec:app-coverage-training}.}
We believe that this contributes to the poor performance of \Bel on several datasets.
Finally, we found that \Bel or \CT do not necessarily recover the accuracy of the baseline.
\red{Remember that we refer to the accuracy in terms of the $\argmax$-prediction of $\pi_\theta$.}
When training from scratch, accuracy can be 2-6\% lower while still \emph{reducing} inefficiency.
\red{This is interesting because \CT is still able to improve inefficiency, highlighting that cross-entropy training is not appropriate for \CP.}

\textbf{Further Results:}
\tabref{tab:further-results} includes additional results for \CT to ``conformalize'' ensembles on CIFAR10 (left) and with lower confidence levels $\alpha$ on EMNIST (right).
In the first example, we consider applying \CP to an ensemble of models.
Ensemble \CP methods such as \citep{YangARXIV2021} cannot improve \Ineff over the best model of the ensemble, \ie, 3.10 for \Thr.
Instead, training an MLP on top of the ensemble's logits can improve \Ineff to 2.40 and additionally using \CT to 2.35.
The second example shows that \CT, trained for $\alpha{=}0.01$, generalizes very well to significantly smaller confidence levels, \eg, $\alpha{=}0.001$ on EMNIST.
In fact, the improvement of \CT (without \Lclass) in terms of inefficiency is actually more significant for lower confidence levels.
We also found \CT to be very stable regarding hyper-parameters, see \appref{sec:app-parameters}.
Only too small batch sizes (\eg, $|B|{=}100$ on MNIST) prevents convergence.
This is likely because of too few examples ($|\Bcal|{=}50$) for calibration with $\alpha{=}0.01$ during training.
More results, \eg, on binary datasets or including additional hyper-parameter ablation can be found in \appref{sec:app-results}.

\begin{figure}[t]
    \vspace*{-0.2cm}
    \centering
    \hspace*{-0.3cm}
    \includegraphics[height=2.45cm]{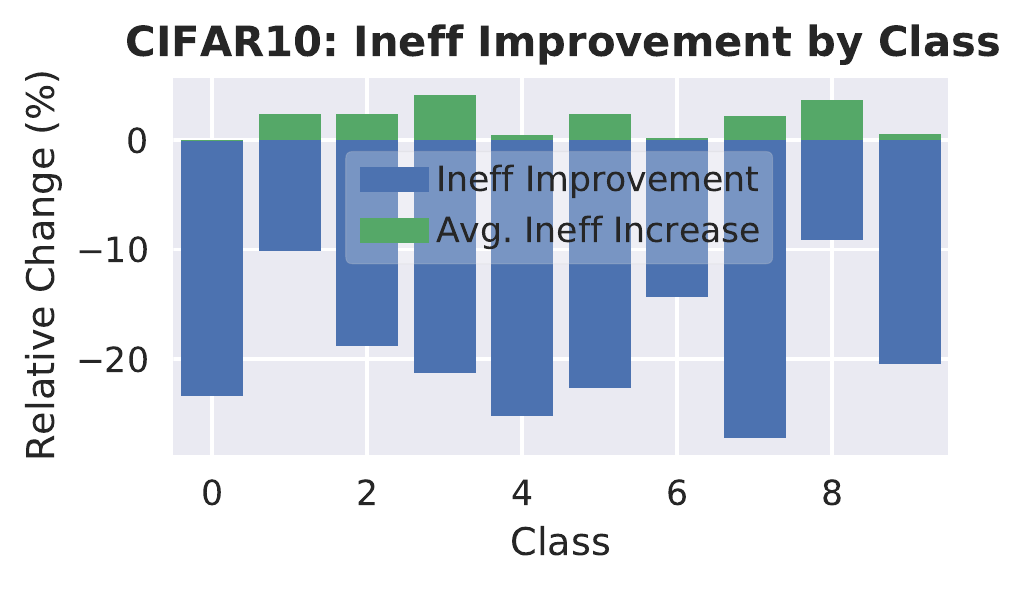}
    \hspace*{-0.1cm}
    \includegraphics[height=2.45cm]{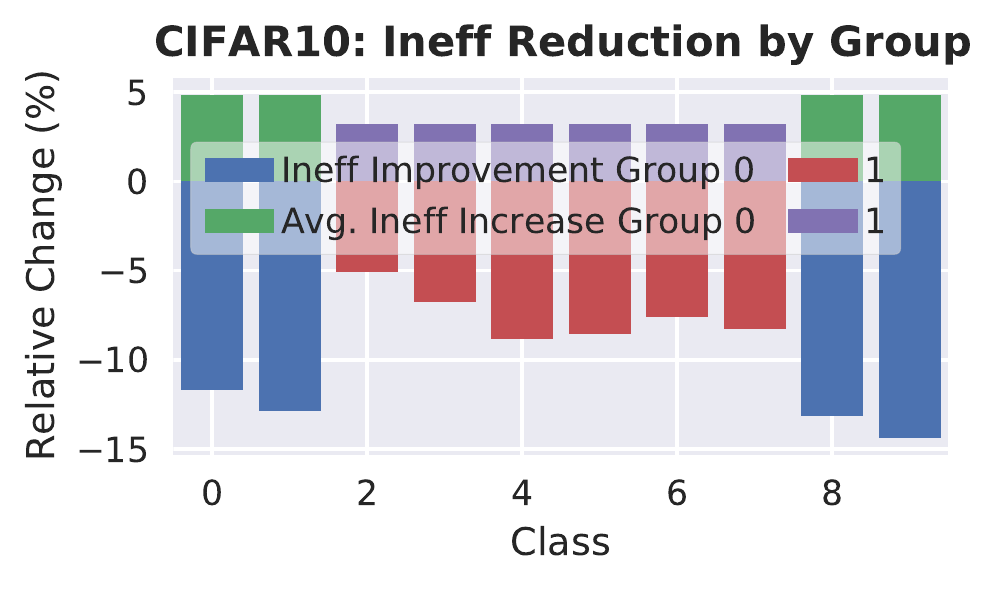}
    \hspace*{-0.1cm}
    \includegraphics[height=2.45cm]{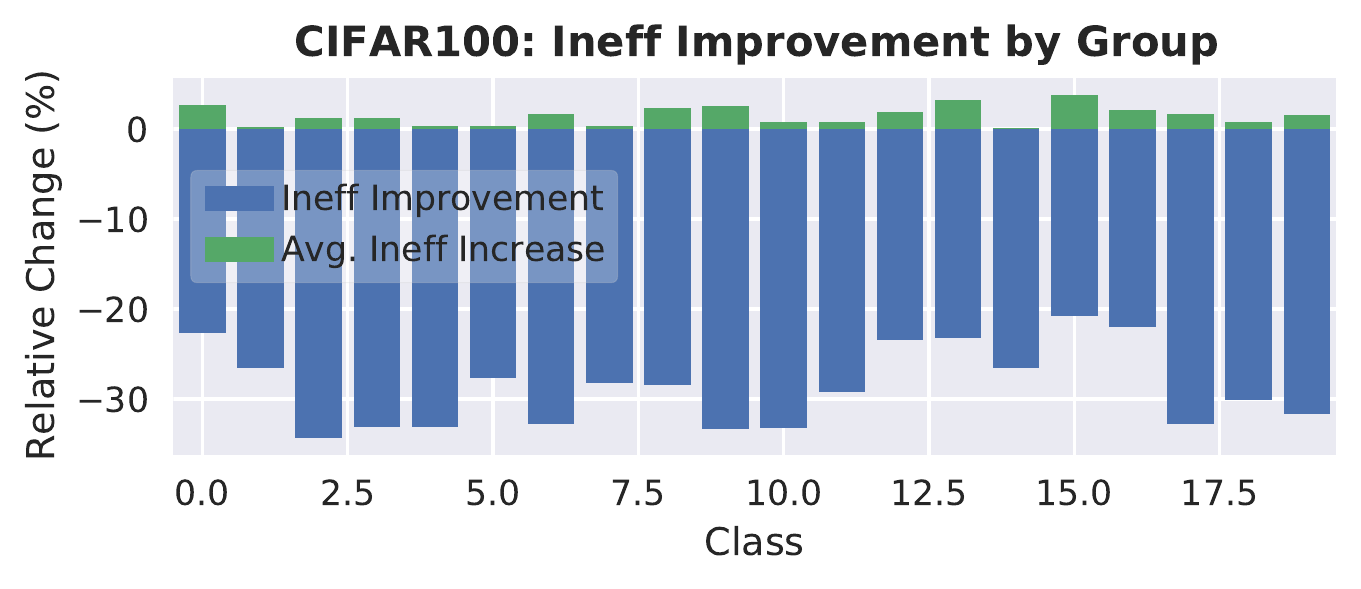}
    \vspace*{-22px}
    \caption{
    \textbf{Shaping Class-Conditional Inefficiency on CIFAR:}
    Possible inefficiency reductions, in percentage change, per class ({\color{seaborn1}blue}) and the impact on the overall, average inefficiency across classes ({\color{seaborn2}green}).
    \textit{Left:} Significant inefficiency reductions are possible for all classes on CIFAR10.
    \textit{Middle:} The same strategy applies to groups of classes, \eg, ``vehicles'' vs ``animals'', as well.
    \textit{Right:} Similarly, on CIFAR100, we group classes by their coarse class ($20$ groups à $5$ classes), see \citep{Krizhevsky2009}, allowing inefficiency improvements of more than 30\% per individual group.
    }
    \label{fig:size}
    \vspace*{-0.1cm}
\end{figure}
\begin{figure}[b]
    \vspace*{-0.1cm}
    \centering
    \begin{minipage}[t]{0.32\textwidth}
        \vspace*{0px}
        
        \includegraphics[height=2.7cm]{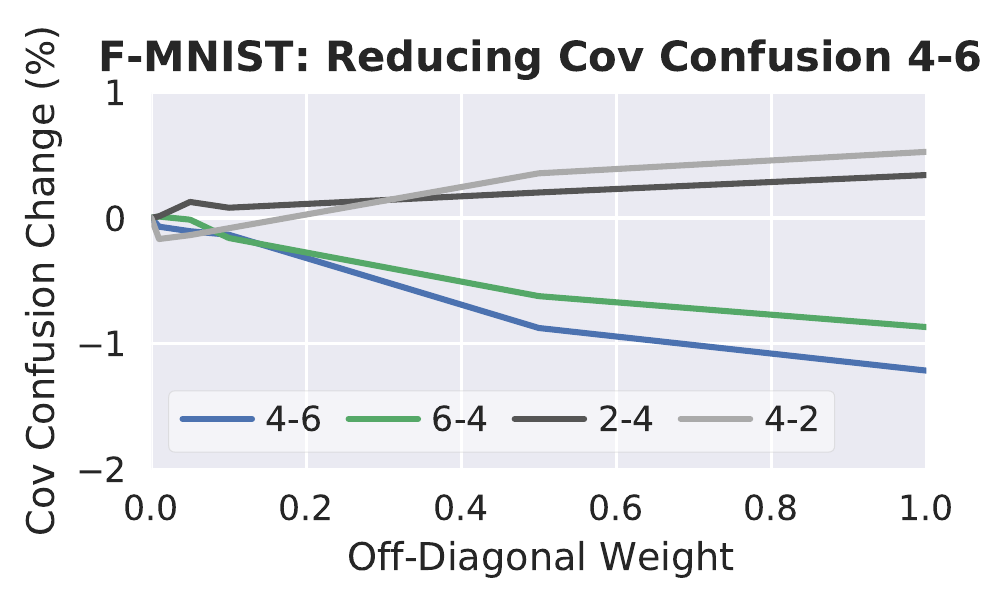}
    \end{minipage}
    \begin{minipage}[t]{0.32\textwidth}
        \vspace*{0px}
        
        \includegraphics[height=2.7cm]{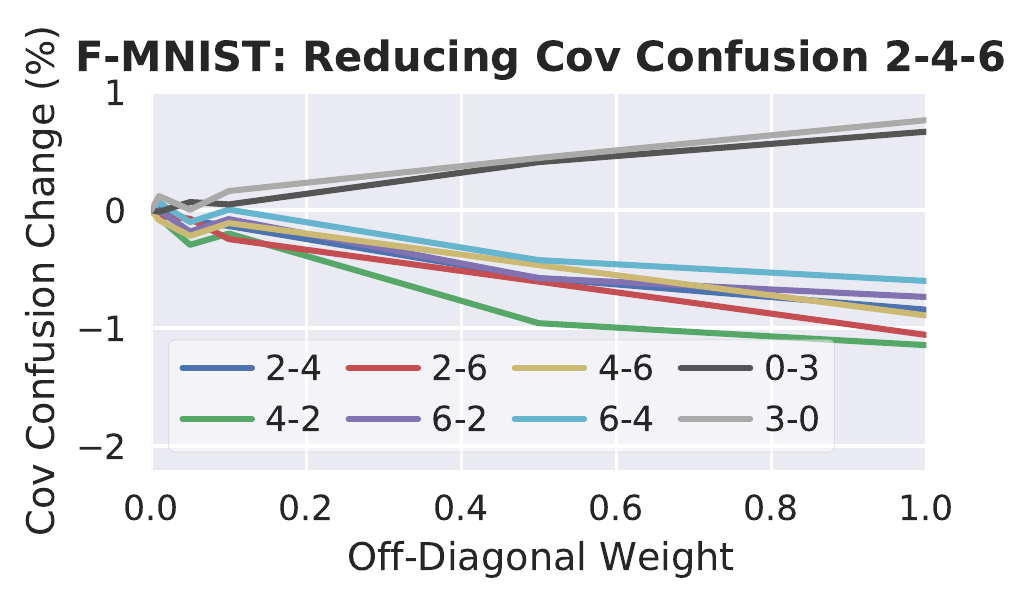}
    \end{minipage}
    \begin{minipage}[t]{0.33\textwidth}
        \vspace*{0px}
        
        \includegraphics[height=2.7cm]{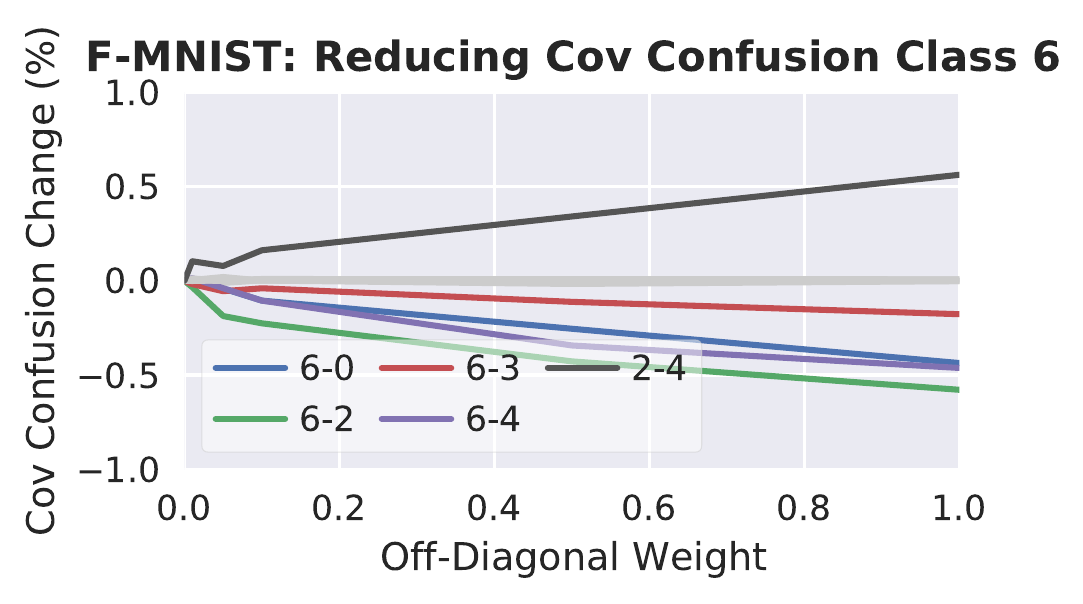}
    \end{minipage}
    \vspace*{-10px}
    \caption{
    \textbf{Controlling Coverage Confusion:}
    Controlling coverage confusion using \CT with \Lclass and an increasing penalty $L_{y,k}{\,>\,}0$ on Fashion-MNIST.
    For classes 4 and 6 (``coat'' and ``shirt''), coverage confusion $\Sigma_{y,k}$ and $\Sigma_{k,y}$ decreases significantly ({\color{seaborn1}blue} and {\color{seaborn2}green}).
    However, confusion of class 4 with class 2 (``pullover'') might increase ({\color{gray}gray}).
    \CT can also reduce coverage confusion of multiple pairs of classes (\eg, additionally considering class 2).
    Instead, we can also penalize confusion for each pair $(y, k)$, $k \in [K]$, \eg, $y{\,=\,}6$.
    Here, $L_{y,k}{\,>\,}0$, but $L_{y,k}{\,=\,}0$, \ie, \Cov confusion is not reduced symmetrically.
    }
    \label{fig:confusion}
    \vspace*{-0.2cm}
\end{figure}

\subsection{Conformal Training for Applications: Case Studies}
\label{subsec:experiments-shaping}

For the second part, we focus on \CT trained with \ThrLP and evaluated using \Thr.
We follow \secref{subsec:conformal-training-applications} and start by reducing class- or group-conditional inefficiency using \CT (without \Lclass), before demonstrating reductions in coverage confusion of two or more classes and avoiding mis-coverage between groups of classes (with \Lclass).
Because this level of control over the confidence sets is not easily possible using \Bel or standard \CP, we concentrate on \CT only:

\textbf{Shaping Conditional Inefficiency:}
We use \CT to reduce class-conditional inefficiency for specific classes or a group of classes, as defined in \eqnref{eq:conditional-ineff}.
In \figref{fig:baseline}, inefficiency is shown to vary widely across classes:
On CIFAR10, the more difficult class 3 (``cat'') obtains higher inefficiency than the easier class 1 (``automobile'').
Thus, in \figref{fig:size}, we use $\omega{=}10$ as described in \secref{subsec:conformal-training-applications} to reduce class- or group-conditional inefficiency.
We report the \emph{relative} change in percentage, showing that inefficiency reductions of 20\% or more are possible for many classes, including ``cat'' on CIFAR10 (left, {\color{seaborn1}blue}).
This is also possible for two groups of classes, ``vehicles'' vs. ``animals'' (middle).
However, these reductions usually come at the cost of a slight increase in average inefficiency across all classes ({\color{seaborn2}green}).
On CIFAR100, we consider 20 coarse classes, each containing 5 of the 100 classes (right).
Again, significant inefficiency reductions per coarse class are possible.
These observations generalize to all other considered datasets and different class groups, see \appref{sec:app-ineff}.

\begin{figure}[t]
    \vspace*{-0.2cm}
    \centering
    \begin{minipage}[t]{0.52\textwidth}
        \vspace*{0px}
        
        \small
        \begin{tabular}[t]{|l|c|c|c||c|c|c|}
            \hline
            \multicolumn{7}{|c|}{CIFAR10: $K_0{=}$ 3 (``cat'') vs. $K_1{=}$ Others}\\
            \multicolumn{7}{|c|}{CIFAR100: $K_0{=}$ ``human-made vs. $K_1{=}$ ``natural''}\\
            \hline
            \hline
            & \multicolumn{3}{c||}{CIFAR10}
            & \multicolumn{3}{c|}{CIFAR100}\\
            \hline
            && \multicolumn{2}{c||}{\MisCov $\downarrow$}
            && \multicolumn{2}{c|}{\MisCov $\downarrow$}\\
            \hline
            Method & \Ineff & $0{\rightarrow}1$ & $1{\rightarrow}0$
            & \Ineff & $0{\rightarrow}1$ & $1{\rightarrow}0$\\
            \hline
            \hline
            \CT & \bfseries 2.84 & 98.92 & 36.52 & \bfseries 10.44 & 40.09 & 29.6\\
            $L_{K_0, K_1}{=}1$ & 2.89 & \bfseries 91.60 & 34.74 & 16.50 & \bfseries 15.77 & 70.26\\
            $L_{K_1, K_0}{=}1$ & 2.92 & 97.36 & \bfseries 26.43 & 11.35 & 45.37 & \bfseries 17.56\\
            \hline
        \end{tabular}
    \end{minipage}
    \begin{minipage}[t]{0.01\textwidth}
        \vspace*{0px}
        
        {\color{black!50!white}\rule{0.5px}{3cm}}
    \end{minipage}
    \begin{minipage}[t]{0.3\textwidth}
        \vspace*{-6px}
    
        \includegraphics[height=1.8cm]{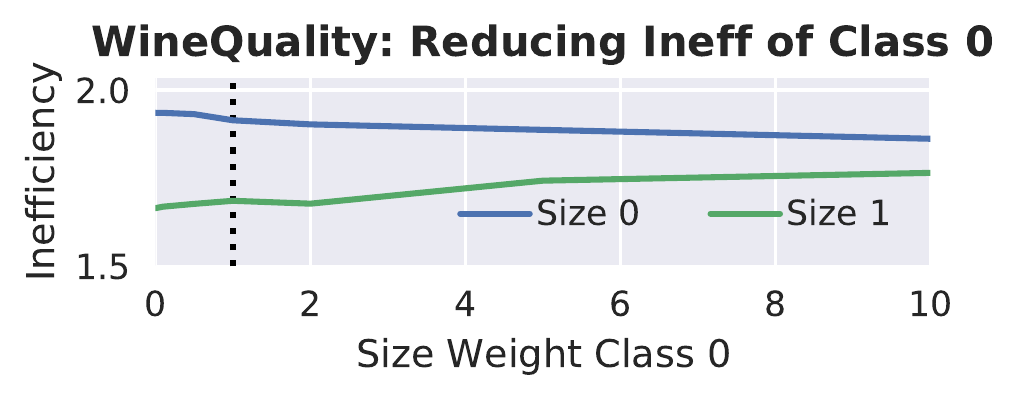}
        
        \includegraphics[height=1.8cm]{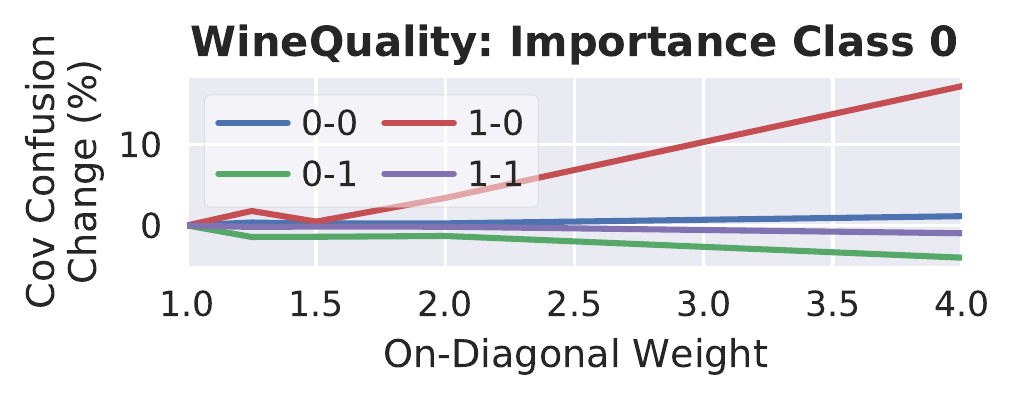}
    \end{minipage}
    \vspace*{-10px}
    \caption{
    \textbf{\textit{Left:} Reducing Mis-Coverage:}
    Following \secref{subsec:conformal-training-applications}, \CT allows to reduce mis-coverage on CIFAR.
    We consider $K_0{=}\{3\}$ (\ie, ``cat'') vs. all other classes on CIFAR10 (left) and ``human-made'' vs. ``natural'' on CIFAR100 ($|K_0|{=}35$, $|K_1|{=}65$, right).
    On CIFAR10, both $\MisCov_{0\rightarrow1}$ and $\MisCov_{1\rightarrow1}$ can be reduced significantly without large impact on inefficiency.
    For CIFAR100, in contrast, \Ineff increases more significantly.
    \textbf{\textit{Right:} Binary Class-Conditional Inefficiency and Coverage:}
    We plot inefficiency by class (top) and coverage confusion (bottom) on WineQuality.
    We can reduce inefficiency for class 0 (``bad''), the minority class, at the expense of higher inefficiency for class 1 (``good'') and boost class-conditional coverage for class 0.
    }
    \label{fig:miscoverage}
    \vspace*{-0.1cm}
\end{figure}

\textbf{Avoiding Coverage Confusion:}
Next, we use \CT to manipulate the coverage confusion matrix as defined in \eqnref{eq:coverage-confusion}.
Specifically, we intend to reduce coverage confusion of selected sets of classes.
Using a non-zero entry $L_{y,k}{\,>\,}0$, $y{\,\neq\,}k$ in \Lclass, as described in \secref{subsec:conformal-training-applications}, \figref{fig:confusion} (left) shows that coverage confusion can be reduced significantly for large enough $L_{y,k}$ on Fashion-MNIST:
Considering classes 4 and 6 (``coat'' and ``shirt'') confusion can be reduced by roughly $1\%$.
However, as accuracy stays roughly the same and coverage is guaranteed, this comes at the cost of increasing coverage confusion for other class pairs, \eg, 2 (``pullover'') and 4.
\CT can also be used to reduce coverage confusion of multiple class pairs (middle) or a whole row in the coverage confusion matrix $\Sigma_{y,k}$ with fixed $y$ and $y{\neq}k{\in}[K]$.
\figref{fig:confusion} (right) shows the results for class 6:
coverage confusion with, \eg, classes 0 (``t-shirt''), 2 or 4 ({\color{seaborn1}blue}, {\color{seaborn2}green} and {\color{seaborn4}violet}) is reduced roughly $0.5\%$ each at the cost of increased confusion of classes 2 and 4 (in {\color{gray}gray}).
These experiments can be reproduced on other datasets, \eg, MNIST or CIFAR10 in \appref{sec:app-confusion}.

\textbf{Reducing Mis-Coverage:}
We can also address unwanted ``overlap'' of two groups of classes using \CT and \Lclass.
In \figref{fig:miscoverage} (left) we explicitly measure mis-coverage as defined in \eqnref{eq:mis-coverage}.
First, on CIFAR10, we consider a singleton group $K_0{=}\{3\}$ (``cat'') and $K_1{=}[K]\setminus\{3\}$:
The \CT baseline $\MisCov_{0\rightarrow1}$ tells us that 98.92\% of confidence sets with true class $3$ also contain other classes.
Given an average inefficiency of $2.84$ this is reasonable.
Using $L_{3,k} = 1$, $k \neq 3$, this can be reduced to 91.6\%.
Vice-versa, the fraction of confidence sets of class $y{\neq}3$ containing class $3$ can be reduced from 36.52\% to 26.43\%.
On CIFAR100, this also allows to reduce overlap between ``human-made'' (35 classes) and ``natural'' (65 classes) things, \eg, $\MisCov_{0\rightarrow1}$ reduces from 40.09\% to 15.77\%, at the cost of a slight increase in inefficiency.
See \appref{sec:app-miscoverage} for additional results.

\textbf{Binary Datasets:}
Finally, in \figref{fig:miscoverage} (right), we illustrate that the above conclusions generalize to the binary case:
On WineQuality, we can control inefficiency of class 0 (``bad wine'', minority class with $\sim$37\% of examples) at the expense of increased inefficiency for class 1 (``good wine'', top).
Similarly, we can (empirically) improve class-conditional coverage for class 0 (bottom) or manipulate coverage confusion of both classes, see \appref{sec:app-binary}.
\section{Conclusion}
\label{sec:conclusion}

We introduced \textbf{conformal training (\CT)}, a novel method to train conformal predictors \emph{end-to-end} with the underlying model.
This addresses a major limitation of conformal prediction (\CP) in practice:
The model is fixed, leaving \CP little to no control over the predicted confidence sets.
In thorough experiments, we demonstrated that \CT can improve inefficiency of state-of-the-art \CP methods such as \Thr \citep{SadinleJASA2019} or \APS \citep{RomanoNIPS2020}.
More importantly, motivated by medical diagnosis, we highlighted the ability of \CT to manipulate the predicted confidence sets in various ways.
First, \CT can ``shape'' the class-conditional inefficiency distribution, \ie, reduce inefficiency on specific classes at the cost of higher inefficiency for others.
Second, \CT allows to control the coverage-confusion matrix by, \eg, reducing the probability of including classes other than the ground truth in confidence sets.
Finally, this can be extended to explicitly reduce ``overlap'' between groups of classes in the predicted confidence sets.
In all cases, \CT does \emph{not} \red{lose} the (marginal) coverage guarantee provided by \CP.
\clearpage
\section*{Ethics Statement}

Recent deep learning based classifiers, as used in many high-stakes applications, achieve impressive accuracies on held-out test examples.
However, this does \emph{not} provide sufficient guarantees for safe deployment. 
Conformal prediction (\CP), instead, predicts \emph{confidence sets} equipped with a guarantee that the true class is included with specific, user-specified probability.
These confidence sets also provide intuitive uncertainty estimates.
We specifically expect \CP to be beneficial in the medical domain, improving trustworthiness among doctors and patients alike by providing performance guarantees and reliable uncertainty estimates.
Yet, the current work does \emph{not} contain experiments with personal/sensitive medical data.
The presented results are on standard benchmark datasets only.

However, these benefits of \CP may not materialize in many applications unless \CP can be better integrated into existing classifiers.
These are predominantly deep networks, trained end-to-end to, \eg, optimize classification performance.
\CP, in contrast, is agnostic to the underlying model, being applied as ``wrapper'' post-training, such that the obtained confidence sets may not be optimal, \eg, in terms of size (\emph{inefficiency}) or composition (\ie, the included classes).
Especially in the medical domain, constraints on the confidence sets can be rather complex.
Our \emph{conformal training (\CT)} integrates \CP into the training procedure, allowing to optimize very specific objectives defined on the predicted confidence sets -- without \red{losing} the guarantees.
In medical diagnosis, smaller confidence sets may avoid confusion or anxiety among doctors or patients, ultimately leading to better diagnoses.
For example, we can reduce inefficiency (\ie, the ambiguity of predicted conditions) for conditions that are particularly difficult for doctors to diagnose.
Alternatively, \CT allows to avoid confusion between low- and high-risk conditions within the confidence sets.

Generally, beyond medical diagnosis, we believe \CT to have positive impact in settings where additional constraints on confidence sets are relevant \emph{in addition} to the guarantees and uncertainty estimates provided by \CP.

\section*{Reproducibility Statement}

In order to ensure reproducibility, we include a detailed description of our experimental setup in \appref{sec:app-experimental-setup}.
We discuss all necessary information for conformal training (\CT) as well as our baselines.
This includes architectures, training procedure and hyper-parameters, as well as pre-processing/data augmentation if applicable.
Furthermore, we describe our evaluation procedure which includes multiple calibration/test splits for conformal prediction (\CP) at test time as well as multiple training runs to capture randomness in the used calibration examples and during training.
To this end, \tabref{tab:app-datasets} reports the training/calibration/test splits of all used datasets and \tabref{tab:app-parameters} the used hyper-parameters for \CT.
While \algref{alg:ct} already summarizes the used (smooth) threshold \CP methods and our \CT, \appref{sec:app-code} (specifically \algref{alg:app-ct}) lists the corresponding Python implementation of these key components.

\bibliography{bibliography}
\bibliographystyle{iclr2022_conference}

\clearpage
\begin{appendix}
    \renewcommand{\topfraction}{0.9}
    \renewcommand{\bottomfraction}{0.8}
    \setcounter{topnumber}{2}
    \setcounter{bottomnumber}{2}
    \setcounter{totalnumber}{4}
    \setcounter{dbltopnumber}{2}
    \renewcommand{\dbltopfraction}{0.9}
    \renewcommand{\textfraction}{0.07}
    \renewcommand{\floatpagefraction}{0.7}
    \renewcommand{\dblfloatpagefraction}{0.7}
    \setcounter{figure}{0}  
    \renewcommand{\thefigure}{\Alph{figure}} 
    \setcounter{table}{0}  
    \renewcommand{\thetable}{\Alph{table}} 
    \setcounter{algorithm}{0}  
    \renewcommand{\thealgorithm}{\Alph{algorithm}} 
    \section{Overview and Outline}

In the appendix, we discuss an additional baseline, called \emph{coverage training (\CovT)}, provide additional details on our experimental setup and include complementary results.
Specifically, the appendix includes:
\begin{itemize}
    \item Additional discussion of related work in \secref{sec:app-related-work};
    \item Formal statement of the coverage guarantee obtained through conformal prediction (\CP) in \secref{sec:app-guarantee};
    \item \red{Differentiable implementation of \APS in \secref{sec:app-differentiable-aps}};
    \item Discussion of \emph{coverage training (\CovT)} and \citep{BellottiARXIV2021} in \secref{sec:app-coverage-training};
    \item Details on our experimental setup, including dataset splits, model details and used hyper-parameters for \CT, in \secref{sec:app-experimental-setup};
    \item Experiments regarding random training and test trials in \secref{sec:app-trials};
    \item Hyper-parameter ablation on MNIST in \secref{sec:app-parameters};
    \item \CovT and \CT ablation on MNIST and Fashion-MNIST in \secref{sec:app-ablation};
    \item Complete inefficiency (\Ineff) results on all datasets in \secref{sec:app-results};
    \item Effect of (standard) \CT on class-conditional inefficiency and coverage (\Cov) confusion in \secref{sec:app-conformal-training};
    \item Additional results for \CT shaping the class-conditional inefficiency distribution in \secref{sec:app-ineff};
    \item More experiments for \CT manipulating coverage confusion in \secref{sec:app-confusion};
    \item Complementary results for \CT reducing mis-coverage (\MisCov) in \secref{sec:app-miscoverage};
    \item Class-conditional inefficiency and coverage confusion on binary datasets in \secref{sec:app-binary};
    \item Python- and Jax \citep{Bradbury2018} code for \CT in \secref{sec:app-code}.
\end{itemize}

\section{Related Work}
\label{sec:app-related-work}

Conformal prediction (\CP) builds on early work by \cite{Vovk2005} considering both regression, see \eg, \citep{RomanoNIPS2019} for references, and classification settings, \eg \citep{RomanoNIPS2020,AngelopoulosICLR2020,CauchoisARXIV2020,HechtlingerARXIV2018}.
Most of these approaches follow a \emph{split} \CP approach \citep{LeiAMAI2013} where a held-out calibration set is used, as in the main paper, however, other variants based on cross-validation \citep{VovkAMAI2013} or jackknife (\ie, leave-one-out) \citep{BarberARXIV2019b} are available. 
These approaches mostly provide marginal coverage.
\cite{VovkML2013,BarberARXIV2019} suggest that it is generally difficult or impossible to obtain conditional coverage.
However, \cite{RomanoNIPS2020} work towards \emph{empirically} better conditional coverage and \cite{SadinleJASA2019} show that efficient \emph{class-conditional} coverage is possible.
\cite{AngelopoulosICLR2020} extend the work by \cite{RomanoNIPS2020} to obtain smaller confidence sets at the expense of the obtained empirical conditional coverage.
\CP has also been studied in the context of ensembles \citep{YangARXIV2021}, allowing to perform model selection based on inefficiency while keeping coverage guarantees.
The work of \cite{BatesARXIV2021} can be seen as a \CP extension in which a guarantee on an arbitrary, user-specified risk can be obtained, using a conformal predictor similar to \citep{SadinleJASA2019}.
Our conformal training (\CT) follows the split \CP approach and is specifically targeted towards classification problems.
Nevertheless, extensions to regression, or other \CP formulations such as \citep{BatesARXIV2021} during training, are possible.
Beyond that, \CT is agnostic to the \CP method used at test time and can thus be seen as complementary to the \CP methods discussed above.
\red{This means that \CT can easily be combined with approaches such as \citep{BatesARXIV2021} or class-conditional conformal predictors \citep{SadinleJASA2019} at test time.}

In terms of \emph{learning} to predict confidence sets, our approach has similarities to the \emph{multiple choice learning} of \cite{GuzmanriveraNIPS2012} which yields multiple possible outputs in structured prediction settings (\eg, image segmentation).
However, the obtained prediction sets are fixed size and no coverage guarantee is provided.
Concurrent work by \cite{BellottiARXIV2021} is discussed in detail in \appref{sec:app-coverage-training}.

\section{Coverage Guarantee}
\label{sec:app-guarantee}

Following \cite{RomanoNIPS2020}, we briefly state the coverage guarantee obtained by \CP in formal terms:
Given that the learning algorithm used is invariant to permutations of the training examples, and the calibration examples $\{(X_i, Y_i)\}_{i \in \Ical}$ are exchangeably drawn from the same distribution encountered at test time, the discussed \CP methods satisfy
\begin{align}
    P(Y \in C(X)) \geq 1 - \alpha.
\end{align}
As highlighted in \citep{RomanoNIPS2020}, this bound is near tight if the scores $E(x_i)$ are almost surely distinct:
\begin{align}
    P(Y \in C(X)) \leq 1 - \alpha + \frac{1}{|\Ical| + 1}.
\end{align}
Note that this is the case for \APS due to the uniform random variable $U$ in \eqnref{eq:aps}.
\citep{RomanoNIPS2020} notes that there is generally no guarantee on conditional coverage, as this requires additional assumptions.
However, class-conditional coverage can be obtained using \Thr as outlined in \citep{SadinleJASA2019}.
Moreover, \cite{SadinleJASA2019} show that \Thr is the most efficient conformal predictor given a fixed model $\pi_\theta$, \ie, minimizes inefficiency.
We refer to \citep{SadinleJASA2019} for exact statements of the latter two findings.

\section{Differentiable \APS}
\label{sec:app-differentiable-aps}

\red{Our differentiable implementation closely follows the one for \Thr outlined in \secref{subsec:differentiable-conformal-predictors}. The main difference is the conformity score $E(x, k)$ computation, \ie,
\begin{align}
E_\theta(x, k) := \pi_{\theta,y^{(1)}}(x) + \ldots + \pi_{\theta,y^{(k - 1)}}(x) + U\pi_{\theta,y^{(k)}}(x)\label{eq:aps},
\end{align}
where $\pi_{\theta,y^{(1)}}(x) \geq \ldots \geq \pi_{\theta, y^{(K)}}(x)$ and $U$ is a uniform random variable in $[0,1]$ to break ties. As in the calibration step, we use an arbitrary smooth sorting approach for this.
This implementation could easily be extended to include the regularizer of \cite{AngelopoulosICLR2020}, as well.}

\section{Coverage Training}
\label{sec:app-coverage-training}

\begin{figure}[b]
    \vspace*{-0.1cm}
    \centering
    \begin{minipage}[t]{0.525\textwidth}
        \vspace*{-7.5px}
        
        \begin{tcolorbox}[arc=0mm,colback=white,colframe=black,boxrule=0.1mm,bottom=0.25mm,top=0.25mm,left=0.25mm,right=0.25mm]
            \begin{algorithmic}[1]
                \footnotesize 
                \Function{CoverageTraining}{$\tau$, $\lambda$}
                    \For{mini-batch $B$}
                        \State $C(x_i; \tau) :=$ \Call{SmoothPred}{$\pi_\theta(x_i)$, $\tau$}, $i{\in}B$
                        \State $\mathcal{L}_B := \sum_{i \in B} \mathcal{L}(C_\theta(x_i; \tau), y_i)$
                        \State $\Omega_B := \sum_{i \in B}\Omega(C_\theta(x_i;\tau))$
                        \State $\Delta := \nabla_\theta \nicefrac{1}{|B|}(\mathcal{L}_B + \lambda\Omega_B)$
                        \State update parameters $\theta$ using $\Delta$
                    \EndFor
                \EndFunction
            \end{algorithmic}
        \end{tcolorbox}
    \end{minipage}
    \hfill
    \begin{minipage}[t]{0.465\textwidth}
        \vspace*{-7px}
        
        \captionof{algorithm}{
        \textbf{Coverage Training (\CovT)}:
        Compared to \algref{alg:ct} for \CT, \CovT simplifies training by not differentiating through the calibration step and avoiding splitting the batch $B$ in half.
        However, fixing the threshold $\tau$ can be a problem and training requires both coverage and size loss.
        }
        \label{alg:app-covt}
    \end{minipage}
    \vspace*{-0.2cm}
\end{figure}

As intermediate step towards \emph{conformal training (\CT)}, we can also ignore the calibration step and just differentiate through the prediction step, \ie, $C_\theta(X; \tau)$.
This can be accomplished by fixing the threshold $\tau$.
Then, $\pi_\theta$ essentially learns to produce probabilities that yield ``good'' confidence sets $C_\theta(X; \tau)$ for the chosen threshold $\tau$.
Following \algref{alg:app-covt}, \emph{coverage training (\CovT)} computes $C_\theta(X;\tau)$ on each mini-batch using a fixed $\tau$.
The model's parameters $\theta$ are obtained by solving
\begin{align}
    \min_\theta \log\left(\mathbb{E}\left[\mathcal{L}(C_\theta(X;\tau), Y) + \lambda \Omega(C_\theta(X;\tau))\right]\right).\label{eq:app-coverage-training}
\end{align}
Again, $\mathcal{L}$ is the classification loss from \eqnref{eq:classification-loss} and $\Omega$ the size loss from \eqnref{eq:size-loss}.
The classification loss has to ensure that the true label $y$ is in the predicted confidence set $C_\theta(X;\tau)$ as the calibration step is missing.
In contrast to \CT, \CovT strictly requires both classification and size loss during training.
This is because using a fixed threshold $\tau$ yields trivial solutions for both classification and size loss when used in isolation (\ie, $\mathcal{L}$ is minimized for $C_\theta(X;\tau) = [K]$ and $\Omega$ is minimized for $C_\theta(X;\tau) = \emptyset$).
Thus, balancing both terms in \eqnref{eq:app-coverage-training} using $\lambda$ is crucial during training.
As with \CT, the threshold $\tau$ is re-calibrated at test time to obtain a coverage guarantee.
Choosing $\tau$ for training, in contrast, can be difficult:
First, $\tau$ will likely evolve during training (when $\pi_\theta$ gets more and more accurate) and, second, the general ballpark of reasonable thresholds $\tau$ depends on the dataset as well as model and is difficult to predict in advance.

In concurrent work by \cite{BellottiARXIV2021} (referred to as \Bel), the problem with fixing a threshold $\tau$ is circumvented by using \ThrL during training, \ie, \Thr on logits.
As the logits are unbounded, the threshold can be chosen arbitrarily, \eg, $\tau = 1$.
As \Belc also follows the formulation of \eqnref{eq:app-coverage-training}, the approach can be seen as a special case of \CovT.
However, a less flexible \emph{coverage loss} is used during training:
Instead of \Lclass, the loss is meant to enforce a specific coverage level $(1 - \alpha)$ on each mini-batch.
This is done using a squared loss on coverage:
\begin{align}
    \Lcov := \left[\left(\frac{1}{|B|}\sum_{i \in B} C_{\theta,y_i}(x_i;\tau)\right) - (1 - \alpha)\right]^2\label{eq:app-coverage-loss}
\end{align}
for a mini-batch $B$ of examples.
In contrast to \eqnref{eq:app-coverage-training}, \Lcov is applied per batch and not per example.
For the size loss, \cite{BellottiARXIV2021} uses $\kappa = 0$ in \eqnref{eq:size-loss}.
Besides \emph{not} providing much control over the confidence sets, \Lcov also encourages coverage $(1 - \alpha)$ instead of perfect coverage.
Nevertheless, this approach is shown to improve inefficiency of \ThrL on various UCI datasets \citep{Dua2019} using linear logistic regression models.
The experiments in the main paper show that this generalizes to non-linear models and more complex datasets.
Nevertheless, \Belc is restricted to \ThrL which is outperformed significantly by both \Thr and \APS.
Thus, \Belc is consistently outperformed by \CT in terms of inefficiency improvements.
Moreover, the approach \emph{cannot} be used for any of the studied use cases in \secref{subsec:conformal-training-applications}.

Using \CovT with \Thr and \APS remains problematic.
While we found $\tau \in [0.9, 0.99]$ (or $[-0.1, -0.01$ for \ThrLP) to work reasonably on some datasets, we had difficulties on others, as highlighted in \secref{sec:app-ablation}.
Moreover, as \CovT requires balancing coverage $\mathcal{L}$ and size loss $\Omega$, hyper-parameter optimization is more complex compared to \CT.
By extension, these problems also limit the applicability of \Belc.
Thus, we would ideally want to re-calibrate the threshold $\tau$ after each model update.
Doing calibration on a larger, held-out calibration set, however, wastes valuable training examples and compute resources.
Thus, \CT directly calibrates on each mini-batch and also differentiates through the calibration step itself to obtain meaningful gradients.

\section{Experimental Setup}
\label{sec:app-experimental-setup}

\begin{table}
    \vspace*{-0.2cm}
    \caption{
    \textbf{Used Datasets:}
    Summary of train/calibration/test splits, epochs and models used on all datasets in our experiments.
    The calibration set is usually less than 10\% of the training set.
    On most datasets, the test set is roughly two times larger than the calibration set.
    When computing random calibration/test splits for evaluation, see text, the number of calibration and test examples stays constant.
    * On Camelyon, we use features provided by \cite{WilderIJCAI2020} instead of the original images.
    ** For EMNIST, we use a custom subset of the ``byClass'' split.
    }
    \label{tab:app-datasets}
    \vspace*{-6px}
    \centering
    \small
    \begin{tabular}{|l|c|c|c|c|c|c|c|}
        \hline
        \multicolumn{8}{|c|}{Dataset Statistics}\\
        \hline
         Dataset & Train & Cal & Test & Dimensions & Classes & Epochs & Model\\
         \hline\hline
         Camelyon2016* \citep{BejnordiJAMA2017} & 280 & 100 & 17 & $31$ & 2 & 100 & 1-layer MLP\\
         GermanCredit \citep{Dua2019} & 700 & 100 & 200 & $24$ & 2 & 100 & Linear\\
         WineQuality \citep{CortezDSS2009} & 4500 & 500 & 898 & $11$ & 2 & 100 & 2-layer MLP\\
         MNIST \citep{LecunIEEE1998} & 55k & 5k & 10k & $28\times28$ & 10 & 50 & Linear\\
         EMNIST** \citep{CohenARXIV2017} & 98.8k & 5.2k & 18.8k & $28\times28$ & 52 & 75 & 2-layer MLP\\
         Fashion-MNIST \citep{XiaoARXIV2017} & 55k & 5k & 10k & $28\times28$ & 10 & 150 & 2-layer MLP\\
         CIFAR10 \citep{Krizhevsky2009} & 45k & 5k & 10k & $32\times32\times3$ & 10 & 150 & ResNet-34\\
         CIFAR100 \citep{Krizhevsky2009} & 45k & 5k & 10k & $32\times32\times3$ & 100 & 150 & ResNet-50\\
         \hline
    \end{tabular}
    \vspace*{-0.1cm}
\end{table}
\textbf{Datasets and Splits:}
We consider Camelyon2016 \citep{BejnordiJAMA2017}, GermanCredit \citep{Dua2019}, WineQuality \citep{CortezDSS2009}, MNIST \citep{LecunIEEE1998}, EMNIST \citep{CohenARXIV2017}, Fashion-MNIST \citep{CohenARXIV2017} and CIFAR \citep{Krizhevsky2009} with a fixed split of training, calibration and test examples.
\tabref{tab:app-datasets} summarizes key statistics of the used datasets which we elaborate on in the following.
Except Camelyon, all datasets are provided by Tensorflow \citep{Apadi2015}\footnote{\url{https://www.tensorflow.org/datasets}}.
For Camelyon, we use the pre-computed features of \cite{WilderIJCAI2020} which are based on open source code from the Camelyon2016 challenge\footnote{\url{https://github.com/arjunvekariyagithub/camelyon16-grand-challenge}}.
For datasets providing a default training/test split, we take the last 10\% of training examples as calibration set.
On Camelyon, we use the original training set, but split test examples into $100$ validation and $17$ test examples.
This is because less than $100$ calibration examples are not meaningful for $\alpha{=}0.05$.
As we evaluate 10 random calibration/test splits, the few test examples are not problematic in practice.
On GermanCredit and WineQuality, we manually created training/calibration/test splits, roughly matching 70\%/10\%/20\%.
We use the ``white wine'' subset for WineQuality; to create a binary classification problem, wine with quality 6 or higher is categorized as ``good wine'' (class 1), following \citep{BellottiARXIV2021}.
Finally, for EMNIST, we consider a subset of the ``byClass'' split that contains $52 = 2\cdot 26$ classes comprised of all lower and upper case letters.
We take the first 122.8k examples, split as in \tabref{tab:app-datasets}.

\textbf{Models and Training:}
We consider linear models, multi-layer perceptrons (MLPs) and ResNets \citep{HeCVPR2016} as shown in \tabref{tab:app-datasets}.
Specifically, we use a linear model on MNIST and GermanCredit, 1- or 2-layer MLPs on Camelyon2016, WineQuality and Fashion-MNIST, and ResNet-34/50 \citep{HeCVPR2016} on CIFAR10/100.
Models and training are implemented in Jax \citep{Bradbury2018}\footnote{\url{https://github.com/google/jax}} and the ResNets follow the implementation and architecture provided by Haiku \citep{Hennigan2020}\footnote{\url{https://github.com/deepmind/dm-haiku}}.
Our $l$-layer MLPs comprise $l$ \emph{hidden} layers.
We use $32$, $256$, $128$, $64$ units per hidden layer on Camelyon, WineQuality, EMNIST and Fashion-MNIST, respectively.
These were chosen by grid search over $\{16, 32, 64, 128, 256\}$.
In all cases, we use ReLU activations \citep{NairICML2010} and batch normalization \citep{IoffeICML2015}.
We train using stochastic gradient descent (SGD) with momentum $0.0005$ and Nesterov gradients.
The baseline models are trained with cross-entropy loss, while \CT follows \algref{alg:ct} and \CovT follows \algref{alg:app-covt}.
Learning rate and batch size are optimized alongside the \CT hyper-parameters using grid search, see below.
The number of epochs are listed in \tabref{tab:app-datasets} and we follow a multi-step learning rate schedule, multiplying the initial learning rate by $0.1$ after $\nicefrac{2}{5}$, $\nicefrac{3}{5}$ and $\nicefrac{4}{5}$ of the epochs.
We use Haiku's default initializer.
On CIFAR, we apply whitening using the per-channel mean and standard deviation computed on the training set.
On the non-image datasets (Camelyon, GermanCredit, WineQuality), we whiten each feature individually.
On MNIST, EMNIST and Fashion-MNIST, the input pixels are just scaled to $[-1, 1]$.
Except on CIFAR, see next paragraph, we do \emph{not} use any data augmentation.
Finally, we do \emph{not} use Platt scaling \citep{GuoICML2017} as used in \citep{AngelopoulosICLR2020}.

\begin{table}
    \vspace*{-0.2cm}
    \caption{
    \textbf{Used \CT Hyper-Parameters} with and without \Lclass for \ThrLP during training and \Thr at test time.
    The hyper-parameters for \APS at test time might vary slightly from those reported here.
    The exact grid search performed to obtained these hyper-parameters can be found in the text.
    Note that, while hyper-parameters fluctuate slightly, $\lambda$ needs to be chosen higher when training with \Lclass.
    Additionally, and in contrast to \Belc, $\kappa = 1$ in \eqnref{eq:size-loss} performs better, especially combined with \Lclass.
    Note that dispersion for smooth sorting is fixed to $\epsilon = 0.1$.
    }
    \label{tab:app-parameters}
    \vspace*{-6px}
    \centering
    \small
    \begin{tabular}{|l|c|c|c|c|c|}
        \hline
        \multicolumn{6}{|c|}{\textbf{\CT Hyper-Parameters} (for \ThrLP during training and \Thr at test time)}\\
        \hline
        Dataset, Method
        & \begin{tabular}{@{}c@{}}Batch\\Size\end{tabular}
        & \begin{tabular}{@{}c@{}}Learning\\rate\end{tabular}
        & \begin{tabular}{@{}c@{}}Temp.\\$T$\end{tabular}
        & \begin{tabular}{@{}c@{}}Size\\weight $\lambda$\end{tabular}
        & $\kappa$ in \eqnref{eq:size-loss}\\
        \hline
        \hline
        Camelyon, \CT & 20 & 0.005 & 0.1 & 5 & 1\\
        Camelyon, \CT+\Lclass & \red{10} & \red{0.01} & 0.01 & 5 & 1\\
        \hline
        GermanCredit, \CT & 200 & 0.05 & 1 & 5 & 1\\
        GermanCredit, \CT+\Lclass & 400 & 0.05 & 0.1 & 5 & 1\\
        \hline
        WineQuality, \CT & 100 & 0.005 & 0.5 & 0.05 & 1\\
        WineQuality, \CT+\Lclass & 100 & 0.005 & 0.1 & 0.5 & 1\\
        \hline
        MNIST, \CT & 500 & 0.05 & 0.5 & 0.01 & 1\\
        MNIST, \CT+\Lclass & 100 & 0.01 & 1 & 0.5 & 1\\
        \hline
        EMNIST, \CT & 100 & 0.01 & 1 & 0.01 & 1\\
        EMNIST, \CT+\Lclass & 100 & 0.01 & 1 & 5 & 1\\
        \hline
        Fashion-MNIST, \CT & 100 & 0.01 & 0.1 & 0.01 & 0\\
        Fashion-MNIST, \CT+\Lclass & 100 & 0.01 & 0.1 & 0.5 & 1\\
        \hline
        CIFAR10, fine-tune \CT & 500 & 0.01 & 1 & 0.05 & 0\\
        CIFAR10, fine-tune \CT+\Lclass & 500 & 0.05 & 0.1 & 1 & 1\\
        CIFAR10, ``extend'' \CT & 100 & 0.01 & 1 & 0.005 & 0\\
        CIFAR10, ``extend'' \CT+\Lclass & 500 & 0.05 & 0.1 & 0.1 & 1\\
        \hline
        CIFAR100, fine-tune \CT & 100 & 0.005 & 1 & 0.005 & 0\\
        CIFAR100, fine-tune \CT+\Lclass & 100 & 0.005 & 1 & 0.01 & 1\\
        \hline
    \end{tabular}
    \vspace*{-0.1cm}
\end{table}
\textbf{Fine-Tuning on CIFAR:}
On CIFAR10 and CIFAR100, we train base ResNet-34/ResNet-50 models which are then fine-tuned using \Belc, \CovT or \CT.
We specifically use a ResNet-34 with only 4 base channels to obtain an accuracy of 82.6\%, using only random flips and crops as data augmentation.
The rationale is to focus on the results for \CP at test time, without optimizing accuracy of the base model.
On CIFAR100, we use 64 base channels for the ResNet-50 and additionally employ AutoAugment \citep{CubukARXIV2018} and Cutout \citep{DevriesARXIV2017} as data augmentation.
This model obtains 73.64\% accuracy.
These base models are trained on 100\% of the training examples (without calibration examples).
For fine-tuning, the last layer (\ie, logit layer) is re-initialized and trained using the same data augmentation as applied for the base model, subject to the random training trials described below.
We also consider ``extending'' the ResNet by training a 2-layer MLP with 128 units per hidden layer on top of the features (instead of re-initializing and fine-tuning the logit layer).
All reported results either correspond to fine-tuned (\ie, linear model on features) or extended models (\ie, 2-layer MLP on features) trained on these base models.

\textbf{Hyper-Parameters:}
The final hyper-parameters selected for \CT (for \Thr at test time) on all datasets are summarized in \tabref{tab:app-parameters}.
These were obtained using grid search over the following hyper-parameters:
batch size in $\{1000, 500, 100\}$ for WineQuality, MNIST, EMNIST, Fashion-MNIST and CIFAR, $\{300, 200, 100, 50\}$ on GermanCredit and $\{80, 40, 20, 10\}$ on Camelyon;
learning rate in $\{0.05, 0.01, 0.005\}$;
temperature $T \in \{0.01, 0.1, 0.5, 1\}$;
size weight $\lambda \in \{0.0001, 0.0005, 0.001, 0.005, 0.01, 0.05, 0.1, 0.5, 1, 5, 10\}$ (\cf \eqnref{alg:ct}, right);
and $\kappa \in \{0, 1\}$ (\cf \eqnref{eq:size-loss}).
Grid search was done for each dataset individually on 100\% of the training examples (\cf \tabref{tab:app-datasets}).
That is, for hyper-parameter optimization we did \emph{not} perform random training trials as described next.
The best hyper-parameters according to inefficiency after evaluating 3 random calibration/test splits were selected, both for \Thr and \APS at test time, with and without \Lclass.

\tabref{tab:app-parameters} allows to make several observations.
First, on the comparably small (and binary) datasets Camelyon and GermanCredit, the size weight $\lambda = 5$ is rather high.
For \CT without \Lclass, this just indicates that a higher learning rate could be used.
Then using \Lclass, however, this shows that the size loss is rather important for \CT, especially on binary datasets.
Second, we found the temperature $T$ to have low impact on results, also see \secref{sec:app-parameters}.
On multiclass datasets, the size weight $\lambda$ is usually higher when employing \Lclass.
Finally, especially with \Lclass, using ``valid'' size loss, \ie, $\kappa = 1$, to not penalize confidence sets of size 1, works better than $\kappa = 0$.

\begin{table}[t]
    \vspace*{-0.2cm}
    \caption{
    \textbf{Importance of Random Trials:}
    We report coverage and inefficiency with the corresponding standard deviation across 10 \emph{test} (left) and 10 \emph{training} trials (right).
    \CT was trained using \ThrLP if not stated otherwise.
    For test trials, a fixed model is used.
    Results for training trials additionally include 10 test trials, but the standard deviation is reported only across the training trials.
    These results help to disentangle the impact of test and training trials.
    For example, while \CT with \APS (during training) works in the best case, the standard deviation of 3.1 across multiple training trials indicates that training is \emph{not} stable.
    }
    \label{tab:app-trials}
    \vspace*{-6px}
    \centering
    \small
    \begin{minipage}[t]{0.44\textwidth}
        \vspace*{0px}
        
        \begin{tabular}{|l|c|c|c|}
            \hline
            \multicolumn{4}{|c|}{\textbf{MNIST:} \emph{test} trials, \Cov/\Ineff for \Thr}\\
            \hline
            Method & \Acc & \Cov & \Ineff\\
            \hline\hline
            Baseline & 92.45 & 99.09$\pm$0.2 & 2.23$\pm$0.15\\
            \CT & 90.38 & 99.05$\pm$0.2 & 2.14$\pm$0.13\\
            \CT +\Lclass & 91.14 & 99.03$\pm$0.19 & 2.09$\pm$0.12\\
            \hline
        \end{tabular}
    \end{minipage}
    \begin{minipage}[t]{0.53\textwidth}
        \vspace*{0px}
        
        \begin{tabular}{|l|c|c|c|}
            \hline
            \multicolumn{4}{|c|}{\textbf{MNIST:} \emph{Training} trials, \Cov/\Ineff for \Thr}\\
            \hline
            Method & \Acc & \Cov & \Ineff\\
            \hline\hline
            Baseline & 92.4$\pm0.06$ & 99.09$\pm$0.8 & 2.23$\pm$0.01\\
            \CT & 90.2$\pm$0.12 & 99.03$\pm$0.22 & 2.18$\pm$0.025\\
            \CT +\Lclass & 91.2$\pm0.05$ & 99.05$\pm$0.21 & 2.11$\pm$0.028\\
            \hline
            \CT with \APS & 87.9$\pm$4.81 & 99.09$\pm$0.29 & 5.79$\pm${\color{red}3.1}\\
            \hline
        \end{tabular}
    \end{minipage}
    \vspace*{-0.1cm}
\end{table}

\textbf{Random Training and Test Trials:}
For statistically meaningful results, we perform random \emph{test} and \emph{training} trials.
Following common practice \citep{AngelopoulosICLR2020}, we evaluate \CP methods at test time using 10 random calibration/test splits.
To this end, we throw all calibration and test examples together and sample a new calibration/test split for each trial, preserving the original calibration/test composition which is summarized in \tabref{tab:app-datasets}.
Metrics such as coverage and inefficiency are then empirically evaluated as the average across all test trials.
Additionally, and in contrast to \citep{BellottiARXIV2021}, we consider random training trials:
After hyper-parameters optimization on all training examples, we train 10 models with the final hyper-parameters on a new training set obtained by sampling the original one with up to 5 replacements.
For example, on MNIST, with 55k training examples, we randomly sample 10 training sets of same size with each, on average, containing only $\sim$68\% unique examples from the original training set.
Overall, this means that we report, \eg, inefficiency as average over a total of $10 \cdot 10 = 100$ random training \emph{and} test trials.
As a consequence, our evaluation protocol accounts for randomness at test time (\ie, regarding the calibration set) and at training time (\ie, regarding the training set, model initialization, \etc).

\section{Importance of Random Trials}
\label{sec:app-trials}

In \tabref{tab:app-trials} we highlight the importance of random training and test trials for evaluation.
On the left, we show the impact of trials at test time, \ie, $10$ random calibration/test splits, for a fixed model on MNIST.
While the standard deviation of coverage is comparably small, usually $\leq0.2\%$, standard deviation of inefficiency is higher in relative terms.
This makes sense as coverage is guaranteed, while inefficiency depends more strongly on the sampled calibration set.
The right table, in contrast, shows that training trials exhibit lower standard deviation in terms of inefficiency.
However, training with, \eg, \APS will mainly result in high inefficiency, on average, because of large standard deviation.
In fact, \CT with \APS or \Thr at training time results in worse inefficiency mainly because training is less stable.
This supports the importance of running multiple training trials for \CT.

\section{Impact of Hyper-Parameters}
\label{sec:app-parameters}

\begin{table}
    \vspace*{-0.2cm}
    \caption{
    \textbf{Hyper-Parameter Ablation on MNIST:}
    For \CT without \Lclass, we report inefficiency and accuracy when varying hyper-parameters individually:
    batch size/learning rate, size weight $\lambda$, temperature $T$ and confidence level $\alpha$.
    While size weight $\lambda$ and temperature $T$ have insignificant impact, too small batch size can prevent \CT from converging.
    Furthermore, the chosen hyper-parameters do not generalize well to higher confidence levels $\alpha \in \{0.1, 0.05\}$.
    }
    \label{tab:app-ablation}
    \vspace*{-6px}
    \centering
    \small
    \begin{tabular}{|l|c|c|c|c|c|c|c|c|c|c|}
        \hline
        \multicolumn{10}{|c|}{Batch Size and Learning Rate}\\
        \hline
        Batch Size & 1000 & 1000 & 1000 & \bfseries 500 & 500 & 500 & 100 & 100 & 100\\
        Learning Rate & 0.05 & 0.01 & 0.005 & \bfseries 0.05 & 0.01 & 0.005 & 0.05 & 0.01 & 0.005\\
        \hline
        \hline
        \Ineff & 2.27 & 2.24 & 2.24 & 2.18 & 2.18 & \bfseries 2.17 & 8.04 & 7.32 & 9.66\\
        \Acc & 89.05 & 89.18 & 89.06 & 90.23 & 90.22 & \bfseries 90.27 & 11.5 & 22.46 & 12.13\\
        \hline
    \end{tabular}
    
    \begin{tabular}{|l|c|c|c|c|c|c|c|}
        \hline
        \multicolumn{8}{|c|}{Size Weight $\lambda$}\\
        \hline
        $\lambda$ & 0.001 & 0.005 & \bfseries 0.01 & 0.05 & 0.1 & 1 & 10\\
        \hline
        \hline
        \Ineff & 2.18 & 2.18 & 2.18 & 2.19 & 2.19 & 2.19 & \bfseries 2.16\\
        \Acc & 90.2 & 20.23 & 90.23 & 90.2 & 90.25 & 90.23 & \bfseries 90.26\\
        \hline
    \end{tabular}
    
    \begin{tabular}{|l|c|c|c|c|c|c|c|}
        \hline
        \multicolumn{8}{|c|}{Temperature $T$}\\
        \hline
        $T$ & 0.01 & 0.05 & 0.1 & \bfseries 0.5 & 1 & 5 & 10\\
        \hline
        \hline
        \Ineff & 2.39 & 2.23 & 2.2 & 2.19 & \bfseries 2.18 & 2.2 & 2.29\\
        \Acc & 88.54 & 89.94 & 90.02 & 90.24 & \bfseries 90.28 & 90.05 & 89.63\\
        \hline
    \end{tabular}
    
    \begin{tabular}{|l|c|c|c|c|}
        \hline
        \multicolumn{5}{|c|}{Confidence Level $\alpha$ (during training)}\\
        \hline
        $\alpha$ & 0.1 & 0.05 & \bfseries 0.01 & 0.005\\
        \hline
        \hline
        \Ineff & 8.07 & 7.23 & 2.18 & \bfseries 2.17\\
        \Acc & 12.88 & 39.82 & \bfseries 90.23 & 89.47\\
        \hline
    \end{tabular}
    \vspace*{-0.1cm}
\end{table}
In \tabref{tab:app-ablation}, we conduct ablation for individual hyper-parameters of \CT with \ThrLP and without \Lclass on MNIST.
The hyper-parameters used in the main paper, \cf \tabref{tab:app-parameters}, are highlighted in \textbf{bold}.
As outlined in \secref{sec:app-experimental-setup}, hyper-parameter optimization was conducted on 100\% training examples with only 3 random test trials, while \tabref{tab:app-ablation} shows results using random training \emph{and} test trials.
We found batch size and learning rate to be most impactful.
While batch sizes 1000 and 500 both work, batch size 100 prevents \CT from converging properly.
This might be due to the used $\alpha = 0.01$ which might be too low for batch size 100 where only 50 examples are available for calibration during training.
Without \Lclass, the size weight $\lambda$ merely scales the learning rate and, thus, has little to no impact.
For \CT \emph{with} \Lclass, we generally found the size weight $\lambda$ to be more important for balancing classification loss $\mathcal{L}$ and size loss $\Omega$ in \eqnref{eq:conformal-training-classification}.
Temperature has no significant impact, although a temperature of 0.5 or 1 works best.
Finally, the hyper-parameters do generalize to a lower confidence level $\alpha = 0.005$.
Significantly lower values, \eg, $\alpha = 0.001$, are, however, not meaningful due to the batch size of $500$.
However, significantly higher confidence levels, \eg, $\alpha = 0.1$ or $\alpha = 0.05$, require re-optimizing the other hyper-parameters.

\section{\CovT and \CT Ablation on MNIST and Fashion-MNIST}
\label{sec:app-ablation}

\begin{table}
    \vspace*{-0.2cm}
    \caption{
    \textbf{Ablation for \CovT and \CT on MNIST and Fashion-MNIST:}
    We report inefficiency and accuracy for \citep{BellottiARXIV2021} (\Bel), \CovT and \CT considering various \CP methods for training and testing.
    \Bel outperforms the baseline when using \ThrL, but does not do so for \Thr on MNIST.
    \CovT with \Thr or \APS during training is challenging, resulting in high inefficiency (mainly due to large variation among training trials, \cf \tabref{tab:app-trials}), justifying our choice of \ThrLP for \CT.
    Also \CovT is unable to improve over the \Thr baseline.
    Similar observations hold on Fashion-MNIST where, however, \CovT with \Thr or \APS was not possible.
    }
    \label{tab:app-fashion-mnist}
    \vspace*{-6px}
    \centering
    \small
    \begin{tabular}{|l|c|c|c|c|c|c|c|c|c|c|c|c|c|}
        \hline
        \multicolumn{14}{|c|}{\textbf{MNIST:} Ablation for \CovT and \CT}\\
        \hline
        Method & \multicolumn{3}{c|}{Baseline}
        & \multicolumn{2}{c|}{\Bel}
        & \multicolumn{4}{c|}{\CovT}
        & \multicolumn{4}{c|}{\CT}\\
        \hline
        Train &&&
        & \multicolumn{2}{c|}{\ThrL}
        & \Thr & \APS
        & \multicolumn{2}{c|}{\ThrLP}
        & \ThrLP & \ThrLP
        & \multicolumn{2}{c|}{+\Lclass}\\
        \hline
        Test & \ThrL & \Thr & \APS
        & \ThrL & \Thr
        & \Thr & \APS
        & \Thr & \APS
        & \Thr & \APS
        & \Thr & \APS\\
        \hline\hline
        Avg. \Ineff & 3.57 & 2.23 & 2.5 & 2.73 & 2.7 & 6.34 & 4.86 & 2.5 & 2.76 & 2.18 & 2.16 & 2.11 & 2.14\\
        Avg. \Acc & 92.39 & 92.39 & 92.39 & 81.41 & 90.01 & 83.85 & 88.53 & 92.63 & 92.63 & 90.24 & 90.21 & 91.18 & 91.35\\
        \hline
    \end{tabular}
    
    \begin{tabular}{|l|c|c|c|c|c|c|c|c|c|c|c|}
        \hline
        \multicolumn{12}{|c|}{\textbf{Fashion-MNIST:} Ablation for \CovT and \CT}\\
        \hline
        Method & \multicolumn{3}{c|}{Baseline}
        & \multicolumn{2}{c|}{\Bel}
        & \multicolumn{2}{c|}{\CovT}
        & \multicolumn{4}{c|}{\CT}\\
        \hline
        Train & \multicolumn{3}{c|}{}
        & \multicolumn{2}{c|}{\ThrL}
        & \Thr & \ThrLP
        & \ThrLP & \ThrLP
        & \multicolumn{2}{c|}{+\Lclass}\\
        \hline
        Test & \ThrL & \Thr & \APS
        & \ThrL & \Thr
        & \Thr & \Thr
        & \Thr & \APS
        & \Thr & \APS\\
        \hline\hline
        \Ineff & 2.52 & 2.05 & 2.36 & 1.83 & 1.9 & 4.03 & 2.69 & 1.69 & 1.82 & 1.67 & 1.73\\
        \Acc & 89.16 & 89.16 & 89.16 & 84.29 & 84.61 & 89.23 & 87.48 & 88.86 & 87.43 & 89.23 & 88.69\\
        \hline
    \end{tabular}
    \vspace*{-0.1cm}
\end{table}
In \tabref{tab:app-fashion-mnist}, we present an ablation for \CovT, see \secref{sec:app-coverage-training}, and \CT on MNIST, using a linear model, and Fashion-MNIST, using a 2-layer MLP.
\Belc is generally able to improve inefficiency of \ThrL.
Using \Thr, however, \Belc worsens inefficiency on MNIST significantly, while improving slightly over the baseline on Fashion-MNIST.
As a result, the improvement of \CT over \Belc is also less significant on Fashion-MNIST.
Using \CovT with \Thr or \APS during training works poorly.
As described in \tabref{tab:app-trials}, this is mainly due to a high variation across training runs, \ie, individual models might work well, but training is not stable enough to get consistent improvements.
Thus, on MNIST, inefficiency for \CovT with \Thr and \APS is very high.
Moreover, on Fashion-MNIST, we were unable to train \CovT with \Thr and \APS.
Using \ThrLP, training with \CovT works and is reasonably stable, but does not improve over the baseline.
It does improve over \Belc on MNIST though.
As described in the main paper, we suspect the fixed threshold $\tau$ to be problematic.
Overall, however, only \CT is able to outperform the \Thr baseline on both datasets.
Here, \CT with \Lclass works slightly better than without.

\section{All Inefficiency Results}
\label{sec:app-results}

\begin{table}[t]
    \vspace*{-0.2cm}
    \caption{
    \textbf{Inefficiency and Accuracy on Multiclass Datasets:}
    Complementing \tabref{tab:results} in the main paper, we include results for \CovT on CIFAR10.
    Furthermore, we consider training a non-linear 2-layer MLP on the ResNet features on CIFAR10, \cf \secref{sec:app-experimental-setup}, alongside the ensemble results from the main paper.
    We report inefficiency and accuracy in all cases, focusing on \CT in comparison to \Belc.
    On EMNIST, we additionally consider $\alpha = 0.005, 0.001$ (for the baseline and \CT only).
    As in the main paper, \CT consistently improves inefficiency of \Thr and \APS.
    }
    \label{tab:app-multiclass}
    \vspace*{-6px}
    \centering
    \small
    \begin{tabular}{|l|c|c|c|c|c|c|c|c|c|c|c|}
        \hline
        \multicolumn{12}{|c|}{\textbf{CIFAR10:} Fine-Tuning and ``Extending''}\\
        \hline
        & \multicolumn{3}{c|}{}
        & \multicolumn{6}{c|}{Fine-tuning}
        & \multicolumn{2}{c|}{``Extend''}\\
        \hline
        Method
        & \multicolumn{3}{c|}{Baselines}
        & \Bel
        & \CovT
        & \multicolumn{4}{c|}{\CT}
        & \multicolumn{2}{c|}{\CT}\\
        \hline
        Train
        &&&
        & \ThrL
        & \ThrLP
        & \ThrLP & \ThrLP & \multicolumn{2}{c|}{+\Lclass}
        & \ThrLP & +\Lclass\\
        Test
        & \ThrL & \Thr & \APS
        & \Thr
        & \Thr
        & \Thr & \APS & \Thr & \APS
        & \Thr & \Thr\\
        \hline
        \hline
        \Ineff & 3.92 & 2.93 & 3.3 & 2.93 & 2.84 & 2.88 & 3.05 & 2.84 & 2.93 & 2.89 & 2.96\\
        \Acc & 82.6 & 82.6 & 82.6 & 82.18 & 82.36 & 82.32 & 82.34 & 82.4 & 82.4 & 82.3 & 82.23\\
        \hline
    \end{tabular}
    
    \begin{tabular}{|l|c|c|c|c|c|c|c|}
        \hline
        \multicolumn{8}{|c|}{\textbf{CIFAR10:} Ensemble Results}\\
        \hline
        Method
        & \multicolumn{3}{c|}{(Ensemble Models)}
        & \multicolumn{3}{c|}{Ensemble+MLP}
        & \begin{tabular}{@{}c@{}}Ensemble\\+\CT\end{tabular}\\
        \hline
        Train
        &&&
        &&&
        & \ThrLP\\
        Test
        & \ThrL & \Thr & \APS
        & \ThrL & \Thr & \APS
        & \Thr\\
        \hline
        \hline
        Avg. \Ineff & 4.19 & 3.1 & 3.48 & 3.12 & 2.4 & 2.77 & 2.35\\
        Best \Ineff & 3.74 & 2.84 & 3.17 & 3.0 & 2.33 & 2.71 & 2.3\\
        Avg. \Acc & 80.65 & 80.65 & 80.65 & 85.88 & 85.88 & 85.88 & 85.88\\
        Best \Acc & 82.58 & 82.58 & 82.58 & 86.01 & 86.01 & 86.01 & 86.02\\
        \hline
    \end{tabular}
    
    \begin{tabular}{|l|c|c|c|c|c|c|c|c|c|}
        \hline
        \multicolumn{10}{|c|}{\textbf{EMNIST}}\\
        \hline
        \hline
        Method
        & \multicolumn{3}{c|}{Baselines}
        & \multicolumn{2}{c|}{\Bel}
        & \multicolumn{4}{c|}{\CT}\\
        \hline
        Train
        &&&
        & \ThrL
        & \ThrL
        & \ThrLP & \ThrLP
        & \multicolumn{2}{c|}{+\Lclass}\\
        \hline
        Test
        & \ThrL & \Thr & \APS
        & \ThrL
        & \Thr
        & \Thr & \APS
        & \Thr & \APS\\
        \hline
        \hline
        \Ineff & 5.07 & 2.66 & 4.23 & 3.95 & 3.48 & 2.66 & 2.86 & 2.49 & 2.87\\
        \Ineff, $\alpha{=}0.005$ & 9.23 & 4.1 & 6.04 & -- & -- & 3.37 & -- & -- & --\\
        \Ineff, $\alpha{=}0.001$ & 23.89 & 15.73 & 19.33 & -- & -- & 13.65 & -- & -- & --\\
        \Acc & 83.79 & 83.79 & 83.79 & 80.69 & 80.69 & 77.1 & 77.43 & 77.49 & 78.09\\
        \hline
    \end{tabular}

    \begin{tabular}{|l|c|c|c|c|c|c|c|c|}
        \hline
        \multicolumn{9}{|c|}{\textbf{CIFAR100}}\\
        \hline
        \hline
        Method
        & \multicolumn{3}{c|}{Baselines}
        & \Bel
        & \multicolumn{4}{c|}{\CT}\\
        \hline
        Train
        &&&
        & \ThrL
        & \ThrLP & \ThrLP
        & \multicolumn{2}{c|}{+\Lclass}\\
        \hline
        Test
        & \ThrL & \Thr & \APS
        & \Thr
        & \Thr & \APS
        & \Thr & \APS\\
        \hline
        \hline
        \Ineff & 19.22 & 10.63 & 16.62 & 10.91 & 10.78 & 12.99 & 10.44 & 12.73 \\
        \Acc & 73.36 & 73.36 & 73.36 & 72.65 & 72.02 & 72.78 & 73.27 & 72.99\\ 
        \hline
    \end{tabular}
    \vspace*{-0.1cm}
\end{table}
\begin{table}[t]
    \vspace*{-0.2cm}
    \caption{
    \textbf{Inefficiency and Accuracy on Binary Datasets.}
    Experimental results on the binary datasets WineQuality, GermanCredit and Camelyon.
    While we include \APS on WineQuality, we focus on \Thr on GermanCredit and Camelyon due to slightly lower inefficiency.
    However, \ThrL, \Thr and \APS perform very similarly on all tested binary datasets.
    Generally, \CT does not improve significantly over the baseline.
    * On Camelyon, we report the best results without training trials as sub-sampling the $280$ training examples is prohibitively expensive.
    }
    \label{tab:app-binary}
    \vspace*{-6px}
    \centering
    \small
    \begin{tabular}{|l|c|c|c|c|c|c|c|c|c|}
        \hline
        \multicolumn{10}{|c|}{\textbf{WineQuality}}\\
        \hline
        \hline
        Method
        & \multicolumn{3}{c|}{Baselines}
        & \Bel
        & \CovT
        & \multicolumn{4}{c|}{\CT}\\
        \hline
        Train
        & \multicolumn{3}{c|}{}
        & \ThrL
        & \ThrLP
        & \ThrLP & \ThrLP
        & \multicolumn{2}{c|}{+\Lclass}\\
        \hline
        Test
        & \ThrL
        & \Thr
        & \APS
        & \Thr
        & \Thr
        & \Thr & \APS
        & \Thr & \APS\\
        \hline
        \hline
        \Ineff, $\alpha{=}0.01$ & 1.76 & 1.76 & 1.79 & 1.77 & 1.81 & 1.75 & 1.82 & 1.74 & 1.77\\
        \Ineff, $\alpha{=}0.05$ & 1.48 & 1.49 & 1.53 & 1.57 & 1.50 & 1.51 & -- & 1.52 & --\\
        \Acc & 82.82 & 82.82 & 82.82 & 71.3 & 81.5 & 73.8 & 74.24 & 73.91 & 73.91\\
        \hline
    \end{tabular}
    
    \begin{tabular}{|l|c|c|c|c|c|c|}
        \hline
        \multicolumn{7}{|c|}{\textbf{GermanCredit}}\\
        \hline
        \hline
        Method
        & \multicolumn{3}{c|}{Baselines}
        & \Bel
        & \multicolumn{2}{c|}{\CT}\\
        \hline
        Train
        &&&
        & \ThrL
        & \ThrLP & +\Lclass\\
        \hline
        Test
        & \ThrL & \Thr & \APS
        & \Thr
        & \Thr & \Thr\\
        \hline
        \hline
        \Ineff & 1.89 & 1.86 & 1.90 & 1.85 & 1.88 & 1.77\\
        \Acc & 74.4 & 74.4 & 74.4 & 72.35 & 72.81 & 69.5\\
        \hline
    \end{tabular}
    \begin{tabular}{|l|c|c|c|c|c|c|}
        \hline
        \multicolumn{7}{|c|}{\textbf{Camelyon}* $\alpha{=}0.05$}\\
        \hline
        \hline
        Method
        & \multicolumn{3}{c|}{Baselines}
        & \Bel
        & \multicolumn{2}{c|}{\CT}\\
        \hline
        Train
        &&&
        & \ThrL
        & \ThrLP & +\Lclass\\
        \hline
        Test
        & \ThrL & \Thr & \APS
        & \Thr
        & \Thr & \Thr\\
        \hline
        \hline
        Best \Ineff & 1.41 & 1.47 & 1.59 & 1.25 & 1.2 & 1.25\\
        Best \Acc & 88 & 88 & 88 & 92 & 91.5 & 85\\
        \hline
    \end{tabular}
    \vspace*{-0.1cm}
\end{table}
\tabref{tab:app-multiclass} shows complementary results for \CT on CIFAR10, EMNIST and CIFAR100.
For results on MNIST and Fashion-MNIST, see \tabref{tab:app-ablation}.
On CIFAR10, we also include \CT using a 2-layer MLP on top of ResNet features -- instead of the linear model used in the main paper.
In \tabref{tab:app-multiclass}, this is referred to as ``extending''.
However, inefficiency increases slightly compared to re-initializing and training just the (linear) logit layer.
This shows that the smaller inefficiency improvements on CIFAR shown in the main paper are not due to the linear model used, but rather caused by the features themselves.
We suspect that this is because the features are trained to optimize cross-entropy loss, leaving \CT less flexibility to optimize inefficiency.
In \tabref{tab:app-binary}, we consider three binary datasets, \ie, WineQuality, GermanCredit and Camelyon.
On binary datasets, \ThrL, \Thr and \APS perform very similar.
This already suggests that there is little room for inefficiency improvements.
Indeed, \CT is not able to improve inefficiency significantly.
However, this is partly due to our thorough evaluation scheme:
On Camelyon (using $\alpha{=}0.05$), we do \emph{not} report averages across all training trials, but the results corresponding to the best model.
This is because sub-sampling the training examples is unreasonable given that there are only $280$ of them.
Thus, Camelyon shows that \CT \emph{can} improve inefficiency.
On WineQuality or GermanCredit, however, this is ``hidden'' in reporting averages across 10 training runs.

\section{Effect of \CT on Class-Conditional Inefficiency and Coverage Confusion}
\label{sec:app-conformal-training}

\begin{figure}[t]
    \vspace*{-0.2cm}
    \centering
    \begin{minipage}[t]{0.32\textwidth}
        \vspace*{0px}
        
        \includegraphics[width=\textwidth]{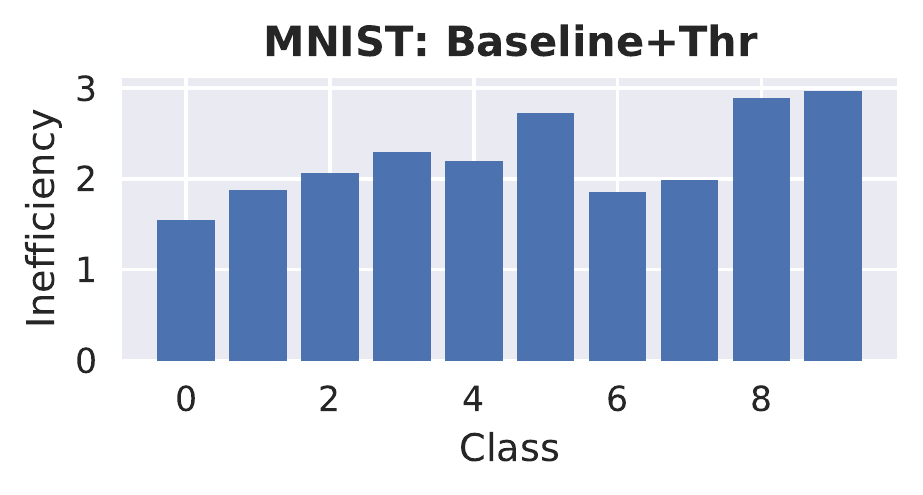}
        
        \includegraphics[width=\textwidth]{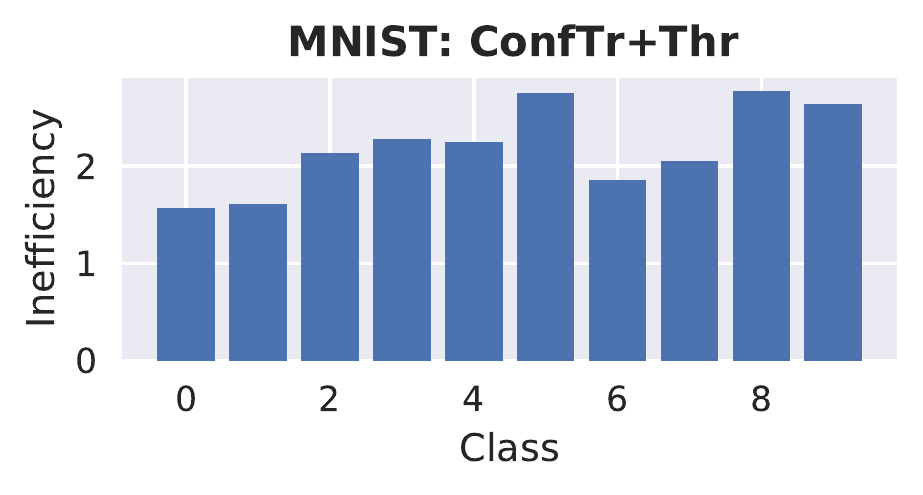}
    \end{minipage}
    \begin{minipage}[t]{0.32\textwidth}
        \vspace*{0px}
        
        \includegraphics[width=\textwidth]{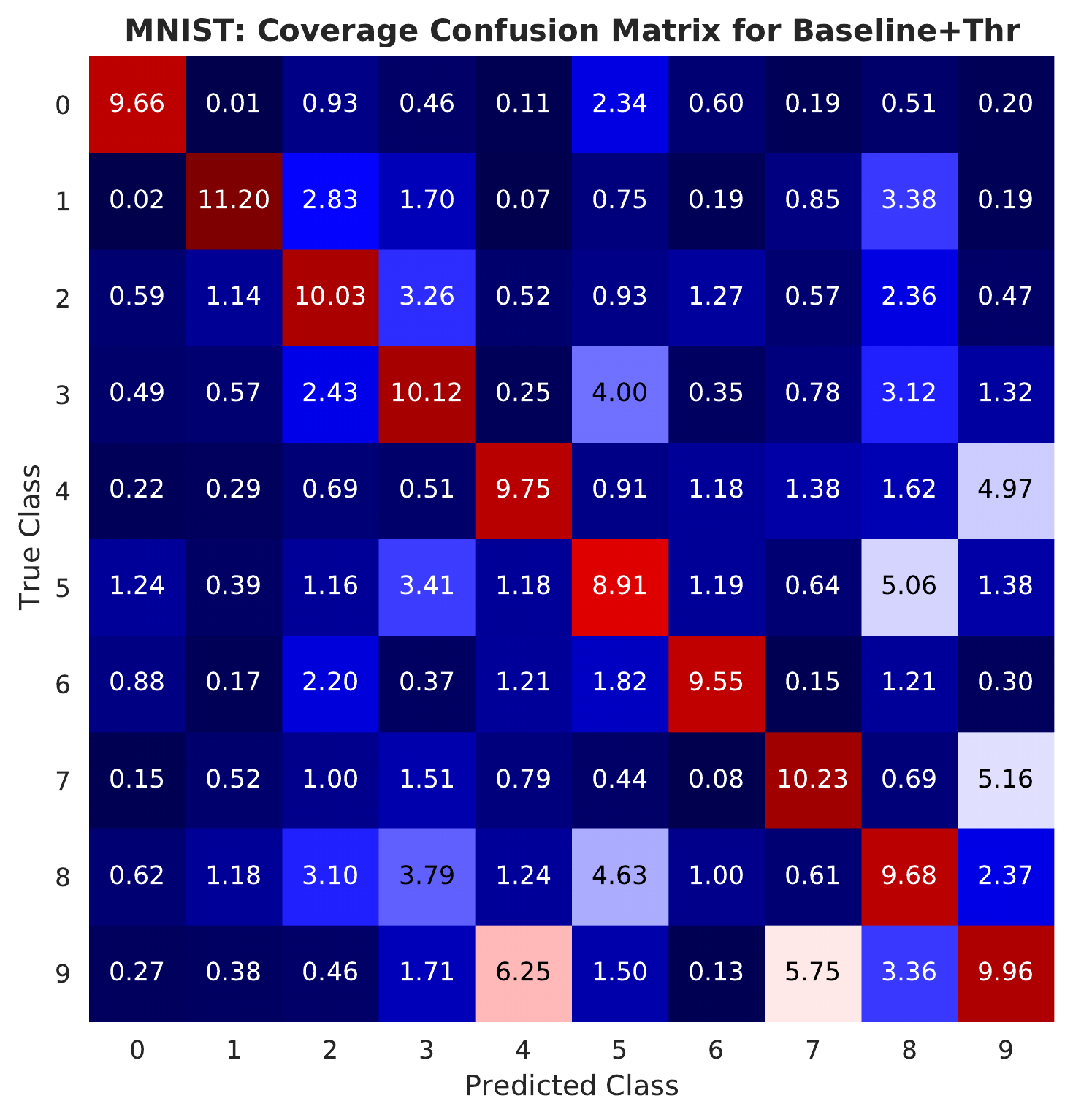}
    \end{minipage}
    \begin{minipage}[t]{0.32\textwidth}
        \vspace*{0px}
        
        \includegraphics[width=\textwidth]{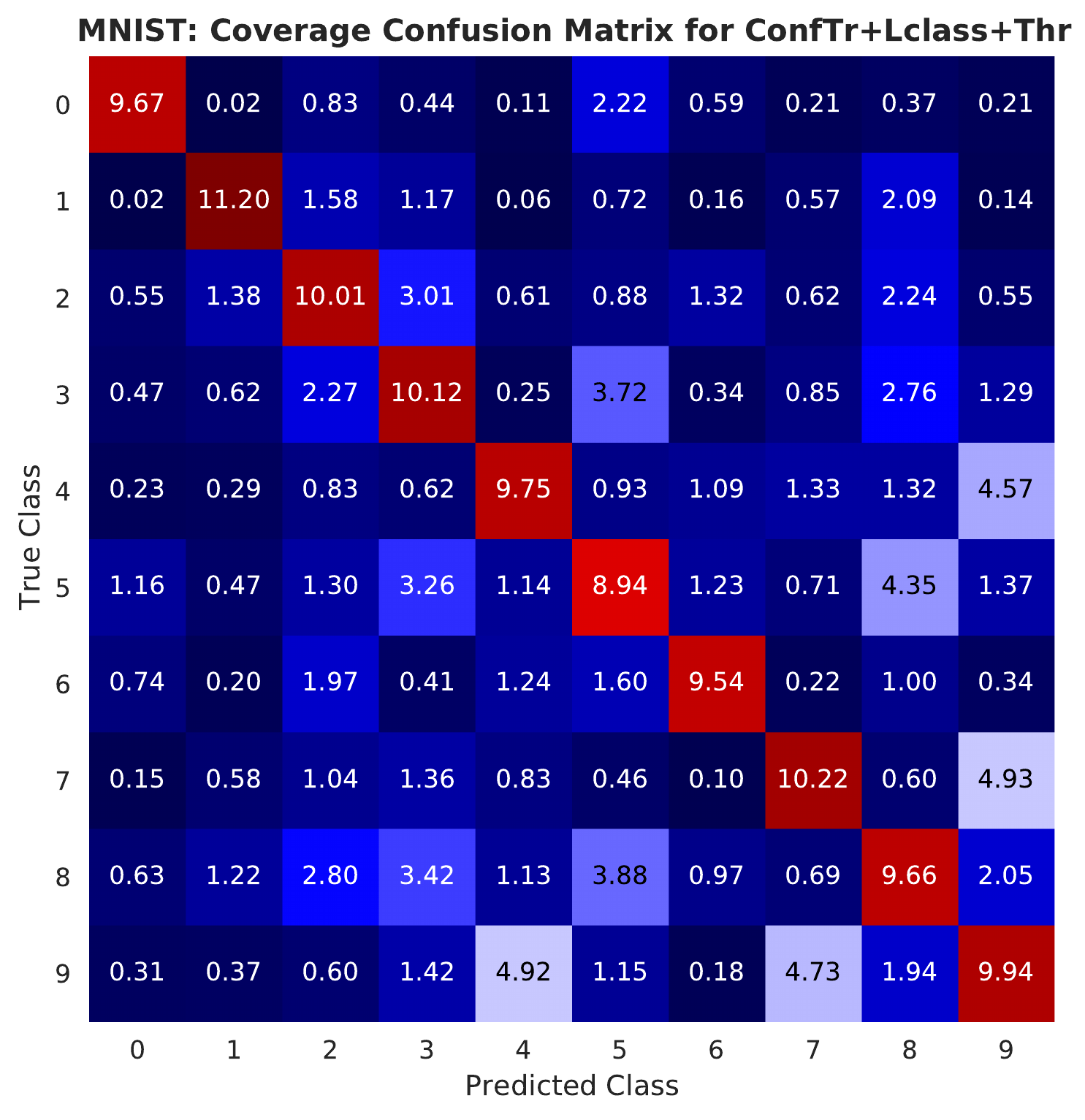}
    \end{minipage}
    
    \begin{minipage}[t]{0.32\textwidth}
        \vspace*{0px}
        
        \includegraphics[width=\textwidth]{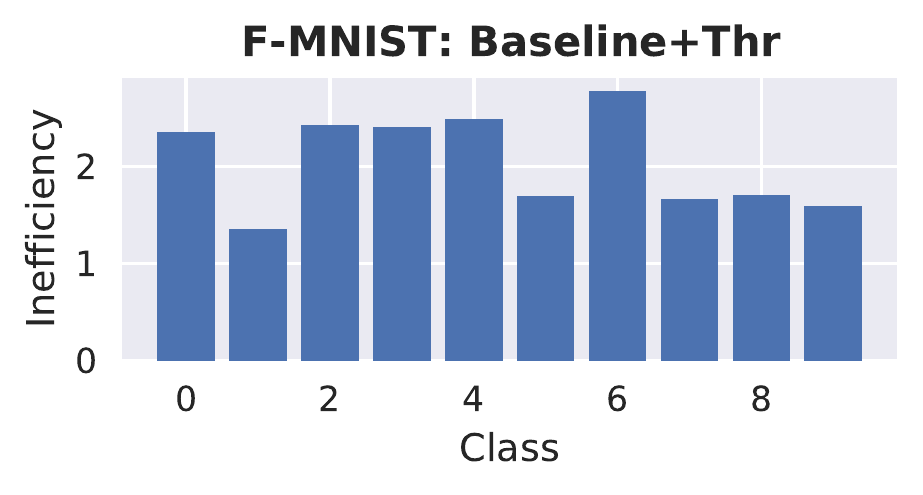}
        
        \includegraphics[width=\textwidth]{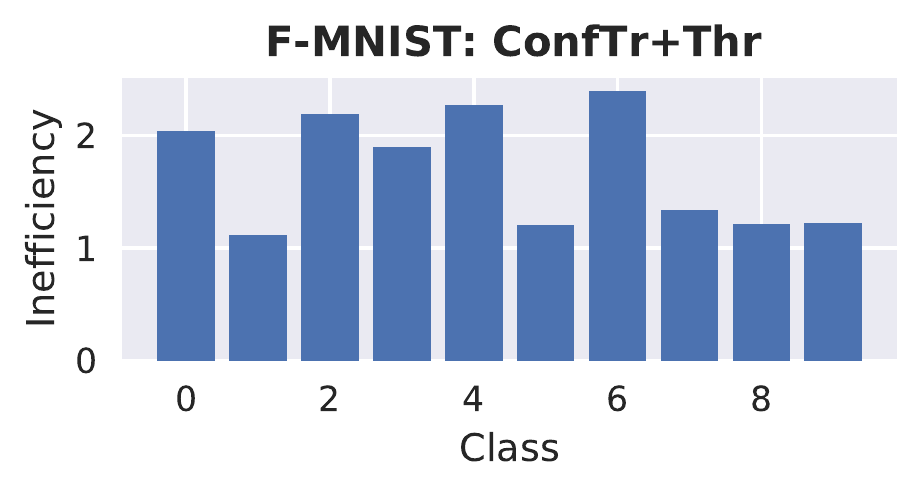}
    \end{minipage}
    \begin{minipage}[t]{0.32\textwidth}
        \vspace*{0px}
        
        \includegraphics[width=\textwidth]{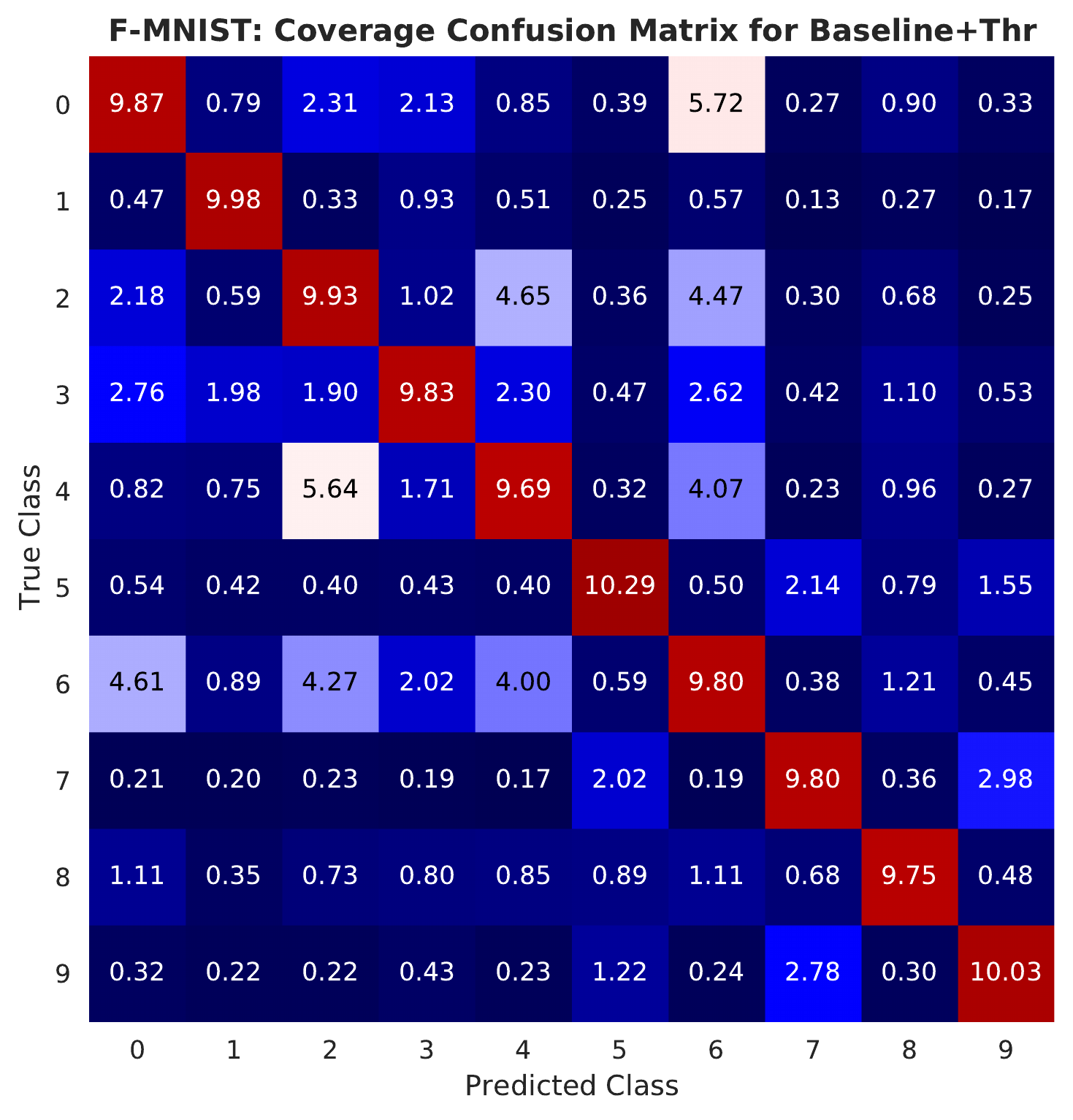}
    \end{minipage}
    \begin{minipage}[t]{0.32\textwidth}
        \vspace*{0px}
        
        \includegraphics[width=\textwidth]{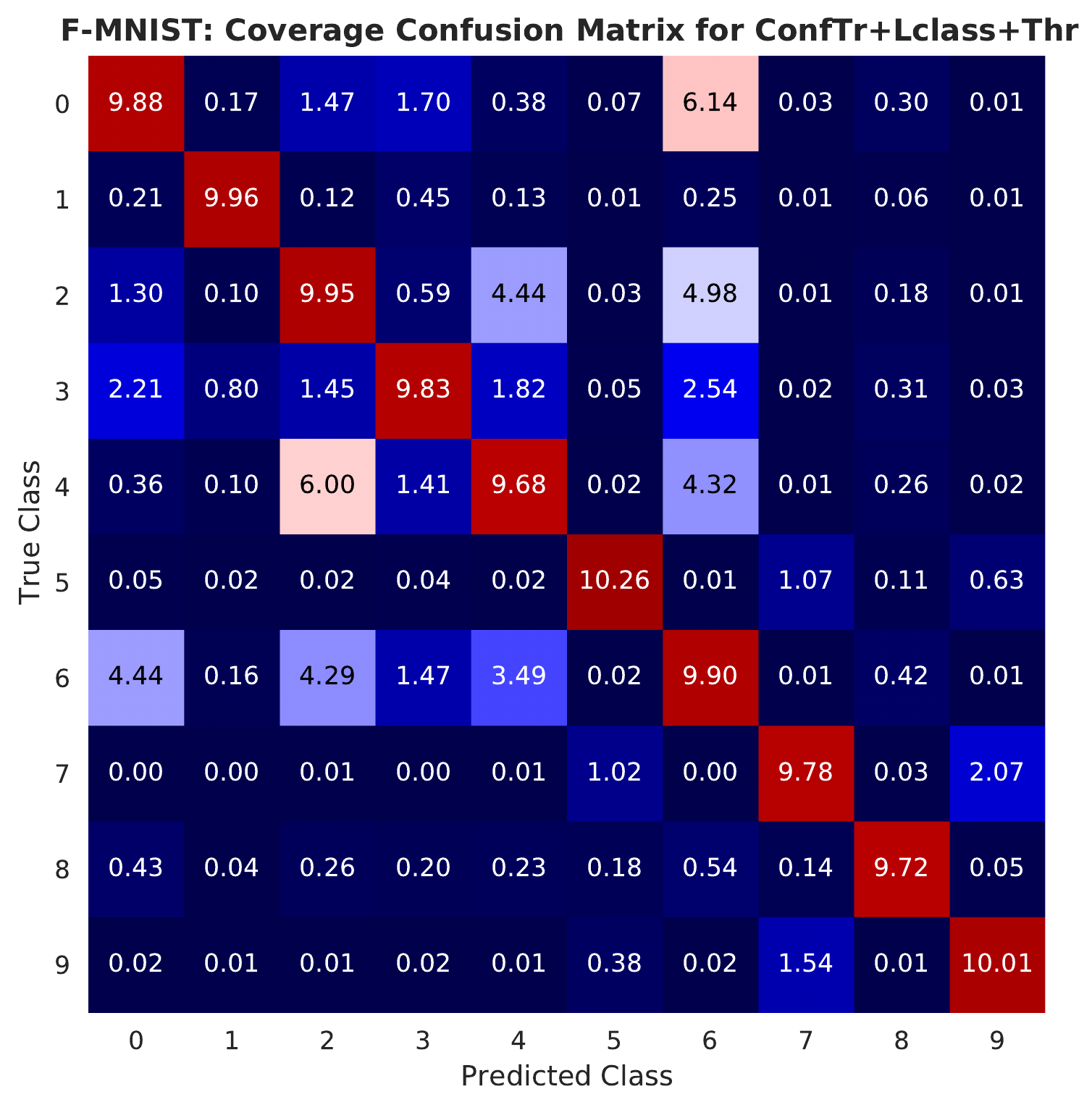}
    \end{minipage}
    
    \begin{minipage}[t]{0.32\textwidth}
        \vspace*{0px}
        
        \includegraphics[width=\textwidth]{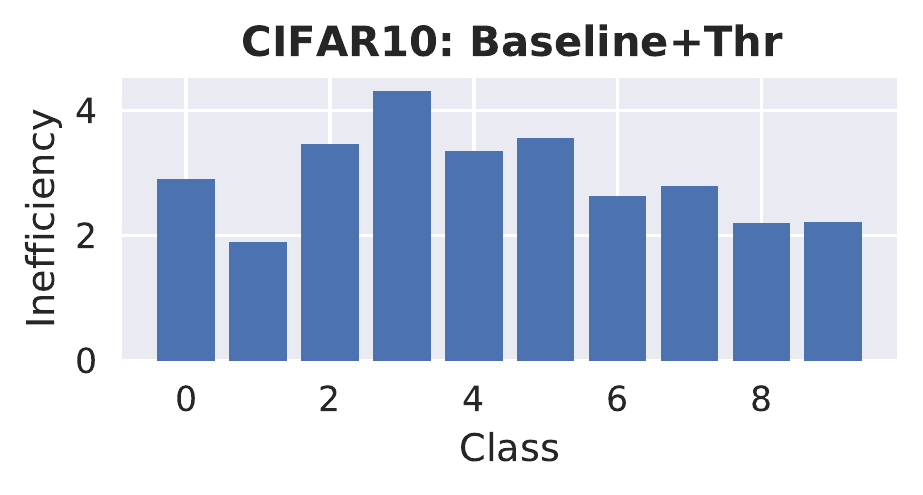}
        
        \includegraphics[width=\textwidth]{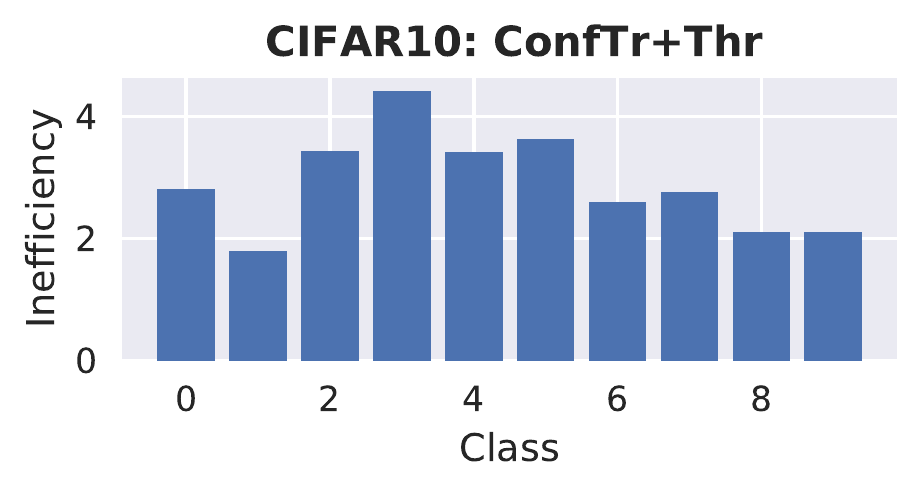}
    \end{minipage}
    \begin{minipage}[t]{0.32\textwidth}
        \vspace*{0px}
        
        \includegraphics[width=\textwidth]{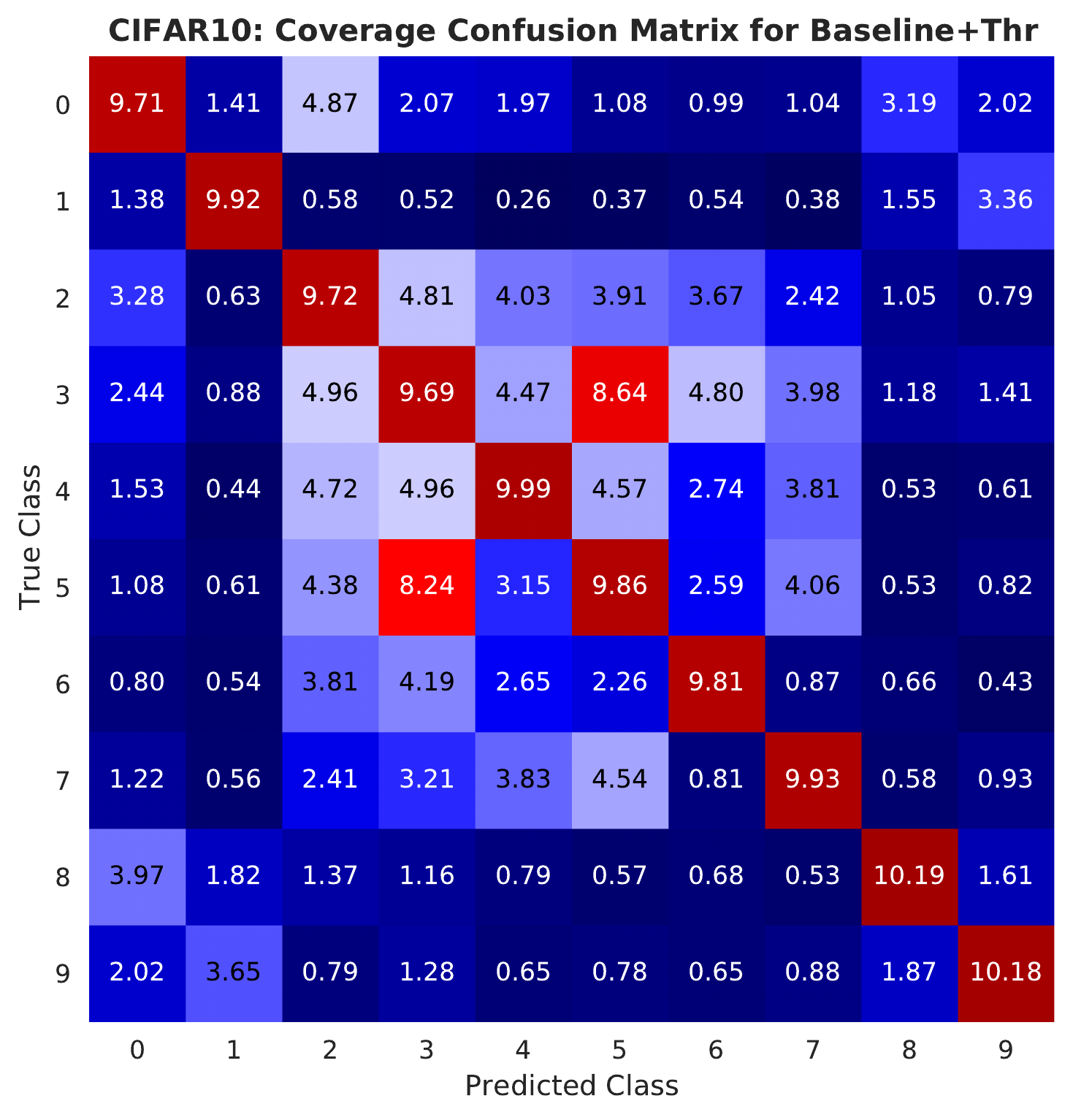}
    \end{minipage}
    \begin{minipage}[t]{0.32\textwidth}
        \vspace*{0px}
        
        \includegraphics[width=\textwidth]{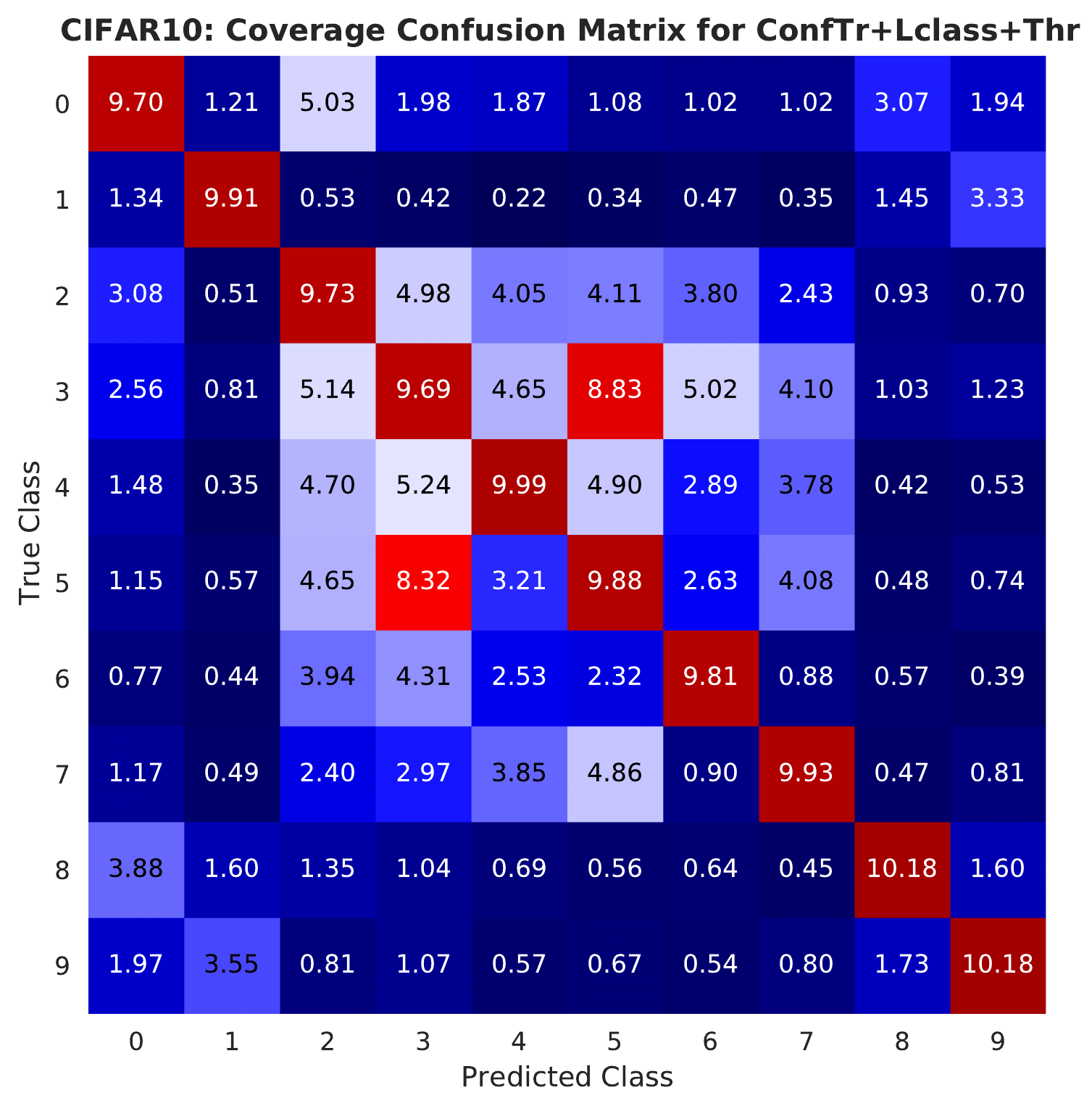}
    \end{minipage}
    \caption{
    \textbf{Class-Conditional Inefficiency and Coverage Confusion:}
    Comparison between baseline and \CT regarding class-conditional inefficiency and coverage confusion $\Sigma$, \cf \secref{subsec:conformal-training-applications}.
    For the inefficiency comparison, we consider \CT \emph{without} \Lclass, while for coverage confusion, \CT was trained with \Lclass.
    As \CT reduces overall inefficiency quite significantly on MNIST and Fashion-MNIST, class-conditional inefficiency is also lower, on average.
    But the distribution across classes remains similar.
    The same holds for coverage confusion, where lower overall inefficiency reduces confusion across the matrix, but the ``pattern'' remains roughly the same.
    On CIFAR10, \CT does not improve average inefficiency significantly, such that the confusion matrix remains mostly the same.
    }
    \label{fig:app-ct}
    \vspace*{-0.1cm}
\end{figure}

\figref{fig:app-ct} shows that standard \CT (without \Lclass) does not have a significant influence on the class-conditional inefficiency distribution compared to the baseline.
Similarly, \CT with \Lclass and identity loss matrix $L = I_K$ does not influence coverage confusion besides reducing overall inefficiency.
Specifically, on MNIST, Fashion-MNIST and CIFAR10, we show the class-conditional inefficiency distribution (left) as well as the coverage confusion matrices (middle and right) for the baseline and \CT.
On the left, we consider \CT without \Lclass, and on the right with \Lclass.
As can be seen, only an overall reduction of inefficiency is visible, the distribution of $\Ineff[y]$, \cf \eqnref{eq:conditional-ineff}, across classes $y$ remains roughly the same.
For coverage confusion $\Sigma$ from \eqnref{eq:coverage-confusion}, the same observation can be made, \ie, an overall reduction of inefficiency also reduces confusion, but the spatial pattern remains the same.
Thus, in the main paper and the following experiments, we always highlight the improvement over standard \CT, without \Lclass for reducing class-conditional inefficiency and with \Lclass for changing coverage confusion or improving \MisCov.

\section{Shaping Class-Conditional Inefficiency on Other Datasets}
\label{sec:app-ineff}

\begin{figure}[b]
    \vspace*{-0.1cm}
    \centering
    \begin{minipage}[t]{0.24\textwidth}
        \vspace*{-4px}
        
        \includegraphics[height=1.85cm]{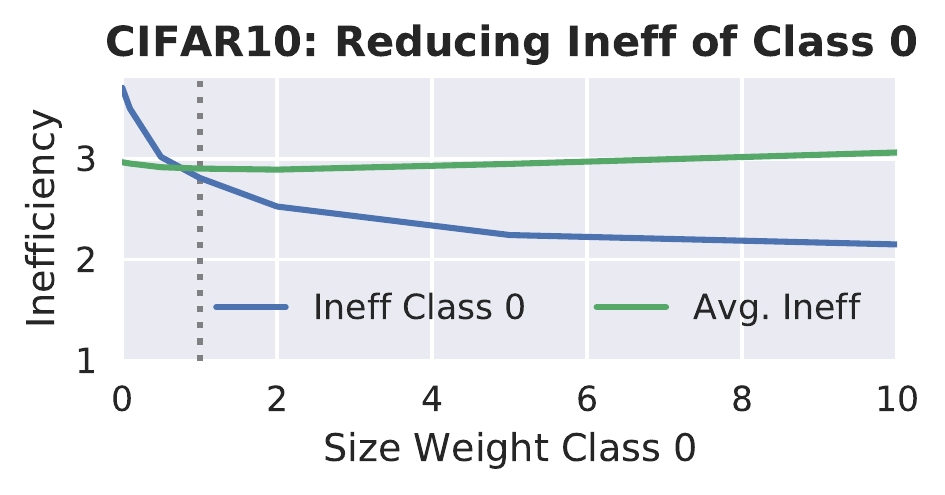}
    \end{minipage}
    \begin{minipage}[t]{0.24\textwidth}
        \vspace*{-4px}
        
        \includegraphics[height=1.85cm]{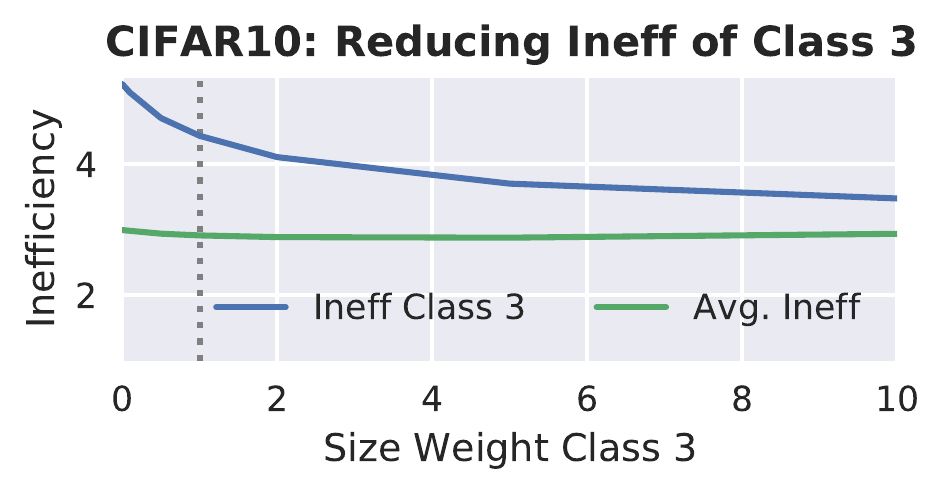}
    \end{minipage}
    \hskip 2px
    \begin{minipage}[t]{0.23\textwidth}
        \vspace*{0px}
        
        \begin{center}
            \hphantom{aaa}{\sffamily\fontsize{4}{5}\bfseries CIFAR100: Reducing \Ineff of Group 9}
        \end{center}
        \vspace*{-5px}
        
        \includegraphics[height=1.45cm,clip,trim={0 0 0 0.65cm}]{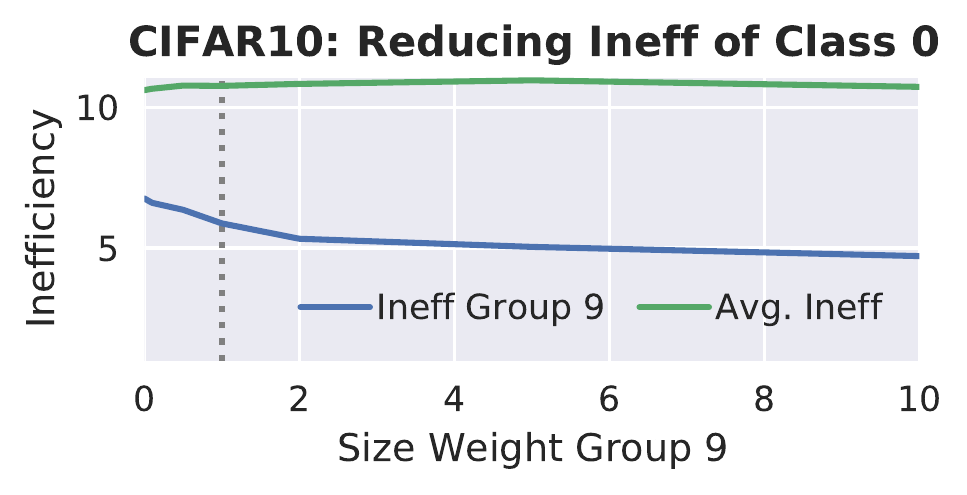}
    \end{minipage}
    \begin{minipage}[t]{0.23\textwidth}
        \vspace*{0px}
        
        \begin{center}
            \hphantom{aaa}{\sffamily\fontsize{4}{5}\bfseries CIFAR100: Reducing \Ineff of Group 15}
        \end{center}
        \vspace*{-5px}
        
        \includegraphics[height=1.45cm,clip,trim={0 0 0 0.65cm}]{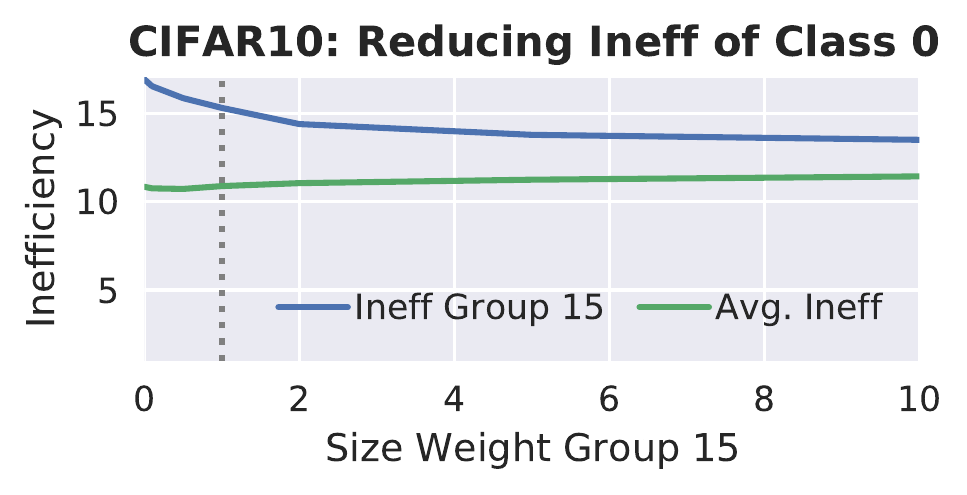}
    \end{minipage}
    \vspace*{-12px}
    \caption{
    \textbf{Reducing Class- and Group-Conditional Inefficiency on CIFAR.}
    Results, complementary to \figref{fig:size}, showing the impact of higher size weights $\omega$ in \eqnref{eq:size-loss} for classes 0 and 3 (``airplane'' and ``cat'') on CIFAR10 and coarse classes 9 and 15 (``large man-made outdoor things'' and ``reptiles'') on CIFAR100.
    \CT allows to reduce inefficiency ({\color{seaborn1}blue}) in all cases, irrespective of whether inefficiency is generally above or below average ({\color{seaborn2}green}).
    }
    \label{fig:app-size-1}
    \vspace*{0.2cm}
    \centering
    \begin{minipage}[t]{0.24\textwidth}
        \includegraphics[width=\textwidth]{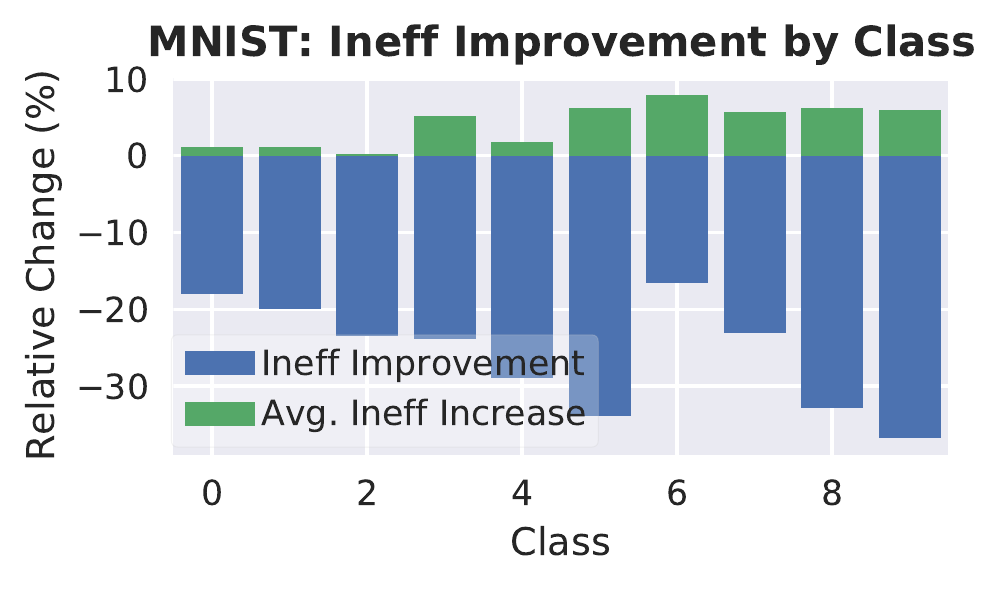}
    \end{minipage}
    \begin{minipage}[t]{0.24\textwidth}
        \includegraphics[width=\textwidth]{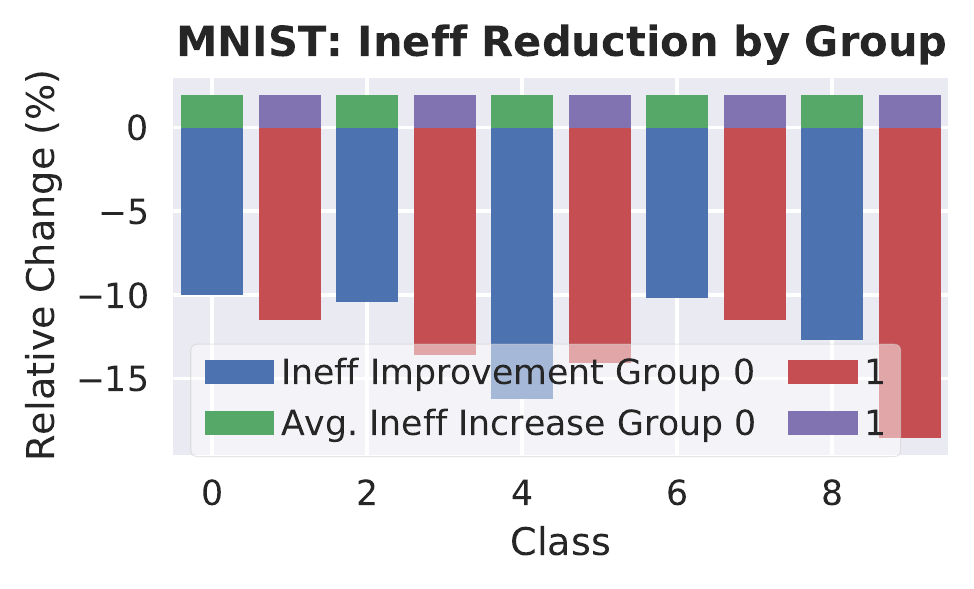}
    \end{minipage}
    \begin{minipage}[t]{0.24\textwidth}
        \includegraphics[width=\textwidth]{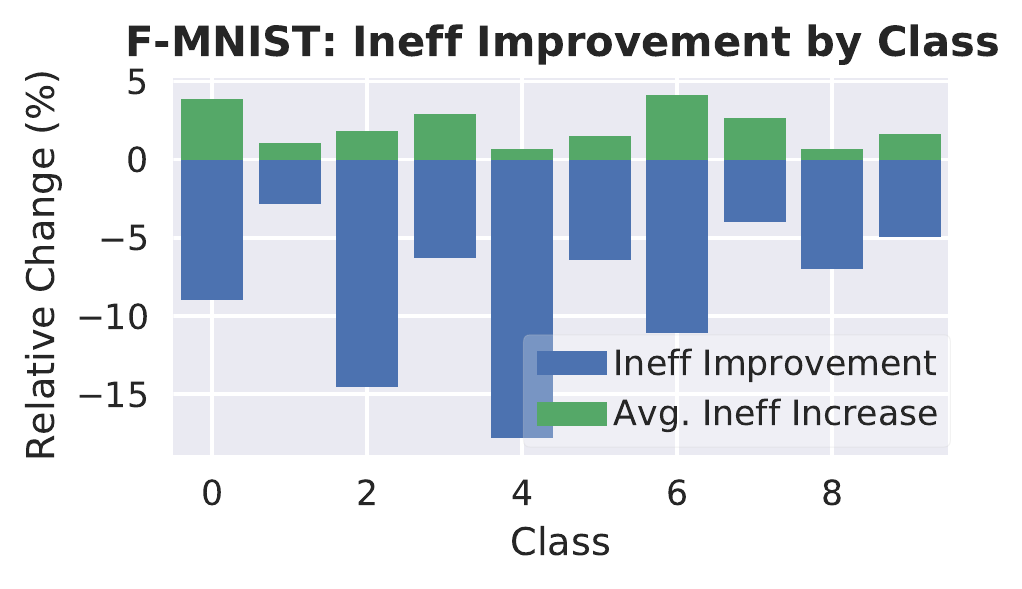}
    \end{minipage}
    \begin{minipage}[t]{0.24\textwidth}
        \includegraphics[width=\textwidth]{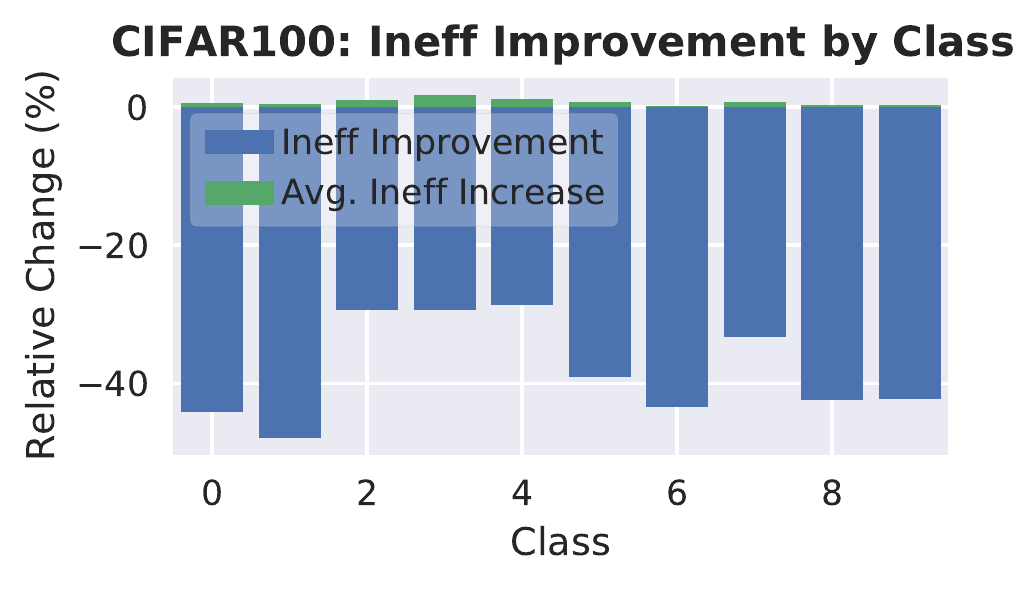}
    \end{minipage}
    \vspace*{-10px}
    \caption{
    \textbf{Relative Class and Group-Conditional Inefficiency Improvements:}
    Complementing the main paper, we plot the possible (relative) inefficiency reduction by class or group (``odd'' vs ``even'') on MNIST and Fashion-MNIST.
    On CIFAR100, we consider the first 10 classes for brevity.
    In all cases, significant per-class or -group inefficiency reductions are possible.
    }
    \label{fig:app-size-2}
    \vspace*{-0.2cm}
\end{figure}
\figref{fig:app-size-1} and \ref{fig:app-size-2} provide complementary results demonstrating the ability of \CT to shape the class- or group-conditional inefficiency distribution.
First, \figref{fig:app-size-1} plots inefficiency for individual classes on CIFAR10 and coarse classes on CIFAR100.
In both cases, significant inefficiency reductions are possible for high weights $\omega$ in \eqnref{eq:size-loss}, irrespective or whether the corresponding (coarse) class has above-average inefficiency to begin with.
This means that inefficiency reduction is possible for easier and harder classes alike.
Second, \figref{fig:app-size-2} plots the relative inefficiency changes, in percentage, possible per-class or group on MNIST, Fashion-MNIST and CIFAR100.
For CIFAR100, we show only the first 10 classes for brevity.
In all cases, significant inefficiency reductions are possible, at the expense of a slight increases in average inefficiency across all classes.
Here, MNIST is considerably easier than Fashion-MNIST: higher inefficiency reductions are possible per class and the cost in terms of average inefficiency increase is smaller.
On CIFAR100, inefficiency reductions of 40\% or more are possible.
This is likely because of the high number of classes, \ie, \CT has a lot of flexibility to find suitable trade-offs during training.

\section{Manipulating Coverage Confusion on Other Datasets}
\label{sec:app-confusion}

\begin{figure}[t]
    \vspace*{-0.1cm}
    \centering
    \begin{minipage}[t]{0.29\textwidth}
        \vspace*{0px}
        \includegraphics[height=4cm,clip,trim={0 0 0 0.8cm}]{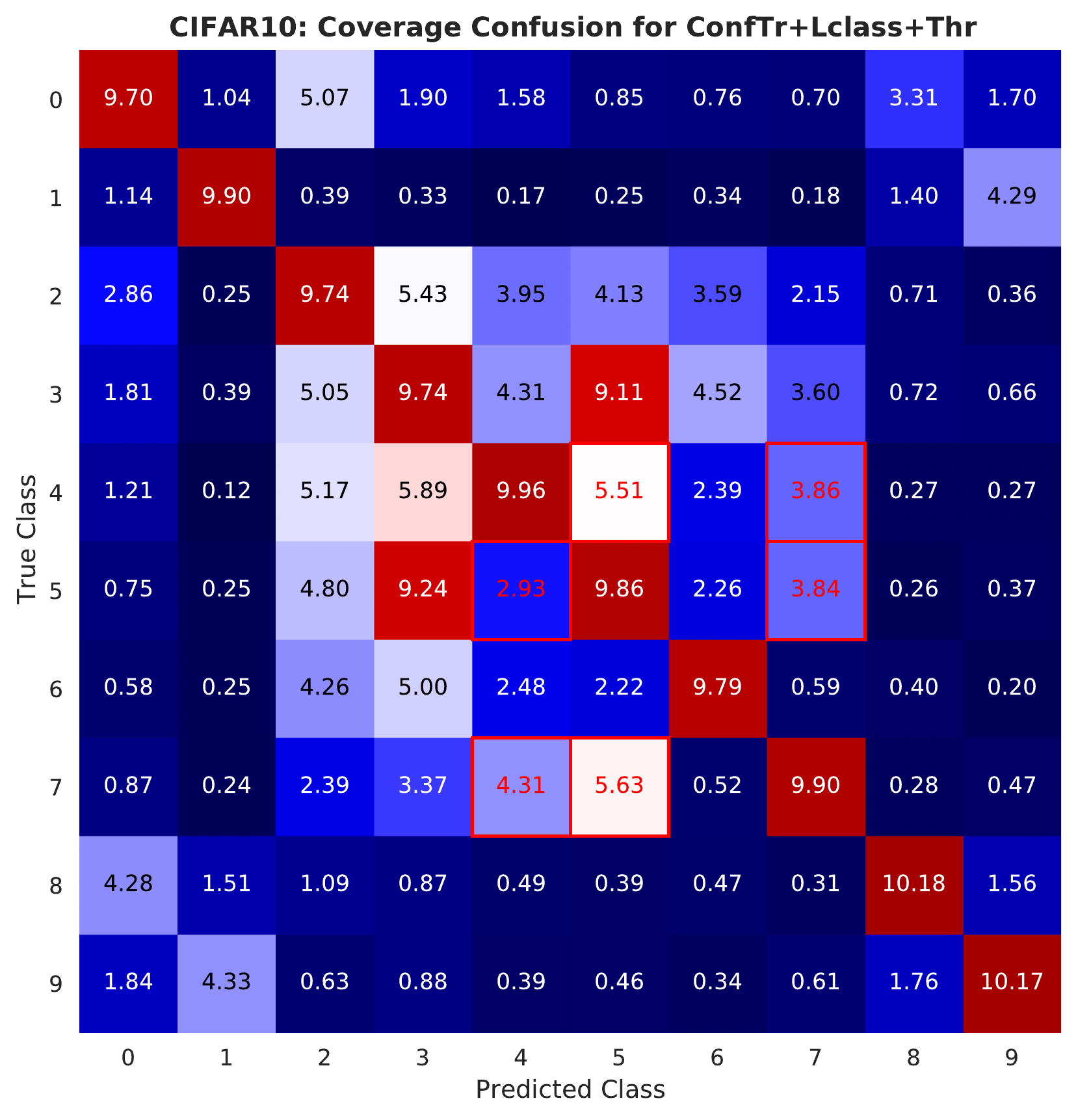}
    \end{minipage}
    \begin{minipage}[t]{0.29\textwidth}
        \vspace*{0px}
        
        \includegraphics[height=4cm,clip,trim={1cm 0 1cm 0.8cm}]{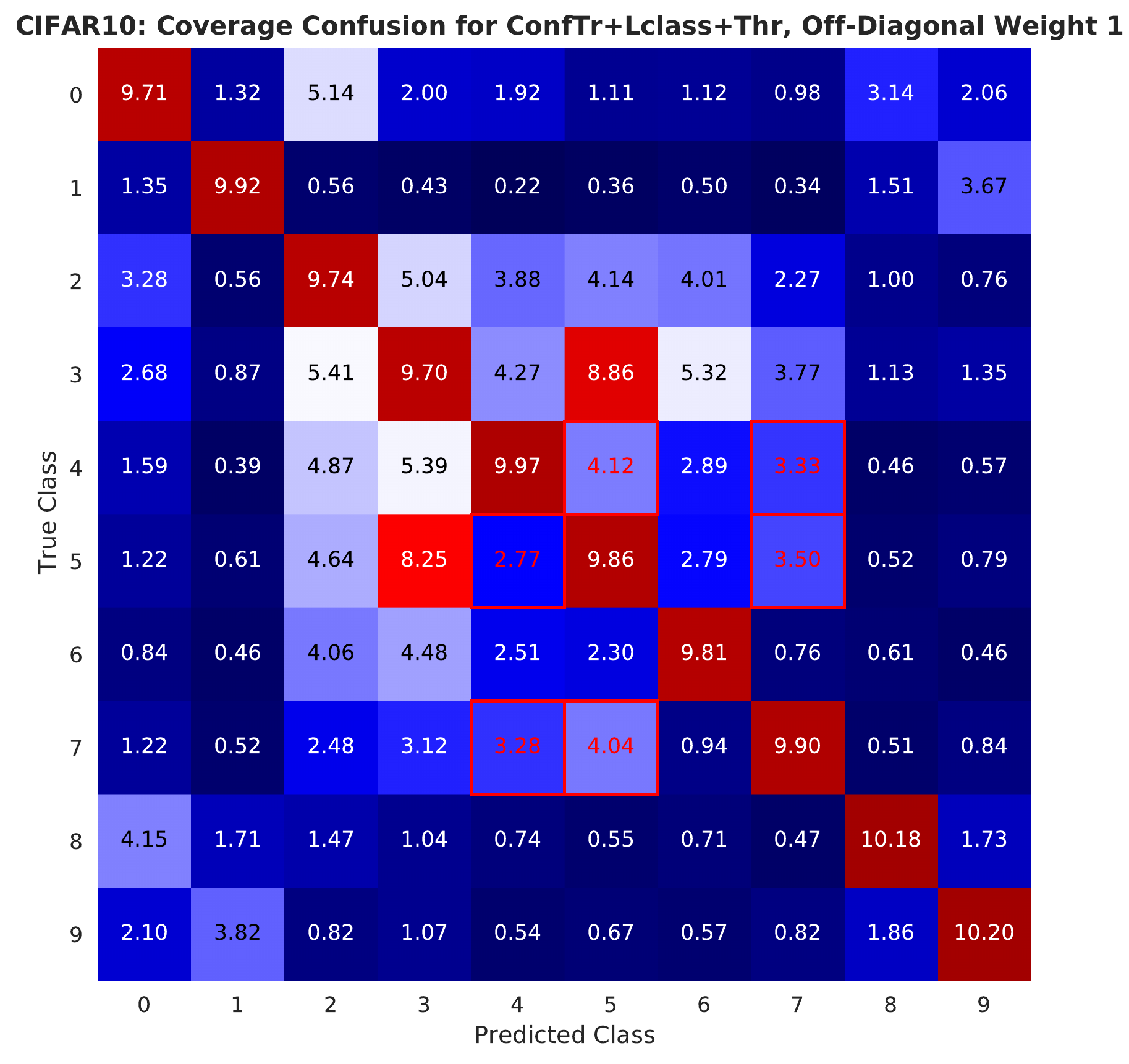}
    \end{minipage}
    \begin{minipage}[t]{0.38\textwidth}
        \vspace*{2px}
        
        \caption{
        \textbf{Full Coverage Confusion Matrix on CIFAR10:}
        We plot the full coverage confusion matrices $\Sigma$ from \eqnref{eq:coverage-confusion} on CIFAR10 for the \CT baseline (with \Lclass, left) and \CT with $L_{y,k} = 1$ in \eqnref{eq:classification-loss} for classes $y, k \in \{4, 5, 7\}$ (right, highlighted in {\color{red}red}).
        }
        \label{fig:app-confusion-1}
    \end{minipage}
    \vspace*{0.1cm}
    \centering
    
    \begin{minipage}[t]{0.01\textwidth}
        \vspace*{6px}
        
        \scriptsize
        \rotatebox{90}{CIFAR10\hskip 10px F-MNIST}
    \end{minipage}
    \begin{minipage}[t]{0.32\textwidth}
        \vspace*{0px}
        
        \raisebox{1.5px}{\includegraphics[height=0.7cm]{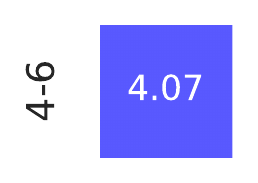}}
        \hskip -4px
        \includegraphics[height=0.75cm,clip,trim={0 1.125cm 0 0}]{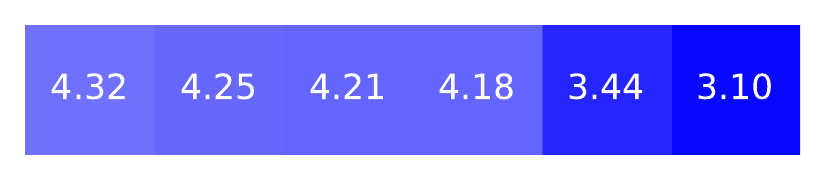}
        \vskip -4px

        \raisebox{1.5px}{\includegraphics[height=0.7cm]{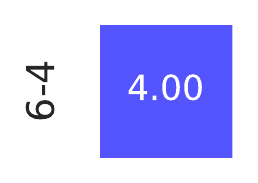}}
        \hskip -4px
        \includegraphics[height=0.75cm,clip,trim={0 1.125cm 0 0}]{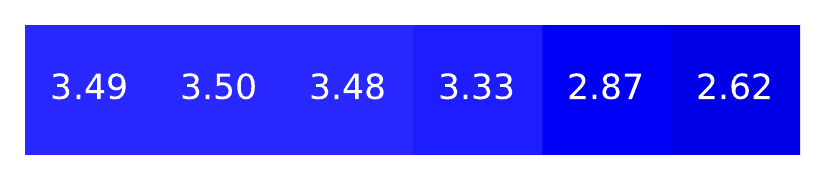}
        \vskip -4px
        
        \raisebox{1.5px}{\includegraphics[height=0.7cm]{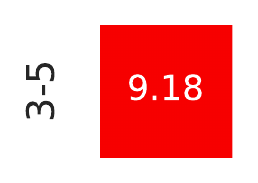}}
        \hskip -4px
        \includegraphics[height=0.75cm,clip,trim={0 1.125cm 0 0}]{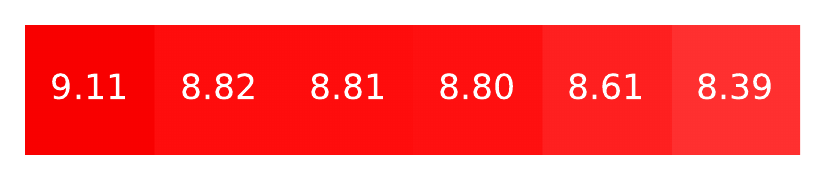}
        \vskip -4px

        \raisebox{4.75mm}{\includegraphics[height=0.7cm]{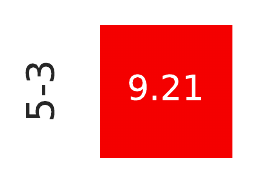}}
        \hskip -4px
        \includegraphics[height=1.225cm,clip,trim={0 0 0 0}]{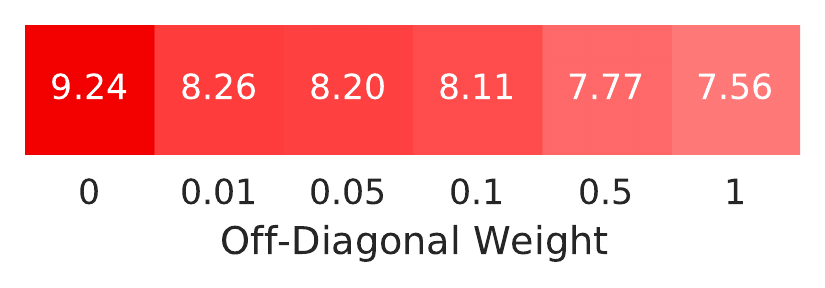}
    \end{minipage}
    \begin{minipage}[t]{0.32\textwidth}
        \vspace*{0px}
        
        \includegraphics[height=3cm]{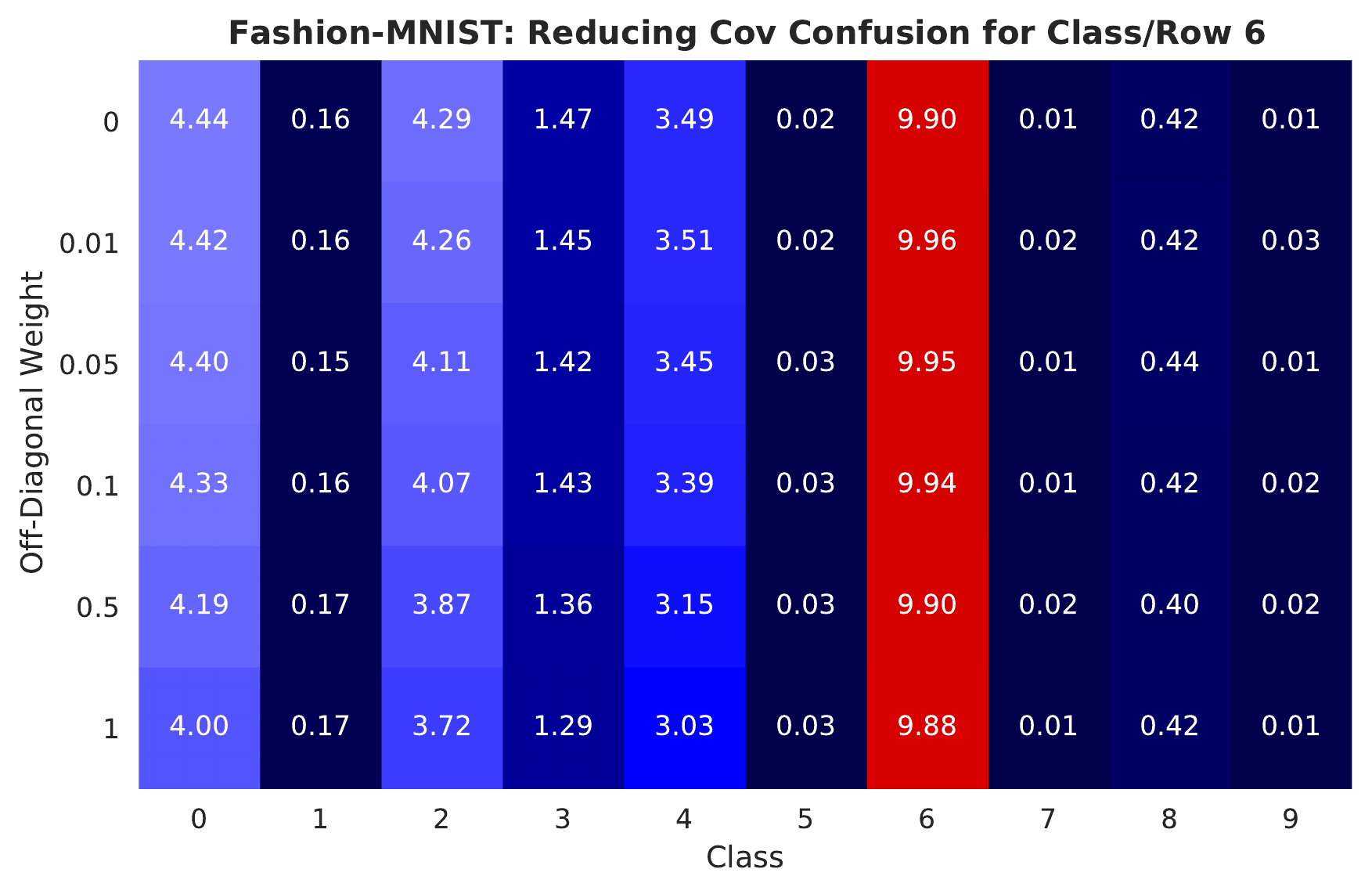}
    \end{minipage}
    \begin{minipage}[t]{0.32\textwidth}
        \vspace*{0px}
        
        \includegraphics[height=3cm]{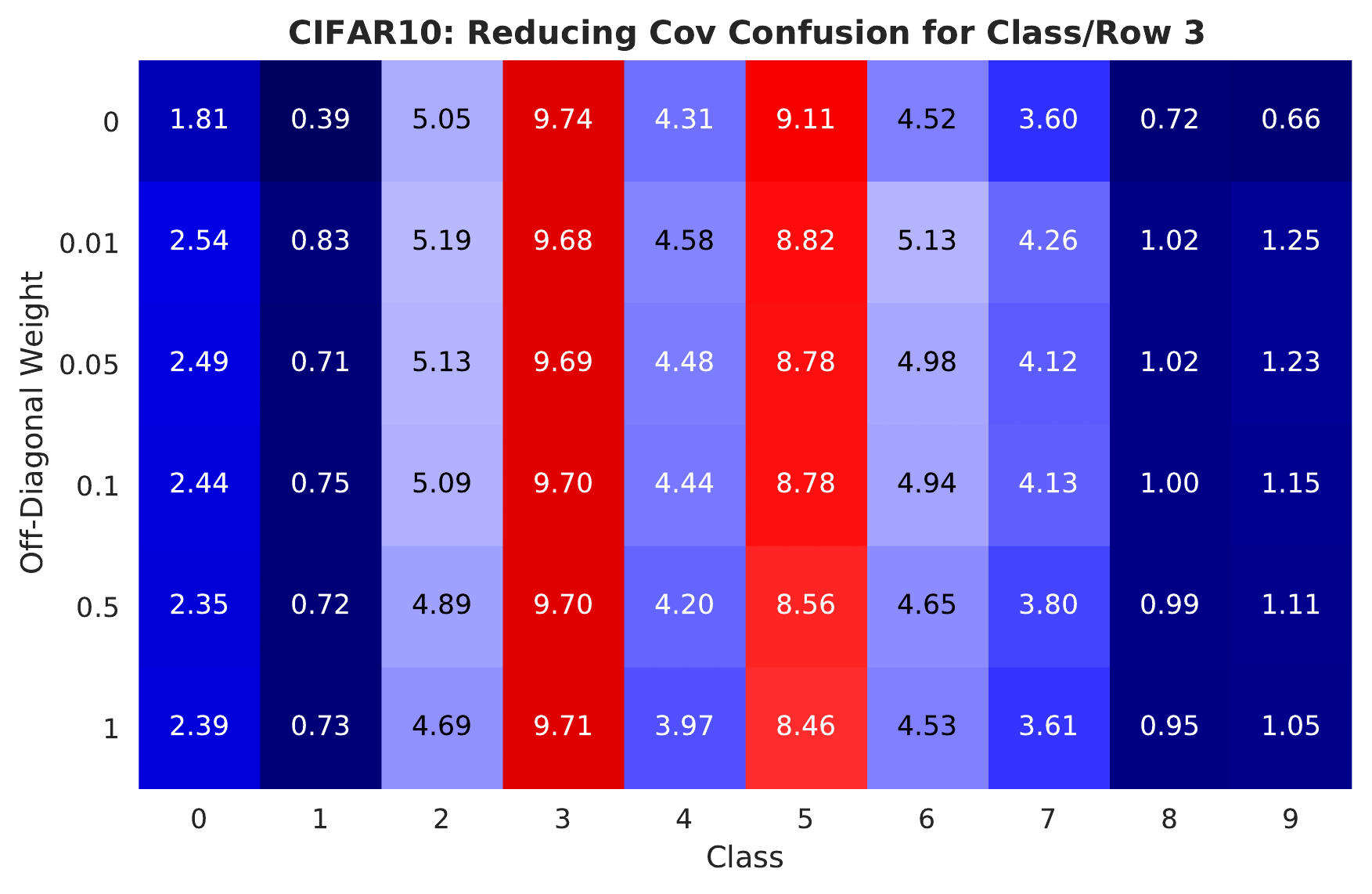}
    \end{minipage}
    \vspace*{-10px}
    \caption{
    \textbf{Coverage Confusion Changes on Fashion-MNIST and CIFAR10:}
    \textit{Left:}
    coverage confusion change when targeting classes 4 and 6 (``coat'' and ``shirt'') on Fashion-MNIST and 3 and 5 (``cat'' and ``dog'') on CIFAR10.
    The separate cell on the left is the \CT baseline which is, up to slight variations, close to $L_{y,k} = 0$.
    \textit{Middle and right:}
    coverage confusion for a whole row, \ie, $\Sigma_{y,k}$ with fixed class $y$ and all $k \neq y$.
    We show row 6 on Fashion-MNIST and 3 on CIFAR10.
    In both cases, coverage confusion can be reduced significantly.
    }
    \label{fig:app-confusion-2}
    \vspace*{-0.1cm}
\end{figure}

\begin{figure}[b]
    \vspace*{-0.1cm}
    \centering
    \begin{minipage}[t]{0.23\textwidth}
        \vspace*{0px}
        
        \includegraphics[height=2cm]{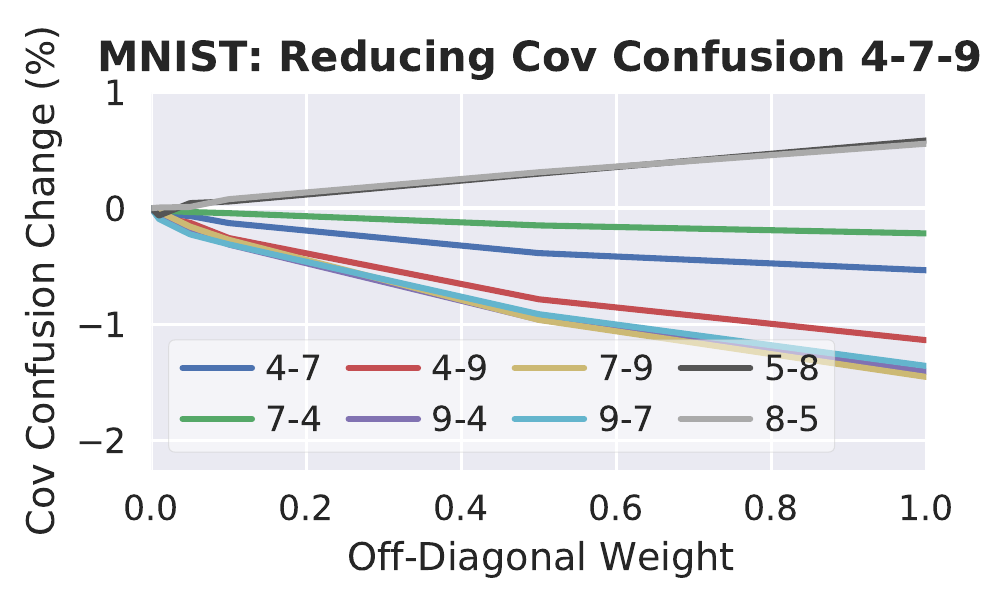}
    \end{minipage}
    \begin{minipage}[t]{0.235\textwidth}
        \vspace*{0px}
        
        \includegraphics[height=2cm]{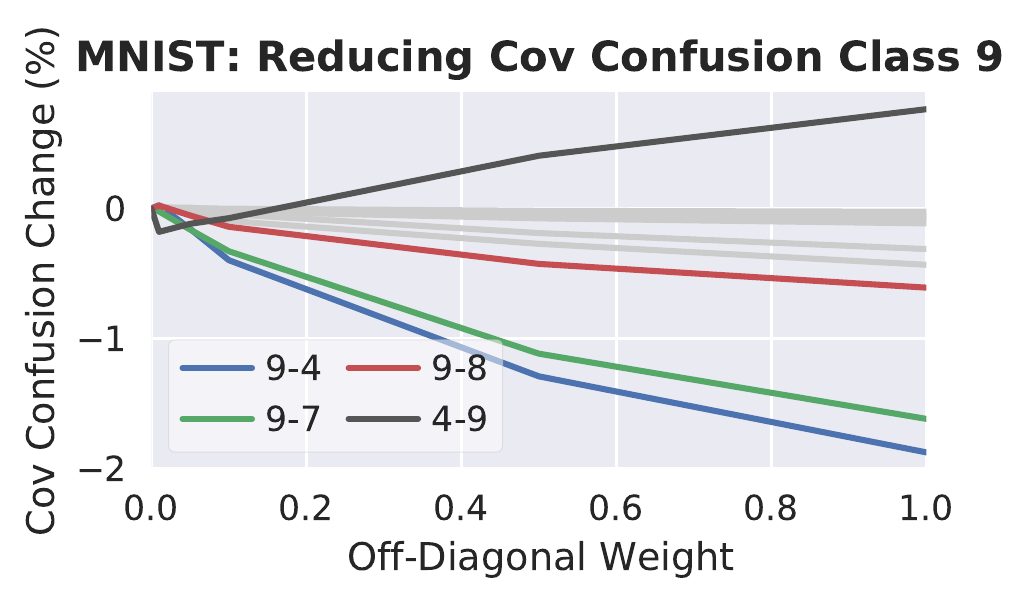}
    \end{minipage}
    \begin{minipage}[t]{0.25\textwidth}
        \vspace*{0px}
        
        \includegraphics[height=2cm]{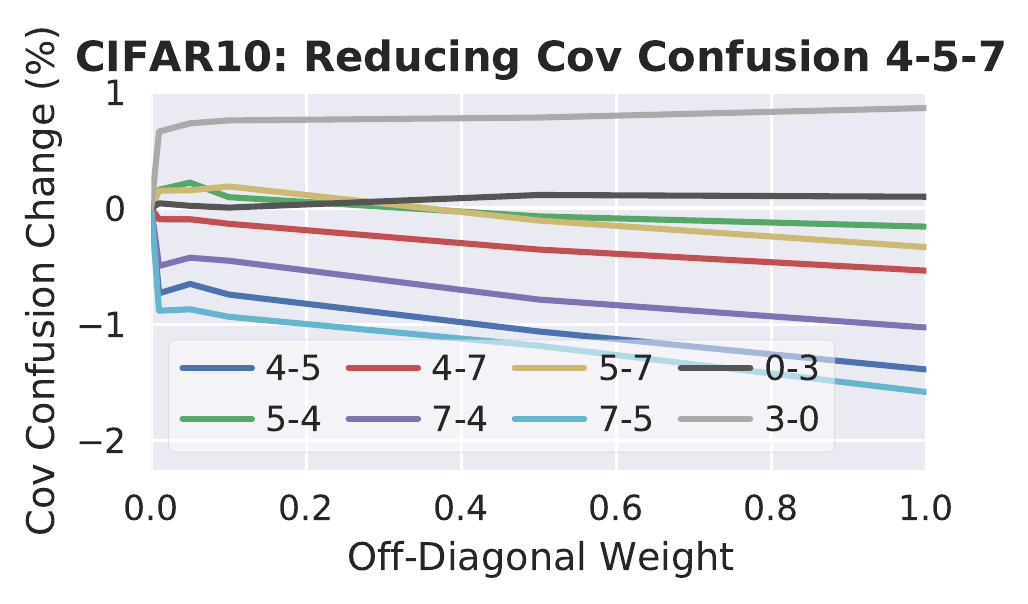}
    \end{minipage}
    \begin{minipage}[t]{0.25\textwidth}
        \vspace*{0px}
        
        \includegraphics[height=2cm]{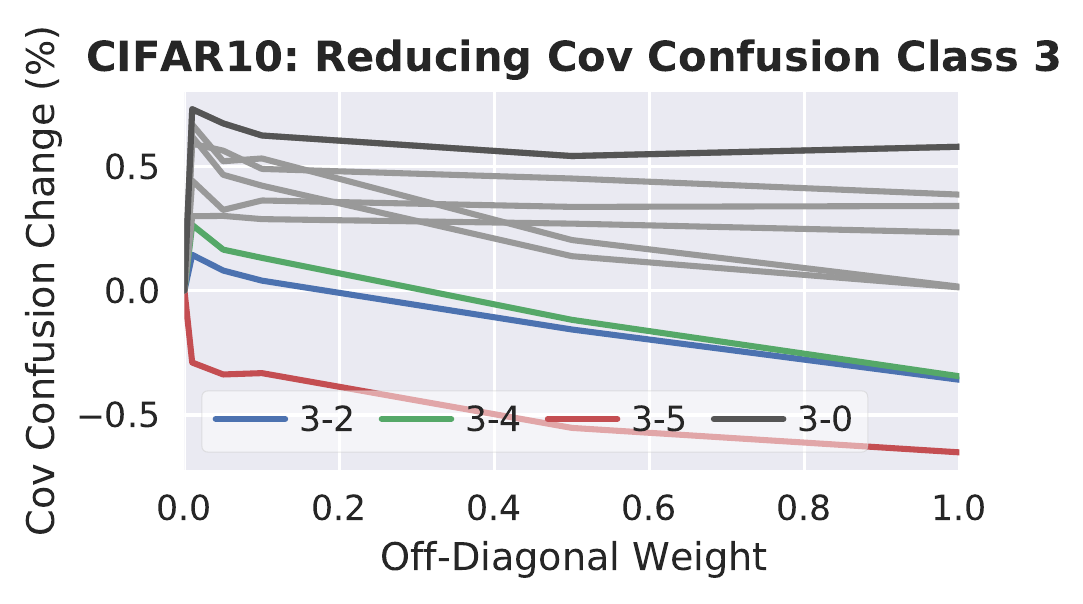}
    \end{minipage}
    \vspace*{-10px}
    \caption{
    \textbf{Coverage Confusion Reduction on MNIST and CIFAR10:}
    Controlling coverage confusion for various class pairs.
    On MNIST, coverage confusion reduction is usually more significant and the reduction scales roughly linear with the weight $L_{y,k}$.
    On CIFAR10, in contrast, coverage confusion cannot always be reduced for multiple class pairs at the same time (see {\color{gray!65!white}light gray}).
    }
    \label{fig:app-confusion-3}
    \vspace*{-0.2cm}
\end{figure}

\figref{fig:app-confusion-1} to \ref{fig:app-confusion-3} provide additional results for reducing coverage confusion using \CT.
First, in \figref{fig:app-confusion-1} we show the full coverage confusion matrices for the \CT baseline (with \Lclass, left) and \CT with $L_{y,k} = 1$, $y \neq k \in \{4,5,7\}$ (right, marked in {\color{red}red}) on CIFAR10.
This allows to get the complete picture of how coverage confusion changes and the involved trade-offs.
As demonstrated in the main paper, coverage confusion for, \eg, classes 4 and 5 (``deer'' and ``dog'') reduces.
However, coverage confusion for other class pairs might increase slightly.
Then, supplementary to \figref{fig:confusion} in the main paper, we provide the actual numbers in \figref{fig:app-confusion-2}.
In particular, we visualize how the actual coverage confusion entries (left) or rows (right) change depending on the off-diagonal weights $L_{y,k}$.
Finally, \figref{fig:app-confusion-3} presents additional results on MNIST and CIFAR10.
From these examples it can be seen that reducing coverage confusion is easier on MNIST, reducing linearly with the corresponding penalty $L_{y,k}$.
Moreover, the achieved reductions are more significant.
On CIFAR10, in contrast, coverage confusion reduces very quickly for small $L_{y,k}$ before stagnating for larger $L_{y,k}$.
At the same time, not all targeted class pairs might yield significant coverage confusion reductions.

\section{\MisCov Results on Additional Datasets}
\label{sec:app-miscoverage}

\begin{table}[t]
    \caption{
    \textbf{Mis-Coverage on MNIST, Fashion-MNIST and CIFAR10:}
    We present inefficiency and mis-coverage for various cases:
    On MNIST, we consider 2 vs. other classes as well as even vs. odd classes.
    In both cases, mis-coverage can be reduced significantly.
    As in the main paper, however, reducing $\MisCov_{0\rightarrow1}$ usually increases $\MisCov_{1\rightarrow0}$ and vice-versa.
    On Fashion-MNIST, we consider 6 (``shirt'') vs. other classes.
    Only on CIFAR10, considering ``vehicles'' vs. ``animals'', mis-coverage cannot be reduced significantly.
    In particular, we were unable to reduce $\MisCov_{1\rightarrow0}$.
    }
    \label{tab:app-miscoverage}
    \vspace*{-6px}
    \centering
    \small
    \begin{minipage}[t]{0.305\textwidth}
        \vspace*{0px}
        
        \small
        \begin{tabular}[t]{|l|c|c|c|}
            \hline
            \multicolumn{4}{|c|}{$K_0{=}$ 2 vs.  $K_1{=}$ Others}\\
            \hline
            MNIST && \multicolumn{2}{c|}{\MisCov $\downarrow$}\\
            \hline
            Method & \Ineff & $0{\rightarrow}1$ & $1{\rightarrow}0$\\
            \hline
            \hline
            \CT & 2.11 & 49.68 & 14.74\\
            $L_{K_0, K_1}{=}1$ & 2.15 & 36.63 & 17.42\\
            $L_{K_1, K_0}{=}1$ & 2.09 & 51.54 & 7.62\\
            \hline
        \end{tabular}
    \end{minipage}
    \begin{minipage}[t]{0.34\textwidth}
        \vspace*{0px}
        
        \small
        \begin{tabular}[t]{|l|c|c|c|}
            \hline
            \multicolumn{4}{|c|}{$K_0{=}$ Even vs. $K_1{=}$ Odd}\\
            \hline
            MNIST && \multicolumn{2}{c|}{\MisCov $\downarrow$}\\
            \hline
            Method & \Ineff & $0{\rightarrow}1$ & $1{\rightarrow}0$\\
            \hline
            \hline
            \CT & 2.11 & 38.84 & 38.69\\
            $L_{K_0, K_1}{=}1$ & 2.16 & 29.36 & 49.08\\
            $L_{K_1, K_0}{=}1$ & 2.09 & 44.3 & 26.08\\
            \hline
        \end{tabular}
    \end{minipage}
    
    \begin{minipage}[t]{0.32\textwidth}
        \vspace*{0px}
        
        \small
        \begin{tabular}[t]{|l|c|c|c|}
            \hline
            \multicolumn{4}{|c|}{$K_0{=}$ 6 (``shirt'') vs.  $K_1{=}$ Others}\\
            \hline
            F-MNIST && \multicolumn{2}{c|}{\MisCov $\downarrow$}\\
            \hline
            Method & \Ineff & $0{\rightarrow}1$ & $1{\rightarrow}0$\\
            \hline
            \hline
            \CT & 1.67 & 80.28 & 20.93\\
            $L_{K_0, K_1}{=}1$ & 1.70 & 72.58 & 25.81\\
            $L_{K_1, K_0}{=}1$ & 1.72 & 81.18 & 17.66\\
            \hline
        \end{tabular}
    \end{minipage}
    \begin{minipage}[t]{0.36\textwidth}
        \vspace*{0px}
        
        \small
        \begin{tabular}[t]{|l|c|c|c|}
            \hline
            \multicolumn{4}{|c|}{$K_0{=}$ ``vehicles'' vs. $K_1{=}$ ``animals''}\\
            \hline
            CIFAR10 && \multicolumn{2}{c|}{\MisCov $\downarrow$}\\
            \hline
            Method & \Ineff & $0{\rightarrow}1$ & $1{\rightarrow}0$\\
            \hline
            \hline
            \CT & 2.84 & 22.22 & 16.45\\
            $L_{K_0, K_1}{=}1$ & 2.92 & 20.00 & 22.69\\
            $L_{K_1, K_0}{=}1$ & 2.87 & 24.76 & 16.73\\
            \hline
        \end{tabular}
    \end{minipage}
\end{table}
\tabref{tab:app-miscoverage} provides mis-coverage results for different settings on MNIST, Fashion-MNIST and CIFAR10.
As in the main paper, we are able to reduce mis-coverage significantly on MNIST and Fashion-MNIST.
Only on CIFAR10, considering ``vehicles'' vs. ``animals'' as on CIFAR100 in the main paper, we are unable to obtain significant reductions.
While, we are able to reduce $\MisCov_{0\rightarrow1}$ slightly from 22.22\% to 20\%, $\MisCov_{{\color{red}1\rightarrow0}}$ increases slightly from 16.45\% to 16.73\% even for high off-diagonal weights used in $L$.
Compared to CIFAR100, this might be due to less flexibility to find suitable trade-offs as CIFAR10 has only 10 classes.
Moreover, mis-coverages on CIFAR10 are rather small to begin with, indicating that vehicles and animals do not overlap much by default.

\section{Additional Results on Binary Datasets}
\label{sec:app-binary}

\begin{figure}[b]
    \centering
    \begin{minipage}[t]{0.24\textwidth}
        \vspace*{0px}
    
        \includegraphics[height=1.4cm]{fig_wine_quality_ct_size_0}
        
        \includegraphics[height=1.4cm]{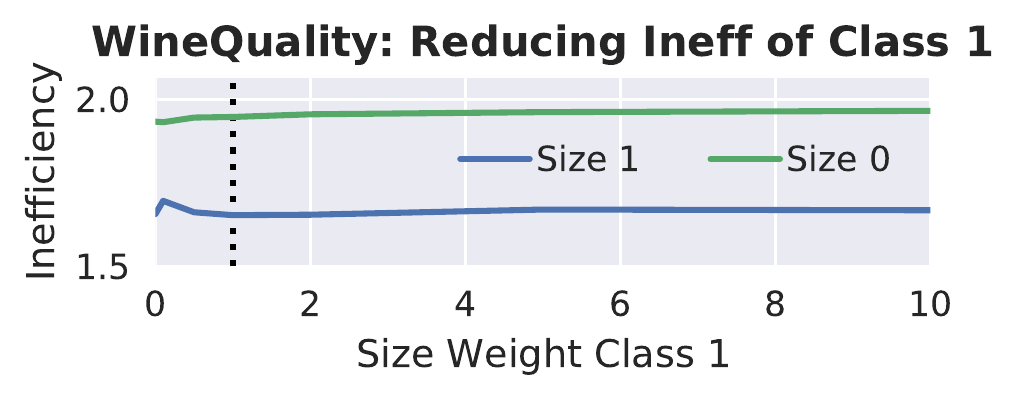}
    \end{minipage}
    \hskip 4px
    \begin{minipage}[t]{0.54\textwidth}
        \vspace*{0px}
    
        \includegraphics[height=1.4cm]{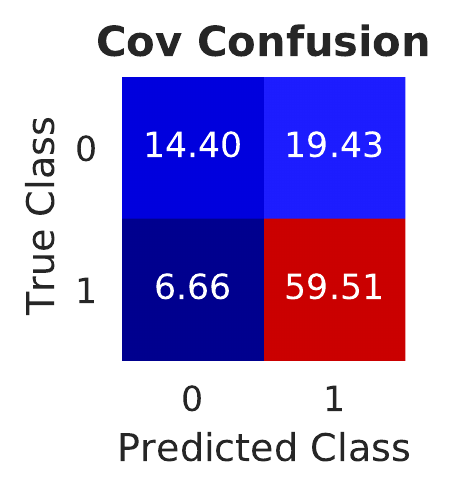}
        \includegraphics[height=1.4cm]{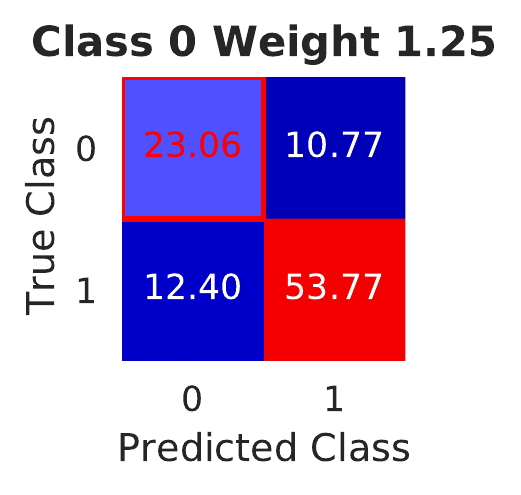}
        \includegraphics[height=1.4cm]{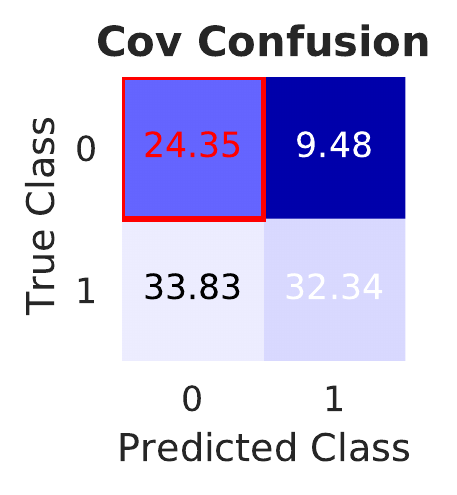}
        \includegraphics[height=1.4cm]{fig_wine_quality_ct_importance_0_0}
        \includegraphics[height=1.4cm]{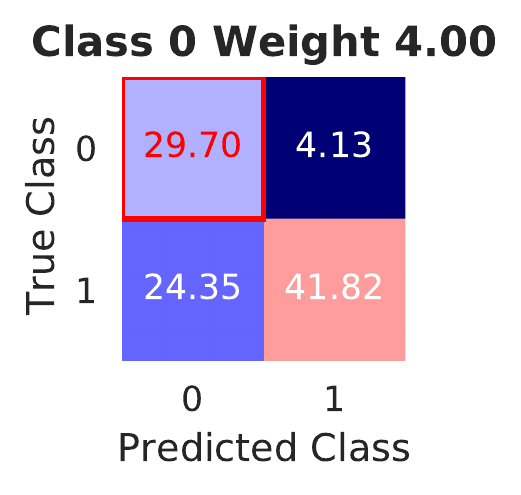}
        
        \includegraphics[height=1.4cm]{fig_wine_quality_ct_confusion}
        \hskip 2px
        \includegraphics[height=1.4cm]{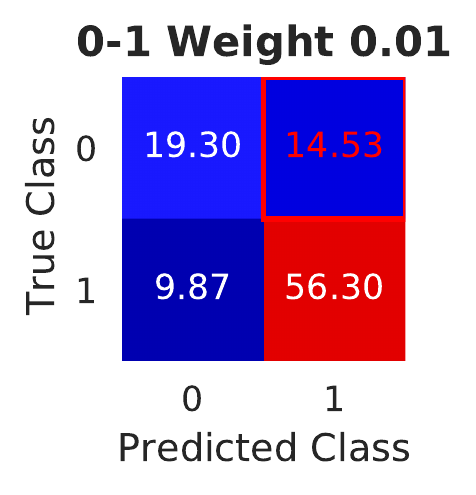}
        \hskip 2.5px
        \includegraphics[height=1.4cm]{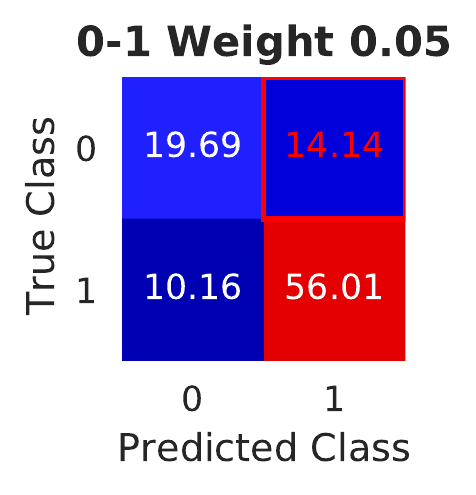}
        \hskip 2.5px
        \includegraphics[height=1.4cm]{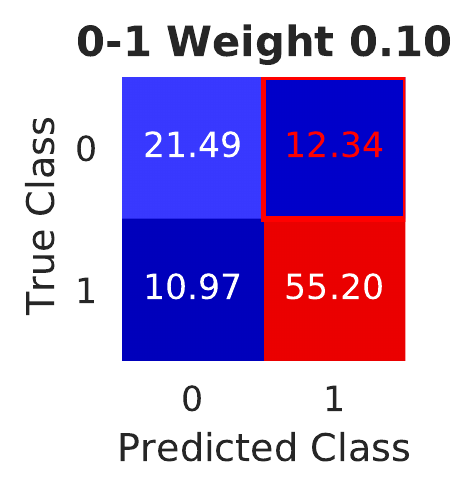}
        \hskip 2.5px
        \includegraphics[height=1.4cm]{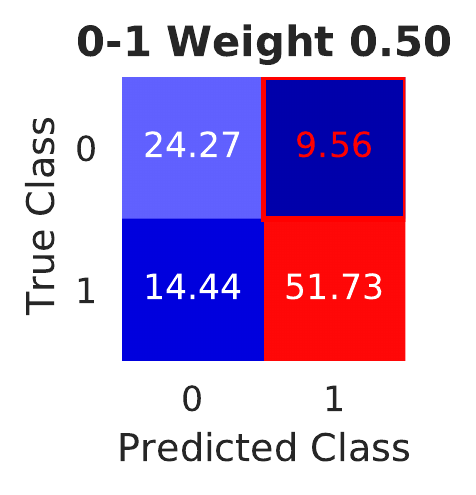}
    \end{minipage}
    \caption{
        \textbf{Manipulating Inefficiency and Coverage Confusion on WineQuality:}
        Complementing \figref{fig:miscoverage} (right) in the main paper, we plot the possible inefficiency reduction for class 1 (``good wine'', left) and full coverage confusion matrices for increased $L_{0,0} > 1$ and $L_{1, 0} > 0$ (right, top and bottom, respectively).
        While we can reduce inefficiency for class 0 (``bad wine''), this is not possible for class 1.
        However, class-conditional coverage for class 0 can be improved significantly and we can reduce coverage confusion $\Sigma_{0, 1}$.
    }
    \label{fig:app-binary-1}
\end{figure}
\figref{fig:app-binary-1} shows results complementing \figref{fig:miscoverage} (right) in the main paper.
Specifically, we show that reducing inefficiency for class 1 (``good wine'') is unfortunately not possible.
This might also be due to the fact that class 1 is the majority class, with $\sim$63\% of examples.
However, in addition to improving coverage conditioned on class 0, we are able to reduce coverage confusion $\Sigma_{0,1}$, \cf \secref{subsec:conformal-training-applications}.
We found that these results generalize to GermanCredit, however, being less pronounced, presumably because of significantly fewer training and calibration examples.

\section{Pseudo Code}
\label{sec:app-code}

\algref{alg:app-ct} presents code in Python, using Jax \citep{Bradbury2018}, Haiku \citep{Hennigan2020} and Optax \citep{Hessel2020}.
We assume access to a smooth sorting routine that allows to compute quantiles in a differentiable way: \mintinline{python}{smooth_quantile}.
Specifically, \algref{alg:app-ct} provides an exemplary implementation of \CT with (smooth) \Thr and \Lclass as outlined in \algref{alg:ct} in the main paper.
\mintinline{python}{smooth_predict_threshold} and \mintinline{python}{smooth_calibrate_threshold} implement differentiable prediction and calibration steps for \Thr.
These implementations are used in \mintinline{python}{compute_loss_and_error} to ``simulate'' \CP on mini-batches during training.
Size loss $\Omega$ from \eqnref{eq:size-loss} and classification loss from \eqnref{eq:classification-loss} are implemented in \mintinline{python}{compute_size_loss} and \mintinline{python}{compute_general_classification_loss}.
Note that the definition of \mintinline{python}{compute_loss_and_error} distinguishes between  \mintinline{python}{trainable_params} and \mintinline{python}{fixed_params}, allowing to fine-tune a pre-trained model.

\clearpage
\begin{algorithm}
\caption{
\textbf{Python Pseudo-Code for \CT:}
We present code based on our Python and Jax implementation of \CT.
In particular, we include smooth calibration and prediction steps for \Thr as well as the classification loss $\Lclass$ and the size loss $\Omega$.
Instead of including a full training loop, \mintinline{python}{compute_loss_and_error} shows how to compute the loss which can then be called using \mintinline{python}{jax.value_and_grad(compute_loss_and_error, has_aux=True)} and used for training using Optax.
Hyper-parameters, including \mintinline{python}{alpha}, \mintinline{python}{dispersion}, \mintinline{python}{size_weight}, \mintinline{python}{temperature}, \mintinline{python}{loss_matrix}, \mintinline{python}{size_weights} and \mintinline{python}{weight_decay}, are not defined explicitly for brevity.
\mintinline{python}{smooth_quantile} is assumed to be a provided differentiable quantile computation method.
Finally, \mintinline{python}{model} can be any Jax/Haiku model.
}
\label{alg:app-ct}
\begin{minted}[mathescape,linenos,numbersep=5pt,gobble=2,frame=lines,framesep=2mm,fontsize=\scriptsize]{python}
  import jax
  import jax.numpy as jnp
  import haiku as hk
  
  def smooth_predict_threshold(
      probabilities: jnp.ndarray, tau: float, temperature: float) -> jnp.ndarray:
    """Smooth implementation of prediction step for Thr."""
    return jax.nn.sigmoid((probabilities - tau) / temperature)
    
  def smooth_calibrate_threshold(
      probabilities: jnp.ndarray, labels: jnp.ndarray,
      alpha: float, dispersion: float) -> float:
    """Smooth implementation of the calibration step for Thr."""
    conformity_scores = probabilities[jnp.arange(probabilities.shape[0]), labels.astype(int)]
    return smooth_quantile(array, dispersion, (1 + 1./array.shape[0]) * alpha)
        
  def compute_general_classification_loss(
      confidence_sets: jnp.ndarray, labels: jnp.ndarray,
      loss_matrix: jnp.ndarray) -> jnp.ndarray:
    """Compute the classification loss Lclass on the given confidence sets."""
    one_hot_labels = jax.nn.one_hot(labels, confidence_sets.shape[1])
    l1 = (1 - confidence_sets) * one_hot_labels * loss_matrix[labels]
    l2 = confidence_sets * (1 - one_hot_labels) * loss_matrix[labels]
    loss = jnp.sum(jnp.maximum(l1 + l2, jnp.zeros_like(l1)), axis=1)
    return jnp.mean(loss)
    
  def compute_size_loss(
      confidence_sets: jnp.ndarray, target_size: int, weights: jnp.ndarray) -> jnp.ndarray:
    """Compute size loss."""
    return jnp.mean(weights * jnp.maximum(jnp.sum(confidence_sets, axis=1) - target_size, 0))

  FlatMapping = Union[hk.Params, hk.State]
  def compute_loss_and_error(
      trainable_params: FlatMapping, fixed_params: FlatMapping, inputs: jnp.ndarray,
      labels: jnp.ndarray, model_state: FlatMapping, training: bool, rng: jnp.ndarray,
  ) -> Tuple[jnp.ndarray, FlatMapping]:
    """Compute classification and size loss through calibration/prediction."""
    params = hk.data_structures.merge(trainable_params, fixed_params)
    # Model is a Haiku model, e.g., ResNet or MLP.
    logits, new_model_state = model.apply(params, model_state, rng, inputs, training=training)
    probabilities = jax.nn.softmax(logits, axis=1)

    val_split = int(0.5 * probabilities.shape[0])
    val_probabilities = probabilities[:val_split]
    val_labels = labels[:val_split]
    test_probabilities = probabilities[val_split:]
    test_labels = labels[val_split:]
    # Calibrate on the calibration probabilities with ground truth labels:
    val_tau = smooth_calibrate_threshold(val_probabilities, val_labels, alpha, dispersion)

    # Predict on the test probabilities:
    test_confidence_sets = smooth_predict_threshold(test_probabilities, val_tau, rng)
    # Compute the classification loss Lclass with a fixed loss matrix L:
    classification_loss = compute_general_classification_loss(
    test_confidence_sets, test_labels, loss_matrix)
    # Optionally set size weights determined by ground truth labels:
    weights = size_weights[test_labels]
    # Compute size loss multiplied by size weight:
    size_loss = size_weight * compute_size_loss(confidence_sets, weights)

    # Compute the log of classification and size loss:
    loss = jnp.log(classification_loss + size_loss + 1e-8)
    loss += weight_decay * sum(jnp.sum(jnp.square(param)) for param in jax.tree_leaves(params))

    return loss, new_model_state
\end{minted}
\end{algorithm}
\end{appendix}

\end{document}